\documentclass[acmtog,screen,balance=true]{acmart}

\renewcommand\footnotetextcopyrightpermission[1]{} % Removes footnote with conference information in first column

\acmSubmissionID{1687}

\usepackage{xspace}
\usepackage{graphicx}
\usepackage[percent]{overpic}
\usepackage{cleveref}
\usepackage{subfig}
\usepackage{tikz}
\usetikzlibrary{positioning}

\newcommand{\edit}[1]{{#1}}

\newcommand{\xx}{\textsf{X}\xspace}
\newcommand{\albedo}{\ensuremath{\mathbf{a}}\xspace}
\newcommand{\normal}{\ensuremath{\mathbf{n}}\xspace}
\newcommand{\material}{\ensuremath{\mathbf{m}}\xspace}
\newcommand{\irra}{\ensuremath{\mathbf{E}}\xspace}
\newcommand{\depth}{\ensuremath{\mathbf{z}}\xspace}
\newcommand{\direct}{\ensuremath{\mathbf{d}}\xspace}
\newcommand{\image}{\ensuremath{\mathbf{I}}\xspace}

\newcommand{\name}{\textsf{RGBX-Next}\xspace}
\newcommand{\rgbx}{\textsf{RGB$\leftrightarrow$X}\xspace}
\newcommand{\rgbxx}{\textsf{RGB$\times$X}\xspace}
\newcommand{\rgbtox}{\textsf{RGB$\rightarrow$X}\xspace}
\newcommand{\xtorgb}{\textsf{X$\rightarrow$RGB}\xspace}

\newcommand{\model}{f_\theta}

\begin{document}

\title{\name: Towards Realistic Generative Rendering from G-Buffers}

%% Authors
\author{Zheng Zeng}
\affiliation{%
  \institution{University of California, Santa Barbara, NVIDIA}
  \country{USA}
}

\author{Marco Salvi}
\affiliation{%
  \institution{NVIDIA}
  \country{USA}
}

\author{Lifan Wu}
\affiliation{%
  \institution{NVIDIA}
  \country{USA}
}

\author{Jan Nov\' ak}
\affiliation{%
  \institution{NVIDIA}
  \country{Czech Republic}
}

\author{Daqi Lin}
\affiliation{%
  \institution{NVIDIA}
  \country{USA}
}

\author{Saeed Hadadan}
\affiliation{%
  \institution{NVIDIA}
  \country{USA}
}

\author{Yichen Sheng}
\affiliation{%
  \institution{NVIDIA}
  \country{USA}
}

\author{Robert Pottorff}
\affiliation{%
  \institution{NVIDIA}
  \country{USA}
}

\author{Shiqiu Liu}
\affiliation{%
  \institution{NVIDIA}
  \country{USA}
}

\author{Ravi Ramamoorthi}
\affiliation{%
  \institution{NVIDIA, University of California, San Diego}
  \country{USA}
}

\author{Ling-Qi Yan}
\affiliation{%
  \institution{Mohamed bin Zayed University of Artificial Intelligence}
  \country{UAE}
}

\author{Milo\v s Ha\v san}
\affiliation{%
  \institution{NVIDIA}
  \country{USA}
}

\begin{abstract}
   % JAN: original version
   % Diffusion models have achieved impressive results in image, video and streaming generation, with minimal prompting. However, the key unresolved issue is the large remaining gap between the controllability of traditional 3D rendering and generative approaches. We believe a viable path forward is generative rendering controlled by traditionally rendered G-buffers. We introduce RGBX-Next, a unified generative rendering and inverse rendering framework for estimating G-buffers from images, videos and streams, and rendering realistic images, videos and streams from G-buffers. The key technical contribution of this paper is a general recipe for finetuning diffusion transformer (DiT) video models into generative rendering and inverse rendering models. We repurpose the frame budget of a DiT video
   % model (in our case, Wan 2.1) into input and output frames of various types
   % (RGB, G-buffer, image, video, input, output). Our \rgbtox models achieve state-of-the art quality intrinsic decompositions; furthermore, we demonstrate state-of-the-art realistic generative rendering using our \xtorgb models, trained on high-quality real video data annotated using our \rgbtox model. We introduce streaming $\rgbtox$ and $\xtorgb$ models for long video processing and generative rendering. All our models will be released; we also believe the design recipes introduced in this
   % paper can be useful for future generative inverse and forward models.

  Diffusion models have achieved impressive results in image, video, and streaming generation. \edit{However, compared to traditional 3D rendering, they still lack precise control over the generated output.
  We believe a viable path forward is to use generative models as learned renderers conditioned on traditionally rendered G-buffers.} We introduce \emph{RGBX-Next}, a unified generative framework for forward and inverse rendering, which allows estimating G-buffers from images, videos, and streams, and rendering realistic images, videos, and streams from G-buffers.
  Our key contribution is a general recipe for finetuning diffusion transformer (DiT) \edit{models} into generative forward and inverse renderers.
  %A text-to-video DiT model supports a variable number of output frames; we repurpose \edit{this frame budget} into \edit{typed input and output frames across modalities, including RGB and G-buffers.}
  %\edit{For better finetuning convergence, we introduce QK type embeddings and clean input tokens, giving the DiT explicit role, modality, and clean conditioning information.}
  %\edit{To improve realism of RGB outputs, we train on real RGB videos paired with G-buffers estimated by our inverse model.}
  %\edit{We further introduce guidance dropout to balance explicit control with generative freedom, and extend the models to long-video streaming with teacher forcing, long-context tokens, and self forcing.}
  \edit{We show that the resulting models achieve high quality in both realistic generative rendering and intrinsic decomposition.}
  We will make all our models publicly available. We believe that the design principles presented in this paper will benefit future research on \edit{controllable} generative forward and inverse \edit{rendering}.
\end{abstract}

%%
%% Generate your CCSCML using http://dl.acm.org/ccs.cfm.
%%

% \ccsdesc[500]{Computing methodologies~Rendering}

%% Keywords
% \keywords{Realist ic Rendering}

\begin{teaserfigure}
    \centering
    \begin{overpic}[width=1\linewidth]{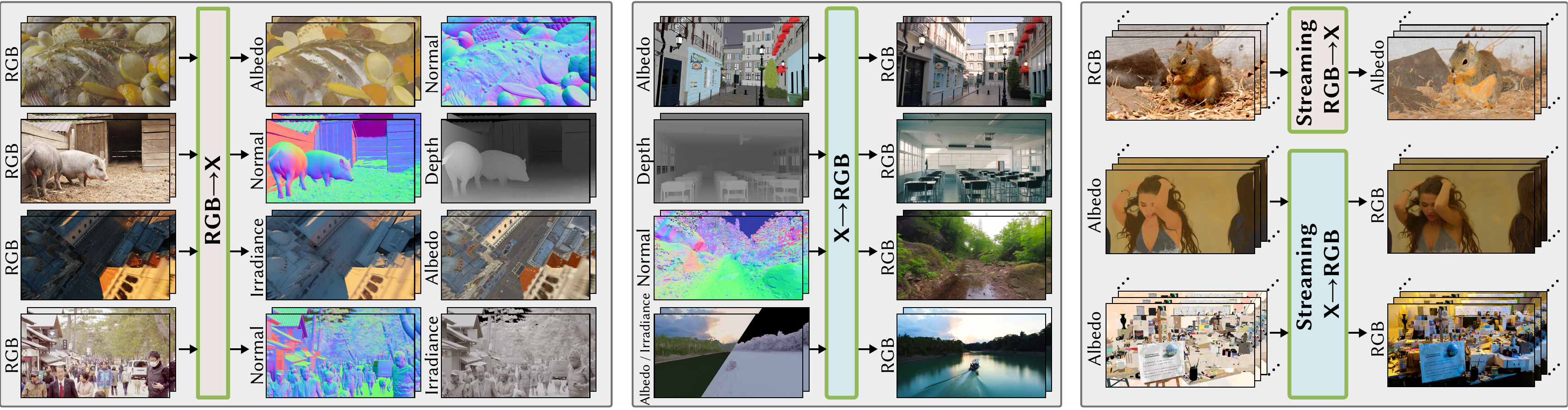}
        % \put(1,1){
        % \begin{tikzpicture} \node[fill=black!10,fill opacity=0.75,rounded corners=1ex, text width=0.7cm,align=left] {\footnotesize Albedo}; \end{tikzpicture}
        % }
        % \put(16.3,1){
        % \begin{tikzpicture} \node[fill=black!10,fill opacity=0.75,rounded corners=1ex, text width=0.7cm,align=left] {\footnotesize Albedo}; \end{tikzpicture}
        % }
        % \put(27.2,1){
        % \begin{tikzpicture} \node[fill=black!10,fill opacity=0.75,rounded corners=1ex, text width=1.0cm,align=left] {\footnotesize Irradiance}; \end{tikzpicture}
        % }
        % \put(42,1){
        % \begin{tikzpicture} \node[fill=black!10,fill opacity=0.75,rounded corners=1ex, text width=1.0cm,align=left] {\footnotesize Irradiance}; \end{tikzpicture}
        % }
        % \put(56.8,1){
        % \begin{tikzpicture} \node[fill=black!10,fill opacity=0.75,rounded corners=1ex, text width=1.0cm,align=left] {\footnotesize Irradiance}; \end{tikzpicture}
        % }
    \end{overpic}
  \caption{We present \name,  a unified generative framework for forward and
inverse rendering, which allows estimating G-buffers from images, videos, and streams, and rendering realistic images, videos, and streams from G-buffers. {\bf Left:} Our \rgbtox model estimates clean G-buffers from 17-frame real video segments. {\bf Middle:} Various \xtorgb models trained on real RGB data produce highly realistic results from synthetically rendered or estimated G-buffers. {\bf Right:} Streaming models finetuned from our video models produce coherent results over hundreds of frames, for \rgbtox as well as \xtorgb.}
  \label{fig:teaser}
\end{teaserfigure}

\maketitle

\section{Introduction}

Traditional physically-based rendering has achieved remarkable success. Progress has been made especially in performance, where previously offline Monte Carlo path tracing methods now deliver real-time performance in recent video games.
%
% The image realism achievable is also impressive; however, this depends on the quality of the 3D assets being rendered. Materials, 3D shapes and their layout, as well as lighting, need to be authored with high detail, expertise, and cost; these can often be afforded only by top game and movie productions.
%
The image realism, impressive as it is, still depends primarily on the quality of 3D assets.
Geometry, materials, and lighting need to be authored with very high detail, requiring expertise and incurring cost that only large game and movie productions can afford.

% On the other hand, video diffusion models have achieved impressive results in image, video and streaming generation \cite{liu2024sora,wan,kong2025hunyuan}. They deliver virtually unlimited content with minimal prompting, and their realism easily surpasses even the most high-end video games. Several streaming video models were introduced with performance already reaching interactivity (e.g. Self Forcing \cite{huang2025selfforcing} reports 10 fps on a single consumer-level NVIDIA 4090, while Seaweed-Apt2 \cite{lin2025apt2} reports 24fps on an H100).
%
On the other hand, diffusion models have achieved impressive results in image, video, and streaming generation \cite{liu2024sora,wan,kong2025hunyuan}. They deliver virtually unlimited content with minimal prompting, and their realism often surpasses even high-end video games. Several streaming video models were introduced with performance already reaching interactivity. For instance, Self Forcing \cite{huang2025selfforcing} and Seaweed-Apt2 \cite{lin2025apt2} models report 10 fps and 24 fps on a single consumer-level and a single datacenter GPU, respectively.

The quality and performance of diffusion models open up the possibility of generative rendering\edit{, where a diffusion model acts as a learned renderer for producing RGB frames,} for interactive and real-time applications. However, the key unresolved issue is the large gap between the controllability of traditional 3D rendering (with precise and temporally stable definitions of materials, objects, lights, and animations) and generative approaches (with very high-level, imprecise text or image control and temporal stability issues).

% We believe that a viable path to building a controllable generative renderer is to use a traditional (potentially simplified) 3D scene to render synthetic G-buffers, encoding per-pixel properties such as albedo, incoming lighting, normals, depth, and/or material parameters, and to train a streaming video diffusion model that turns these synthetic inputs into realistic final RGB frames. To obtain data to train such models, we also need to address the inverse problem of estimating G-buffers from RGB frames with temporal coherence.

We believe that a viable path towards building a controllable generative renderer is to leverage a synthetic \edit{(potentially simplified) 3D scene and condition a diffusion model on G-buffers rendered from that scene.}
In this setup, the diffusion model transforms synthetic information in the G-buffers (e.g., albedo, normals, depth, lighting, and material properties) into a sequence of realistic RGB frames (potentially much more realistic than the G-buffers themselves).
A key requirement for training such a model is a large paired dataset of \edit{realistic RGB frames and corresponding G-buffers}. \edit{Traditional 3D rendering can produce perfectly aligned pairs, but the RGB frames tend to retain a synthetic appearance. Creating sufficiently diverse, realistic scenes, assets, materials, and animations is prohibitively expensive and may be beyond practical human authoring capacity.} We therefore need to address the inverse problem: \edit{estimating temporally coherent G-buffers from real-world RGB videos.}

% Neural rendering from G-buffers has been explored by a number of previous methods \cite{deepshading,Li2022relighting}, but non-generative methods trained on small synthetic datasets hit fundamental limitations.

% The \rgbx framework \cite{zeng2024rgbx} introduced a unified approach for estimating G-buffers from images (\rgbtox), and rendering images from G-buffers (\xtorgb), based on a finetuned image diffusion model, thus building on the rich prior knowledge of the base model. The set of G-buffers (also called \emph{intrinsic channels}) is represented by the symbol \textsf{X}. In principle, an \xtorgb model can turn any set of G-buffers, synthetically rendered by traditional techniques, into final realistic images. However, the original \rgbx models only support static images and their training data is limited to indoor architectural scenes. DiffusionRenderer \cite{liang2025diffusionrenderer} has extended this framework to video segments, and added lighting control through an additional environment map input. However, the total number of frames in this approach is limited to whatever is supported by the base video diffusion model, and it requires the

The \rgbx framework \cite{zeng2024rgbx} introduced a unified approach for estimating G-buffers from RGB images (\rgbtox), and rendering RGB images from G-buffers (\xtorgb); \textsf{X} represents the G-buffers, also known as \emph{intrinsic channels}.
The method relies on a finetuned \emph{image} diffusion model and its rich prior knowledge.
% In principle, an \xtorgb model can turn any set of G-buffers, synthetically rendered by traditional techniques, into final realistic images. However, the original \rgbx models only support static images and their training data is limited to indoor architectural scenes.
DiffusionRenderer \cite{liang2025diffusionrenderer} has extended this framework to video segments, and added lighting control through an additional environment map input. However, the total number of frames is limited to what the base video model supports, and all conditioning properties must be specified for every pixel.
%- This approach has been further extended to relighting video segments \cite{relightingdiffusion} through a conditional input of the environment map.

% In this paper, we introduce \name, a unified generative rendering and inverse rendering framework for estimating G-buffers from images, videos and streams, and rendering realistic images, videos and streams from G-buffers.
In this paper, we introduce \name, a unified generative framework for forward and inverse rendering of images, videos, and streams.
A key technical contribution of this paper is a general recipe for \edit{finetuning diffusion transformer (DiT) models into generative forward (\xtorgb) and inverse (\rgbtox) renderers}.

A text-to-video DiT model (in our case, Wan 2.1 \cite{wan}) supports a variable number of output frames; \edit{we repurpose this frame budget into typed input and output frames across modalities, including RGB and G-buffers, through \emph{frame-wise concatenation}.}
\edit{To improve finetuning convergence, we introduce \emph{QK type embeddings} and \emph{clean input tokens}, giving the DiT explicit token role and modality information.}
%We improve quality by introducing \emph{QK type embedding} and \emph{clean input tokens}.
We further show that multiple input modalities can be packed into a single set of input tokens without loss of quality, using a technique termed \emph{X-patchify}.
We compare our framework to previous work based on channel-wise concatenation, such as DiffusionRenderer \cite{liang2025diffusionrenderer}, and context tokens, such as VACE \cite{jiang2025vace}, showing higher generality, quality and faster convergence of our approach.

\edit{We train on real videos paired with G-buffers estimated by our \rgbtox model, together with captions generated by a vision-language model (VLM) describing whether the videos look like real footage or synthetic content, and use caption-based classifier-free guidance~\cite{ho2022classifier} at inference for improved realism.}
\edit{We further introduce three ways of guidance dropout to balance explicit control with generative freedom: input G-buffers can be blurred, partially specified, or completely dropped during the diffusion iterations, allowing the model to creatively generate missing information. Lighting can be controlled in the same framework by providing an irradiance buffer or an albedo-free direct lighting buffer.}

\edit{Finally, we extend our models beyond fixed-length video clips to long-video streaming.
%We first support streaming with teacher forcing, where each chunk is trained using ground-truth previous outputs, and stitching-frame conditioning, where the previous chunk's boundary frame is reused as context for the next chunk. To break the perfection of teacher forcing, we perturb this stitching frame with brightness, sharpness, and VAE round-trip augmentations;
To provide long-term memory, we add long-context tokens, i.e., clean reference-frame tokens; to better match streaming inference, we apply self forcing \cite{huang2025selfforcing}, which trains the model on its own sampled outputs. Together, these components reduce error accumulation and enable stable long-video generative forward and inverse rendering.}
The contributions of our paper include:
\begin{itemize}
    \item We introduce a unified recipe for turning DiT video model architectures into $\rgbtox$ and $\xtorgb$ models for images and  videos. The recipe combines frame-wise concatenation, QK type embeddings, clean input tokens, and \xx-patchify. It produces high-quality results with fast finetuning, outperforming previous methods such as \rgbx, DiffusionRenderer, and VACE.
    \item We show state-of-the-art performance, in visual results and quantitative metrics, for both \xtorgb and \rgbtox models.
    \item We separate realistic, temporally stable diffuse lighting (irradiance) from real videos with \rgbtox. To our knowledge, this has not been shown before.
    \item We introduce a viable solution for improved realism in \xtorgb. We train on real videos paired with G-buffers estimated by our \rgbtox model, together with VLM-generated captions that describe whether videos look like real footage or synthetic content. At inference, we use caption-based classifier-free guidance to push outputs away from a synthetic look and towards realism.
    \item We introduce three ways to balance control and creativity in \xtorgb. During diffusion, G-buffers can be blurred, partially specified, or completely dropped. This also enables coarse lighting control in generative rendering, based on novel control signals: an irradiance buffer and an albedo-free direct illumination buffer.
    \item We introduce streaming $\rgbtox$ and $\xtorgb$ models for long-video generative forward and inverse rendering. To our knowledge, this has not been shown before. Our design combines teacher forcing, long-context tokens, and self forcing.
    % \item Our \rgbtox models achieve state-of-the art quality intrinsic decompositions.
    % \item We demonstrate state-of-the-art realistic generative rendering using our \xtorgb models, trained on high-quality real video data annotated using our streaming \rgbtox model.
\end{itemize}

The combination of these new contributions enables the high-quality results shown in our supplementary video and defines a new state-of-the-art for generative rendering from G-buffers. Finetuning diffusion models is currently largely an empirical undertaking, with limited theoretical foundations; however, our framework shows how to repurpose a DiT into any multi-input, multi-output, multi-modality model, making a step towards a general understanding of the field. All our models will be released to the community. %; we also believe the design recipes introduced in this paper can be useful for future generative inverse and forward models.

\section{Related Work}
\label{sec:related}
\begin{figure*}[t]
    \centering
    \begin{overpic}[width=1\linewidth]{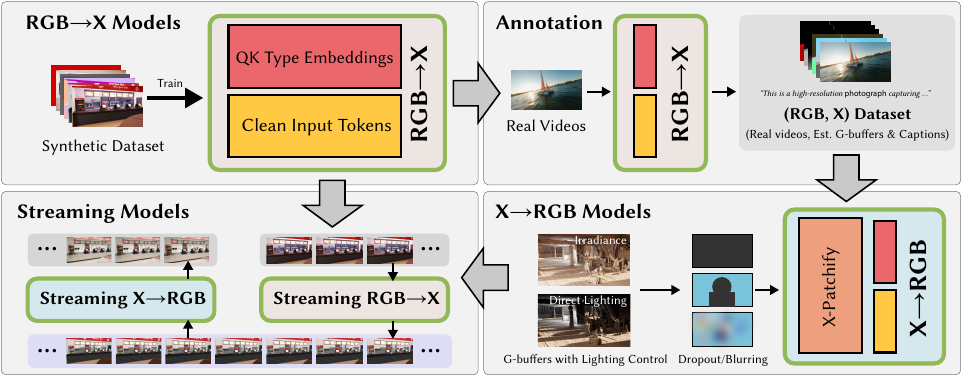}
    \put(6,34){(Sections \ref{sec:rgbtox})}
    \put(53,34){(Section \ref{sec:dataset})}
    \put(68.5,16.8){(Sections \ref{sec:xpatchify}--\ref{sec:lighting})}
    \put(21,16.8){(Section \ref{sec:streaming})}
    \end{overpic}
    \caption{High-level overview of our paper. We first describe a simple version of image and video \rgbx models in \cref{sec:initial-rgbx}, followed by an improved version using our \emph{QK type embedding} and \emph{clean input tokens} techniques in \cref{sec:improved-rgbx}. We estimate G-buffers from real video data with our \rgbtox model in \cref{sec:dataset}, and  train \xtorgb models with \emph{X-patchify} in \cref{sec:xpatchify}. We discuss conditioning dropout and lighting control in \cref{sec:dropout-blurring,sec:lighting}. We introduce streaming extensions of our models in \cref{sec:streaming}.}
    \label{fig:overview}
\end{figure*}
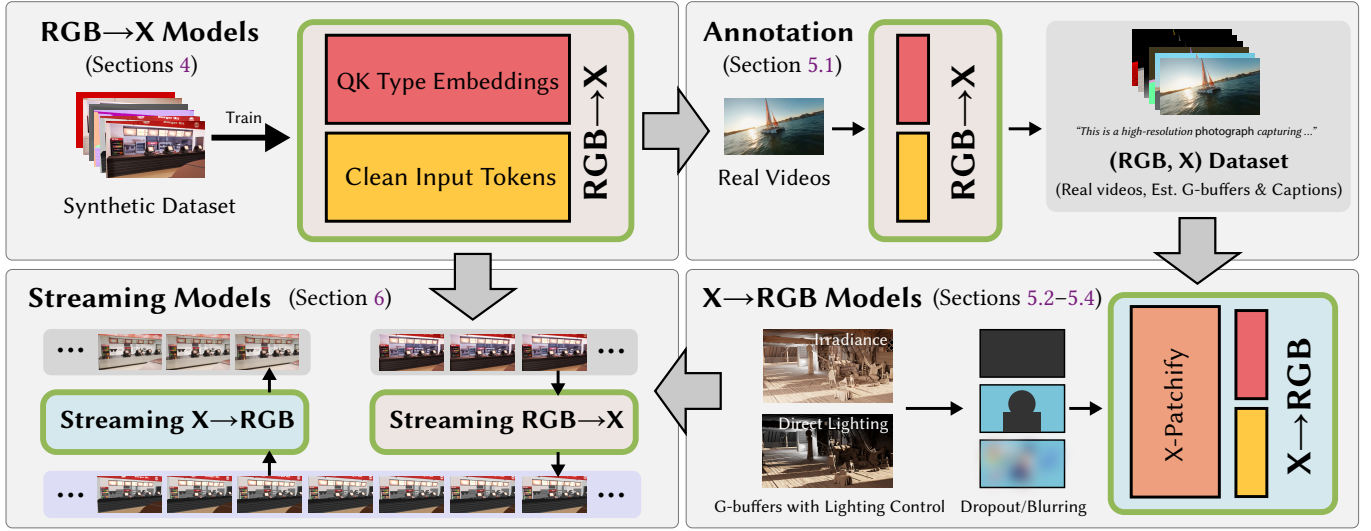

\paragraph{Generative image and video models}

Generative Adversarial Networks (GANs), especially StyleGAN variants \cite{goodfellow2014generative,karras2019style} were the first to achieve remarkably high quality image generation and flexible conditioning.
Diffusion models \cite{ho2020denoising} circumvent the instabilities of adversarial training and offer high-quality image generation with high diversity and broad finetuning opportunities.
While earlier diffusion models operated directly in pixel space \cite{Dalle2}, latent diffusion models (LDM) \cite{StableDiffusion} shifted the generative process to a compressed latent space, drastically reducing computational overhead.
More recently, diffusion models have been extended to video generation and diffusion transformers (DiT) \cite{peebles2023dit} have become the de-facto backbone for large-scale video synthesis with Sora \cite{liu2024sora}, eventually enabling state-of-the-art open source models such as Wan 2.1 \cite{wan}, and Hunyuan Video \cite{kong2025hunyuan}.
To repurpose these pretrained video foundation models for specialized conditioning, adapter-based techniques such as ControlNet \cite{ControlNet} and LoRA \cite{hu2021lora} were introduced for achieving structural and stylistic control.
Our work builds upon DiT models, specifically Wan 2.1, repurposing them into flexible, multi-modal models for realistic generative rendering and inverse rendering.

% generative adversarial networks (GANs) \cite{goodfellow2014generative}
% Achieved very high quality \cite{karras2019style}
% Diffusion models \cite{ho2020denoising}
% Pixel diffusion \cite{Dalle2}
% Latent diffusion models, Stable Diffusion \cite{StableDiffusion}
% Video diffusion models
% Diffusion transformers \cite{peebles2023dit}
% SORA \cite{liu2024sora}
% Wan 2.1 \cite{wan}
% Hunyuan video \cite{kong2025hunyuan}
% finetune pretrained models for new domains or conditioning, such as ControlNet~\cite{ControlNet} and LoRA \cite{hu2021lora}

\paragraph{Generative editing models}

Recent progress enabled semantic editing, where generative priors are steered via multi-modal conditioning.
Models such as Qwen-Image-Edit \cite{wu2025qwenimagetechnicalreport} and Flux Kontext \cite{labs2025flux1kontextflowmatching} leverage the reasoning capabilities of large language model backbones for versatile image manipulation, while VACE \cite{jiang2025vace} introduces an all-in-one framework designed for various video creation and editing tasks.
Our method enables lower-level editing by estimating G-buffers from input images, video, and streams, and then manipulating them.

% Qwen Image Edit \cite{wu2025qwenimagetechnicalreport}
% WAN Vace \cite{jiang2025vace}

\paragraph{Generative video streaming}

Research on video streaming emerged recently, with the goal of interactive frame rates, long-context temporal consistency, and low-latency generation.
FramePack \cite{zhang2025framepack} introduces efficient latent representation packaging tailored for video streaming.
CausVid \cite{yin2025causvid} adopts ``causal'' training protocols, where masked attention is used and ensures the generated latents to be conditioned strictly on the preceding temporal context, enabling true autoregressive generation.
Frameworks such as Self Forcing \cite{huang2025selfforcing}, Seaweed-APT2 \cite{lin2025apt2}, and LongLive \cite{yang2025longlive} have further pushed the limits of low-latency video generation and temporal coherence.
Our framework supports streaming (through a combination of teacher forcing and self forcing training) and is capable of processing long videos for generative rendering and inverse rendering. Distillation using methods such as DMD \cite{yin2024dmd} and DMD2 \cite{yin2024dmd2} is key to interactive performance; we do not apply distillation in this paper, though it is likely that these distillation methods can be applied to speed up our streaming models.

% FramePack \cite{zhang2025framepack}
% "Causal" training (instead of "bidirectional") uses masked attention, so any latent frames only depend on previous latent frames. \cite{yin2025causvid}
% Self forcing \cite{huang2025selfforcing}
% Seeweed APT2 \cite{lin2025apt2}
% LongLive \cite{yang2025longlive}

% LW: Move the following to the next paragraph.
% IRIS \cite{Li2020iris}
% IrisFormer \cite{IrisFormer}
% Ordinal Shading \cite{OrdinalShading}
% Intrinsic Image Diffusion \cite{kocsis2024intrinsic}

\paragraph{Neural forward and inverse rendering}

Pioneering approaches such as Deep Shading \cite{deepshading} employed convolutional neural networks to bridge the gap between synthetic G-buffers and final appearance.
CG-to-real image translation \cite{Bi2019DeepCS} uses a synthetic rendered image instead of G-buffers to condition realistic image synthesis.
CNN and transformer-based neural networks have also achieved significant progress in inverse rendering tasks, such as intrinsic image decomposition \cite{Li2020iris,IrisFormer,InteriorVerse,OrdinalShading}. The lighting editing approach for indoor scenes by \citet{Li2022relighting} is an earlier example of a framework combining neural inverse and forward rendering.

Diffusion models further boosted the achievable intrinsic decomposition quality \cite{kocsis2024intrinsic}.
The \rgbx framework \cite{zeng2024rgbx} unified these tasks by framing both inverse (\rgbtox) and forward (\xtorgb) rendering as diffusion-based image-to-image translation; it also introduced the idea of using the \rgbtox model, applied to real images, to augment the G-buffer training data for the \xtorgb model.
DiffusionRenderer \cite{liang2025diffusionrenderer} expanded this paradigm to video sequences and integrated environment-based lighting control.
Our work builds on these foundations by proposing a highly flexible DiT-based recipe that generalizes across images, videos, and streaming data, achieving high-quality forward and inverse rendering results. We also take inspiration from MaterialPicker \cite{ma2025materialpicker}, which repurposed the frames of a video DiT model for multiple G-buffer modalities and inputs/outputs, though for a different application (conditional generation of SVBRDF maps).
\edit{Concurrent work V-RGBX \cite{fang2025v} studies intrinsic-aware video editing by combining its video \rgbtox and \xtorgb models with keyframe edit propagation over albedo, normal, material, and irradiance controls. Another concurrent work, UniVidX \cite{chen2026unividx}, introduces a unified multimodal video framework whose model flexibly generates among RGB and intrinsic videos.}

% Deep Shading \cite{deepshading}
% CG2Real line of papers (Bi et al. 19)
% Editing of Indoor Scene Lighting from a Single Image \cite{Li2022relighting}
% InteriorVerse \cite{InteriorVerse}
% \rgbx \cite{zeng2024rgbx}
% DiffusionRenderer \cite{liang2025diffusionrenderer}

We do not specifically target relighting, but techniques used in previous relighting work are relevant, and our framework could be used as a foundation for downstream relighting methods. LightIt \cite{kocsis2024lightit}, Neural Gaffer \cite{jin2024neuralgaffer}, and ICLight \cite{zhang2025scaling} demonstrate the power of finetuning large-scale diffusion models with synthetic data for realistic relighting. UniRelight \cite{he2025unirelight} makes use of frame-wise concatenation and an estimated albedo buffer, related to our approach.

%\paragraph{Repurposing DiT video frames.}

\section{Repurposing DiT video models for our tasks}
\label{sec:overview}

%\zheng{Is ``Background'' a good section name here? How about ``''?}

\edit{Our main goal is to achieve controllable, realistic generative rendering: given explicit rendered G-buffers, \xtorgb models should synthesize realistic RGB frames. To train such models on real data, we also need \rgbtox models to recover G-buffers from real video frames.}
\edit{Across these tasks, RGB and different G-buffer channels may appear as either clean conditioning signals or generated outputs, and the available conditions may vary from one application to another.}
\edit{To achieve this, we develop a general \emph{recipe} for turning text-to-video diffusion transformer (DiT) models into multi-modality, multi-input, multi-output models.}

\edit{A central observation is that video DiTs operate on latent-frame sequences. This provides a natural conditioning mechanism: a subset of frame slots can encode clean inputs, while the remaining slots represent noisy outputs, with token-type information specifying each token's role and modality.}
\edit{We first introduce the relevant mechanics of DiT video models (\cref{sec:dit-background}), and then formalize this frame-repurposing mechanism for conditional generation (\cref{sec:dit-repurposing}). We provide a summary of the commonly used symbols throughout the article in \cref{tab:notation}.}

\begin{table}[t]
\centering
\footnotesize
\begin{tabular}{p{0.22\linewidth}|p{0.44\linewidth}|p{0.22\linewidth}}
\edit{\bf Symbol} & \edit{\bf Explanation} & \edit{\bf def.} \\
\hline
\edit{$m$, $\mathcal{M}$} & \edit{Modality index and modality set.} & \edit{\cref{eq:overview-sequence}, \cref{eq:overview-modalities}} \\
\edit{$\mathcal{M}_{\mathrm{in}}$, $\mathcal{M}_{\mathrm{out}}$} & \edit{Input and output modality sets.} & \edit{\cref{eq:overview-modalities}} \\
\edit{$\image$, $\albedo$, $\normal$, $\depth$, $\material$, $\irra$, $\direct$} & \edit{RGB and G-buffer modalities.} & \edit{\cref{eq:overview-modalities}} \\
\edit{$\mathbf{x}^{m}$, $\mathbf{x}^{m}_{t}$, $\boldsymbol{\epsilon}$} & \edit{Clean/noisy latent for modality $m$, and Gaussian noise.} & \edit{\cref{eq:overview-latent}, \cref{eq:overview-flow}} \\
\edit{$\mathbf{v}_{\theta}$, $\mathcal{L}_{\mathrm{FM}}$} & \edit{Predicted flow velocity and flow-matching loss.} & \edit{\cref{eq:overview-velocity}, \cref{eq:overview-flow}} \\
\edit{$\mathcal{P}$, $\mathcal{P}_{\xx}$} & \edit{Patchify and \xx-patchify maps.} & \edit{\cref{eq:overview-patchify}, \cref{eq:x-patchify}} \\
\edit{$\ell=(i,j,k)$, $\mathbf{u}_{\ell}^{m}$} & \edit{Token index and token.} & \edit{\cref{eq:overview-patchify}} \\
\edit{$\mathcal{U}$} & \edit{Internal token layout after patchification.} & \edit{\cref{eq:framewise-concat}, \cref{eq:unified-token-set}} \\
\edit{$\tau=(r,m)$} & \edit{Token type: role $r$ and modality $m$.} & \edit{\cref{sec:improved-rgbx}} \\
\edit{$\mathbf{e}_{Q}^{\tau}$, $\mathbf{e}_{K}^{\tau}$} & \edit{Learned additive query/key type embeddings.} & \edit{\cref{eq:qk-type-embedding}} \\
\edit{$\tilde{t}_{\ell}^{\tau}$} & \edit{Token-specific timestep for clean input tokens.} & \edit{\cref{eq:clean-token-time}} \\
\end{tabular}
\caption{\edit{Core notation used in the paper.}}
\label{tab:notation}
\end{table}

%We would like the framework to support a varying number of input and output G-buffers, and to apply to images, short video segments, and longer video streaming.
%, by reusing the concept of \emph{frames} for different modalities and behaviors.

\subsection{DiT video models}
\label{sec:dit-background}
\begin{figure}[tb]
    \centering
    \includegraphics[width=0.7\linewidth]{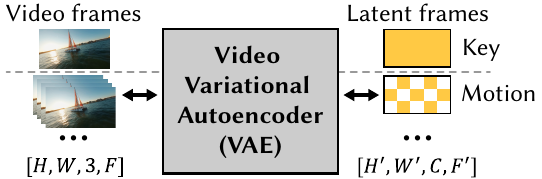}
    \caption{A typical variational auto-encoder (VAE) of recent video models offers $8 \times 8 \times 4$ compression.
    More precisely, it encodes key frames to latent key frames, and further groups of 4 frames into latent motion frames, thus mapping $4k+1$ RGB frames into $k+1$ latent frames of $8\times$ lower spatial resolution and 16 latent channels.
    }
    \vspace{-1em}
    \label{fig:vae}
\end{figure}

The goal of diffusion transformer (DiT)~\cite{peebles2023dit} text-to-video models is to generate a video (sequence of RGB frames) of size $H \times W \times 3 \times F$.
\edit{We denote a data modality by $m$ and the set of modalities under consideration by $\mathcal{M}$; for a standard text-to-video model, $m$ simply refers to RGB video frames. We write the $F$-frame video sequence generated by the model as $\mathbf{Y}_{1:F}^{m}$:}
\begin{equation}
\edit{\mathbf{Y}_{1:F}^{m} = (\mathbf{Y}_{1}^{m}, \ldots, \mathbf{Y}_{F}^{m}), \qquad m \in \mathcal{M}.}
\label{eq:overview-sequence}
\end{equation}
\edit{Recent video models, including Wan \cite{wan}, HunyuanVideo \cite{kong2025hunyuan} and Cosmos \cite{agarwal2025cosmos}, operate in a latent space with a pretrained video variational autoencoder (VAE).
The compression ratio is often $8 \times 8$ spatially and $4$ temporally, encoding $4k+1$ original video frames to $k+1$ latent frames; each key frame becomes a \emph{latent key frame}, and each following group of 4 frames becomes a \emph{latent motion frame}.
As an example, Wan 2.1 video model~\cite{wan} operates on latents $x$ of shape $H' \times W' \times C \times F'$, where $H' = H/8$, $W' = W/8$, $F' = (F+3)/4$ and $C = 16$.} %The native generated clip is $1280 \times 720 \times 3 \times 81$ RGB frames, which becomes $160 \times 90 \times 16 \times 21$ latent frames. The model can operate on different resolutions, aspect ratios and numbers of frames, though the generation quality may degrade as these diverge from "native" resolutions and frame numbers, for which the model has been trained most extensively.
\edit{We denote the VAE encoding of the sequence by $\mathbf{x}^{m}$, a compressed latent tensor that is smaller spatially and temporally than the original frames:}
\begin{equation}
\edit{\mathbf{x}^{m} = \mathcal{E}_{\mathrm{vae}}(\mathbf{Y}_{1:F}^{m}), \qquad
\mathrm{shape}(\mathbf{x}^{m}) = H' \times W' \times C \times F'.}
\label{eq:overview-latent}
\end{equation}
\edit{The map $\mathcal{E}_{\mathrm{vae}}$ is the VAE encoder.}
\edit{The tensor $\mathbf{x}^{m}$ is what the diffusion model sees after VAE compression.}
See \cref{fig:vae} for an illustration of the VAE and the corresponding tensor sizes.

\begin{figure}[tb]
    \centering
    \includegraphics[width=1\linewidth]{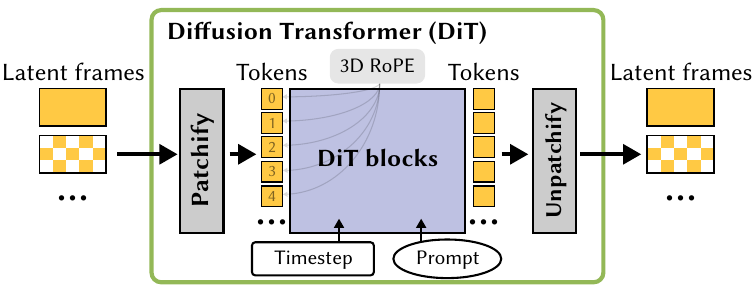}
    \caption{A DiT video model operates on latent frames as inputs and outputs. It patchifies $2 \times 2$ blocks of each latent frame into \emph{tokens}, and operates on the set of tokens in transformer layers, which also receive text prompt embeddings and timestep information through cross-attention. Token order is arbitrary; embeddings are used to tell the model which token is which.}
    \vspace{-1em}
    \label{fig:dit}
\end{figure}

At inference time, the diffusion process produces a sequence of progressively denoised latent estimates, $(x_1, ..., x_0)$, where $x_1$ is pure Gaussian noise and $x_0$ is the final generated latent video.
The DiT model $\model$ features a transformer network that returns flow velocity given a timestep $t$, text prompt $p$ and current partially denoised latent $x_t$. \edit{At each sampling step, the network predicts a flow velocity for the current noisy latent, conditioned on timestep and prompt:}
\begin{equation}
\edit{\mathbf{v}_{\theta} = \model(\mathbf{x}_t, t, p), \qquad t \in [0,1].}
\label{eq:overview-velocity}
\end{equation}
\edit{The vector $\mathbf{v}_{\theta}$ denotes the predicted flow velocity, and the latent $\mathbf{x}_t$ is progressively updated toward smaller timesteps during sampling.}

\edit{We denote the full iterative denoising process by $\operatorname{Sample}_{\theta}(p)$: it starts from Gaussian noise and repeatedly evaluates the DiT under prompt $p$ until reaching the final latent.}

\edit{For training, a clean output latent is linearly mixed with Gaussian noise, and the training target is the velocity from data to noise:}
\begin{equation}
\edit{\begin{aligned}
\mathbf{x}^{m}_t
&= (1-t)\mathbf{x}^{m}_0 + t\boldsymbol{\epsilon},\\
\mathcal{L}_{\mathrm{FM}}
&=
\mathbb{E}_{m,t,\boldsymbol{\epsilon}}\!\left[
\left\| \model(\mathbf{x}^{m}_t,t,p) - (\boldsymbol{\epsilon}-\mathbf{x}^{m}_0) \right\|_2^2
\right].
\end{aligned}}
\label{eq:overview-flow}
\end{equation}
\edit{The loss $\mathcal{L}_{\mathrm{FM}}$ is the flow-matching objective; $\mathbf{x}^{m}_0$ is the clean training latent for modality $m$, $\boldsymbol{\epsilon}$ is Gaussian noise, and the expectation averages over modalities, timesteps, and noise samples.}

The transformer does not operate directly on $x_t$ but rather on tokens, each obtained by ``patchifying'' $2 \times 2 \times C$ blocks of latent frames into a single vector.
\edit{For a latent patch at spatial index $(i,j)$ and latent frame $k$, patchification produces a token from the corresponding local latent block:}
\begin{equation}
\edit{\mathbf{u}_{i,j,k}^{m} =
\mathcal{P}\!\left(\mathbf{x}^{m}_{2i:2i+1,\,2j:2j+1,\,:,k}\right).}
\label{eq:overview-patchify}
\end{equation}
\edit{The map $\mathcal{P}$ is the patchify operation, and $\mathbf{u}_{i,j,k}^{m}$ denotes the resulting token of modality $m$; its length is $T$.}
\edit{For example, in Wan 2.1, a $1280 \times 720$ video with 81 frames produces a $160 \times 90 \times 16 \times 21$ latent tensor, which patchifies into $75,600$ tokens of length $6,144$.}
The number and size of tokens remain unchanged while passing through the DiT layers. Finally, an \emph{unpatchify} operation (also called the DiT \emph{head}) maps each token into a $2 \times 2 \times C$ block of the final latent.

\edit{The DiT identifies each token's spatial position and frame index through positional encoding, implemented as a variant of 3D RoPE \cite{su2024roformer}. The 2D spatial position and frame ID form a triple that modulates the attention queries and keys.}
\edit{In each self-attention layer, RoPE uses this triple to transform the query and key derived from a patchified token:}
\begin{equation}
\edit{\mathbf{Q}_{i,j,k}^{m} =
\mathcal{R}_{i,j,k}\!\left(W_Q\mathbf{u}_{i,j,k}^{m}\right), \qquad
\mathbf{K}_{i,j,k}^{m} =
\mathcal{R}_{i,j,k}\!\left(W_K\mathbf{u}_{i,j,k}^{m}\right).}
\label{eq:overview-rope}
\end{equation}
\edit{The matrices $W_Q$ and $W_K$ are the query and key projections, and $\mathcal{R}_{i,j,k}$ is the 3D RoPE rotation determined by the patch position. Thus the token $\mathbf{u}_{i,j,k}^{m}$ carries local latent content, while RoPE injects position into self-attention through $\mathbf{Q}_{i,j,k}^{m}$ and $\mathbf{K}_{i,j,k}^{m}$.}

Text prompt $p$ is injected into the DiT via cross-attention, while the timestep embedding $t$ conditions the DiT blocks through adaptive LayerNorm modulation~\cite{peebles2023dit}. See \cref{fig:dit} for an illustration of the DiT, tokens, embeddings, and patchify and unpatchify operations.

%\edit{Importantly, after patchification, the DiT operates on a collection of tokens whose spatial and temporal identities are supplied by positional embeddings rather than by their literal order in memory.}
%\edit{This token interface is a useful source of flexibility: in the next subsection, we repurpose latent-frame tokens as clean conditions or noisy outputs, and use type information to specify their roles and modalities.}

\subsection{Repurposing DiT frames}
\label{sec:dit-repurposing}

In our desired forward and inverse rendering tasks, we need to support multiple modalities, some of which act as inputs and some as outputs.
In forward rendering, the model takes one or more G-buffer frames as input and produces RGB frames $\image$ as output. In inverse rendering, this process is reversed: the input is $\image$, and the output is one or more corresponding G-buffers.
\edit{Throughout the paper, RGB denotes final color images or videos, and \textsf{X} denotes the corresponding G-buffer modalities.}
\edit{Thus, our modality set $\mathcal{M}$ contains RGB together with the G-buffer channels that our models may consume or generate:}
\begin{equation}
\edit{\mathcal{M}_{\mathrm{in}}, \mathcal{M}_{\mathrm{out}} \subseteq \mathcal{M}.}
\label{eq:overview-modalities}
\end{equation}
\edit{The subsets $\mathcal{M}_{\mathrm{in}}$ and $\mathcal{M}_{\mathrm{out}}$ specify which modalities are provided as clean inputs and which modalities are outputs.}

\edit{Given the input and output sets above, we place the corresponding latents into different latent frame slots of the video DiT. A text-to-video DiT model is originally designed to take noisy latent frames and produce clean latent frames, which after a number of denoising iterations become the final output of the model. We repurpose this frame interface by treating some frame slots as inputs: output frames have noise added and are denoised, while input frames are provided without noise and the corresponding tokens can be discarded after processing by the DiT.}

\edit{We refer to this mechanism as \emph{frame-wise concatenation}: it arranges clean input (conditioning) latents and noisy output latents in a single frame layout.}

\begin{equation}
\edit{
\mathbf{x}_{\mathrm{fw}}(t)=
\left(
\{\mathbf{x}_{0}^{m}\}_{m\in\mathcal{M}_{\mathrm{in}}},
\{\mathbf{x}_{t}^{m}\}_{m\in\mathcal{M}_{\mathrm{out}}}
\right).}
\label{eq:framewise-latents}
\end{equation}
\edit{$\mathbf{x}_{\mathrm{fw}}(t)$ is the latent collection passed to the denoising model; $\mathbf{x}_{0}^{m}$ is the clean latent for modality $m$, $\mathbf{x}_{t}^{m}$ is the corresponding noisy output latent at diffusion timestep $t$, and $\mathcal{M}_{\mathrm{in}}$ and $\mathcal{M}_{\mathrm{out}}$ choose the input and output modalities.}
\edit{The equations below use frame IDs only to separate input and output roles for clarity; in principle, distinct frame-ID ranges could also mark different modalities, such as RGB and G-buffers.}
\edit{Following the frame-ID convention above, the noisy output latents use the first $K$ latent-frame positions and the clean input latents use the next $K$ positions:}
\begin{equation}
\edit{
\ell_{\mathrm{out}}=(i,j,k), \qquad
\ell_{\mathrm{in}}=(i,j,k+K), \qquad 0\le k<K.}
\label{eq:framewise-frame-ids}
\end{equation}
\edit{The indices $\ell_{\mathrm{out}}$ and $\ell_{\mathrm{in}}$ share the same spatial patch $(i,j)$ and paired frame index $k$, but occupy different RoPE frame-ID positions. Inside the DiT, the latent collection is patchified into separate input and output tokens:}
\begin{equation}
\edit{
\mathbf{u}_{\ell_{\mathrm{in}}}^{m,\mathrm{in}} =
\mathcal{P}\!\left(\mathbf{x}^{m}_{0,\ell_{\mathrm{in}}}\right), \quad
\mathbf{u}_{\ell_{\mathrm{out}},t}^{m,\mathrm{out}} =
\mathcal{P}\!\left(\mathbf{x}^{m}_{t,\ell_{\mathrm{out}}}\right).}
\label{eq:framewise-tokens}
\end{equation}
\edit{The map $\mathcal{P}$ patchifies each latent patch into a token. The internal frame-wise layout is the union:}
\begin{equation}
\edit{\begin{aligned}
\mathcal{U}_{\mathrm{fw}}(t)=
\left\{\mathbf{u}_{\ell_{\mathrm{in}}}^{m,\mathrm{in}} : m \in \mathcal{M}_{\mathrm{in}}\right\}
\cup
\left\{\mathbf{u}_{\ell_{\mathrm{out}},t}^{m,\mathrm{out}} : m \in \mathcal{M}_{\mathrm{out}}\right\}.
\end{aligned}}
\label{eq:framewise-concat}
\end{equation}
\edit{The set $\mathcal{U}_{\mathrm{fw}}(t)$ is the token sequence processed by DiT layers after the latents are patchified. We still write the denoising model as $\model(\mathbf{x}_{\mathrm{fw}}(t),t,p)$ because patchification is an internal part of the DiT.}
\edit{Only output latents have noise added and loss computed on them; clean input latents provide context and are discarded after the DiT pass. The corresponding masked flow-matching loss is}
\begin{equation}
\edit{
\mathcal{L}_{\mathrm{fw}} =
\mathbb{E}\!\left[
\sum_{m\in\mathcal{M}_{\mathrm{out}}}
\left\| \model(\mathbf{x}_{\mathrm{fw}}(t),t,p)^{m} -
(\boldsymbol{\epsilon}^{m}-\mathbf{x}^{m}_{0})\right\|_2^2
\right].}
\label{eq:framewise-loss}
\end{equation}
\edit{The loss $\mathcal{L}_{\mathrm{fw}}$ sums only over $m\in\mathcal{M}_{\mathrm{out}}$, and the superscript $m$ on $\model(\cdot)^m$ selects the predicted velocity for output modality $m$. This formulation is independent of the particular rendering direction: \rgbtox uses RGB as clean input and G-buffers as noisy outputs, while \xtorgb reverses the roles and may use multiple G-buffer inputs.}

\section{\rgbtox models for inverse rendering}
\label{sec:rgbtox}

\subsection{Preliminary \rgbtox model}
\label{sec:initial-rgbx}

\begin{figure}[t]
    \centering
    \includegraphics[width=1\linewidth]{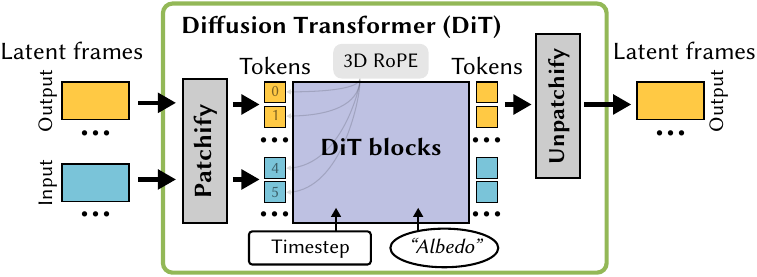}
    \caption{The preliminary \rgbtox model concatenates the output \xx and input RGB, distinguished by frame ID. Input tokens have no noise added and no loss computed on them. The text prompt is repurposed for output modality selection, e.g. "albedo", "normal", etc.}
    \label{fig:frame-wise-concat}
\end{figure}

%In generative inverse rendering (\rgbtox), our goal is to take RGB frames %$\image$ as input, and produce G-buffer frames
%% , for example the albedo frames $\albedo$,
%as output.
%A straightforward way to obtain such a model is to finetune an existing one to understand the text prompt as a switch between different output modalities~\cite{zeng2024rgbx}, e.g. using ``albedo'' or ``normal'' as the prompt.
%To distinguish which frame(s) are input vs. output, we can repurpose the \emph{frame ID} mechanism from the RoPE embedding, where the first $K$ latent frames represent outputs, while the next $K$ are inputs. These techniques already lead to well-performing \rgbtox models.

\edit{In generative inverse rendering (\rgbtox), our goal is to take RGB frames $\image$ as input and produce G-buffer frames as output.}
\edit{The G-buffer modalities considered in our models include albedo $\albedo$, normal $\normal$, depth $\depth$, material properties $\material$ (such as roughness, metallicity, or transparency), diffuse irradiance $\irra$, and direct lighting \direct.}
\edit{Using the frame-wise concatenation formulation from \cref{sec:dit-repurposing}, our preliminary \rgbtox model uses RGB as input and predicts one G-buffer modality at a time:}
\begin{equation}
\edit{
\mathcal{M}_{\mathrm{in}}=\{\image\}, \qquad
\mathcal{M}_{\mathrm{out}}\in
\bigl\{
\{\albedo\},\{\normal\},\{\depth\},\{\material\},\{\irra\},\{\direct\}
\bigr\}.}
\label{eq:rgbtox-modalities}
\end{equation}
\edit{The desired output modality is selected by finetuning the text prompt as a switch~\cite{zeng2024rgbx}, e.g., using ``albedo'' or ``normal'' as the prompt.}

\edit{This construction is intentionally minimal; see \cref{fig:frame-wise-concat}. The frame-ID convention indicates which frames are clean input frames versus noisy output frames, and the prompt indicates the desired output modality. These indirect signals are sufficient to obtain a preliminary \rgbtox model, but they leave role (input/output) and modality only weakly specified. The next \cref{sec:improved-rgbx} replaces these implicit signals with explicit token-level mechanisms.}

\subsection{Improved \rgbtox model}
\label{sec:improved-rgbx}

\begin{figure}[t]
    \footnotesize
    \centering
    \begin{overpic}[width=1\linewidth]{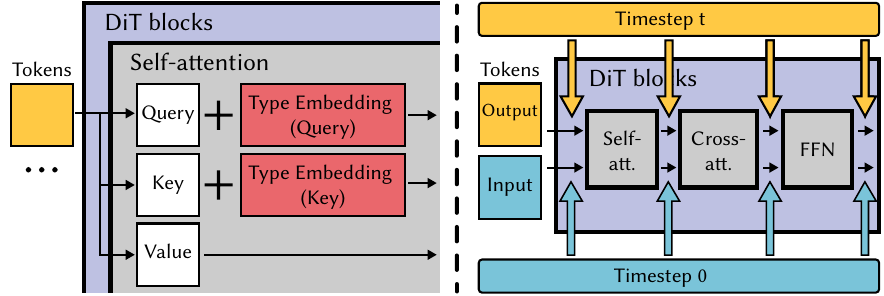}
    \put(28,-3.5){(a)}
    \put(76,-3.5){(b)}
    \end{overpic}
    \caption{(a) Our \emph{QK type embedding} encodes modality and input/output by adding learned offsets to the query and key vectors. (b) Our \emph{clean token} technique modifies the timestep embedding mechanism to always feed timestep 0 (i.e., no noise) to the input tokens, resulting in faster and higher-quality finetuning.}
    \label{fig:qk-clean}
\end{figure}

%In this section, we introduce QK type embedding and clean token embedding that are compatible with frame-wise concatenation and can speed up convergence and improve the results.
%The simplest \rgbtox model above utilizes the frame ID mechanism to distinguish input and output tokens, and the prompt to control the output modality.
%A better approach is to embed this information into the tokens.
% In this section, we introduce QK type embedding and clean token embedding that speed up convergence and improve results of all our models.
% A better approach than using the frame ID to distinguish input and output tokens, and the prompt to control the modality, is to embed this information directly into tokens.
\edit{Frame-wise concatenation already separates clean inputs from noisy outputs, but the baseline identifies these roles only through the frame-ID convention in \cref{eq:framewise-frame-ids}.}
A better approach than using the frame-ID to distinguish input and output tokens is to embed this information directly into tokens.
\edit{For each discrete combination of token role and modality, we define a token type $\tau=(r,m)$, where role $r\in\{\mathrm{in},\mathrm{out}\}$ and $m$ is the modality.}
Instead of directly adding a type embedding to the tokens,
we train the model to add learned embeddings directly to the query and key vectors in the DiT layers; see \cref{fig:qk-clean} (a).
\edit{For a token $\mathbf{u}_{\ell}^{\tau}$, each attention layer applies RoPE as usual and then adds learned type offsets to the attention queries and keys:}
\begin{equation}
\edit{
\mathbf{Q}_{\ell}^{\tau} =
\mathcal{R}_{\ell}\!\left(W_Q\mathbf{u}_{\ell}^{\tau}\right)+\mathbf{e}_{Q}^{\tau}, \qquad
\mathbf{K}_{\ell}^{\tau} =
\mathcal{R}_{\ell}\!\left(W_K\mathbf{u}_{\ell}^{\tau}\right)+\mathbf{e}_{K}^{\tau}.}
\label{eq:qk-type-embedding}
\end{equation}
\edit{The vectors $\mathbf{e}_{Q}^{\tau}$ and $\mathbf{e}_{K}^{\tau}$ are learned additive query/key type embeddings, and $\mathcal{R}_{\ell}$ is the RoPE transform for token index $\ell$. Thus RoPE injects position through a rotation, while QK type embedding injects role and modality through additive offsets, e.g., clean RGB input versus noisy albedo output.}
We show that this modification results in faster finetuning convergence.
%Importantly, it also frees up the text prompt for its original purpose.
We call this modification \emph{QK type embeddings}.

% Using frame-wise concatenation, i.e. having separate sets of tokens for inputs and outputs, also permits using our two new techniques, QK type embedding and clean token embedding, which can further speed up convergence and improve the results.

% In the simplest \rgbtox model above, we reused the text prompt and frame ID mechanisms to tell the model which tokens are which and what is the desired output modality. A better and more flexible solution is to embed this information into the tokens themselves. We could simply learn a mapping from the discrete combinations of input/output and modality to an embedding vector of size $T$ (the token size, which is 6144 in our models) and add the embedding to the tokens. We introduce an improvement: instead of adding to tokens directly, we learn to add the embedding to the query and key vectors in the DiT layers; see \cref{fig:qk-clean} (a). We show that this modification results in faster finetuning convergence; furthermore, it frees the text prompt for its original use. We call this modification \emph{QK type embeddings}.

Furthermore, we note that in a standard DiT~\cite{peebles2023dit}, the same time step embedding is applied through DiT blocks to all tokens. This makes sense, as all tokens are being denoised at the same noise level. However, in our case, some tokens correspond to input frames, which are provided without noise. We therefore modify the timestep embedding to provide $t=0$ to the tokens corresponding to input frames; see \cref{fig:qk-clean} (b). \edit{Equivalently, each token receives an effective timestep}
\begin{equation}
\edit{
\tilde{t}_{\ell}^{\tau} =
\begin{cases}
0, & r(\tau)=\mathrm{in},\\
t, & r(\tau)=\mathrm{out}.
\end{cases}}
\label{eq:clean-token-time}
\end{equation}
\edit{The function $r(\tau)$ extracts the input/output role from the token type, and $\tilde{t}_{\ell}^{\tau}$ is the timestep embedding used for that token. An input token is therefore embedded as a clean latent, while an output token keeps the current diffusion timestep.}
This modification, which we call \emph{clean input tokens}, further improves quality and finetuning convergence.

\edit{Training \rgbtox models only requires paired synthetic data: rendered RGB frames and the corresponding G-buffers. Several public datasets already provide such data, such as Hypersim~\cite{hypersim} and InteriorVerse~\cite{InteriorVerse}. To improve data quality and diversity, we curate an internal collection of synthetic scenes. The collection covers both indoor and outdoor environments. We then render an internal dataset from these scenes. It contains 900 video sequences, each with roughly 100 frames of path-traced RGB and corresponding G-buffers.}

% \paragraph{Results.} \todo{QK Figure} \todo{Clean token Figure}

% \paragraph{Summary.} Information about modality and input/output can be added to the query and key vectors in the DiT layers through a learned embedding, which we call \emph{QK type embeddings}. The timestep embedding for input tokens should be handled specially by feeding $t=0$, which we call \emph{clean tokens}. These techniques improve quality and convergence; they are broadly applicable to any \rgbtox and \xtorgb models.

\begin{figure}[t]
    \footnotesize
    \begin{minipage}{1\linewidth}
        \begin{overpic}[width=0.3279\linewidth]{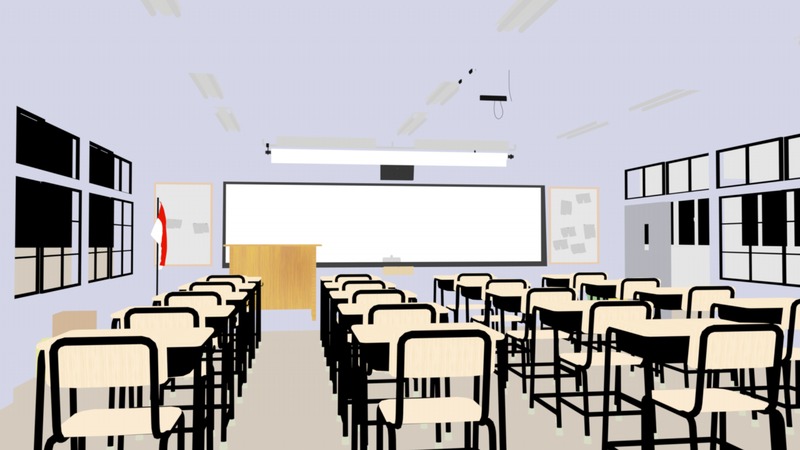}
        \put(27,59){Input Albedo}
        \end{overpic}\hfill
        \begin{overpic}[width=0.3279\linewidth]{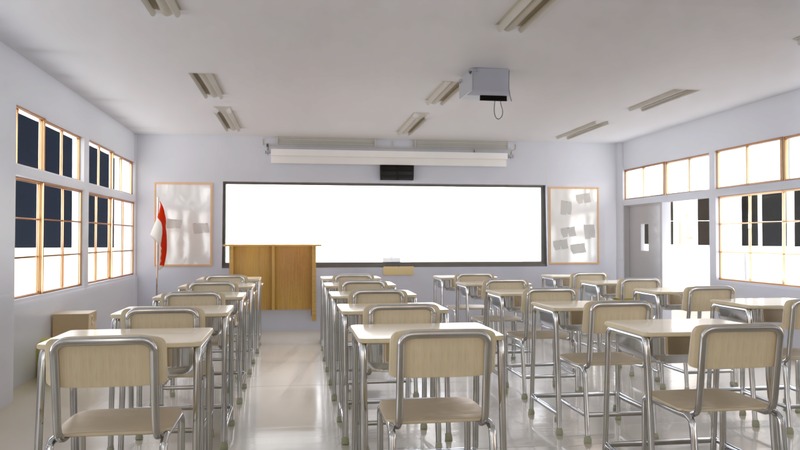}
        \put(5,59){Trained on synthetic data}
        \end{overpic}\hfill
        \begin{overpic}[width=0.3279\linewidth]{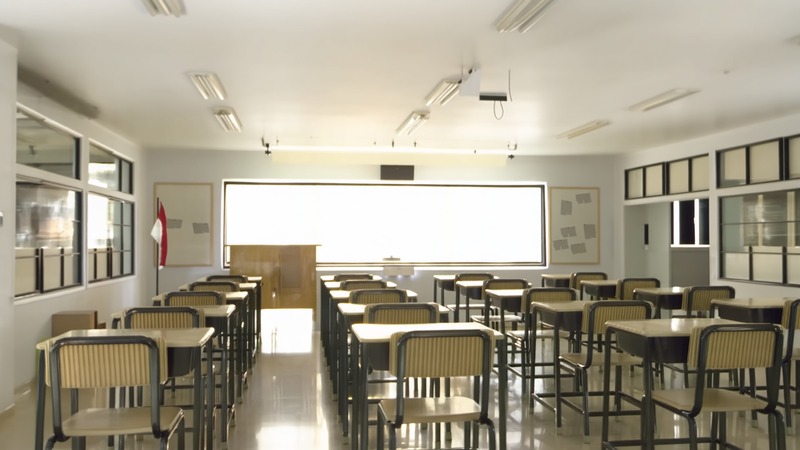}
        \put(9,59){Trained on real videos}
        \end{overpic}
    \end{minipage}
    \caption{Training \xtorgb on synthetic data leads to the model quickly adopting the synthetic look; real data with annotated G-buffers is key to realistic \xtorgb results. }
    \label{fig:x2rgb_cgdata}
\end{figure}
\begin{figure}[t]
    \footnotesize
    \begin{minipage}{1\linewidth}
        \begin{overpic}[width=0.33\linewidth]{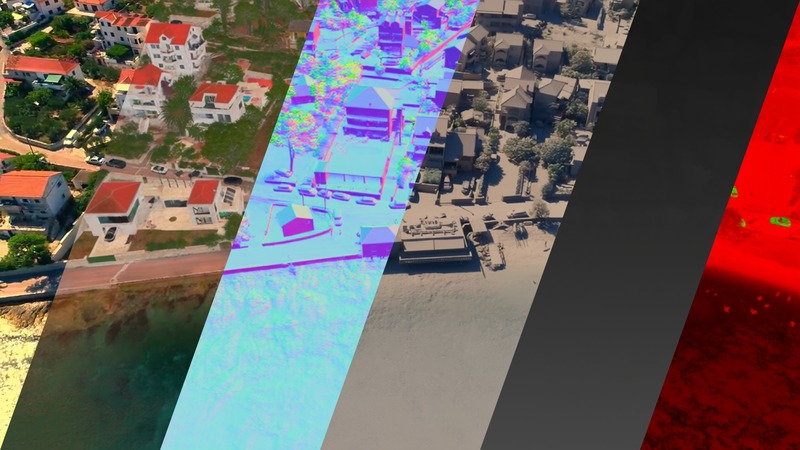}
        \put(8,-10){{Real video \& G-buffers}}
        \end{overpic}\hfill
        \begin{overpic}[width=0.66\linewidth]{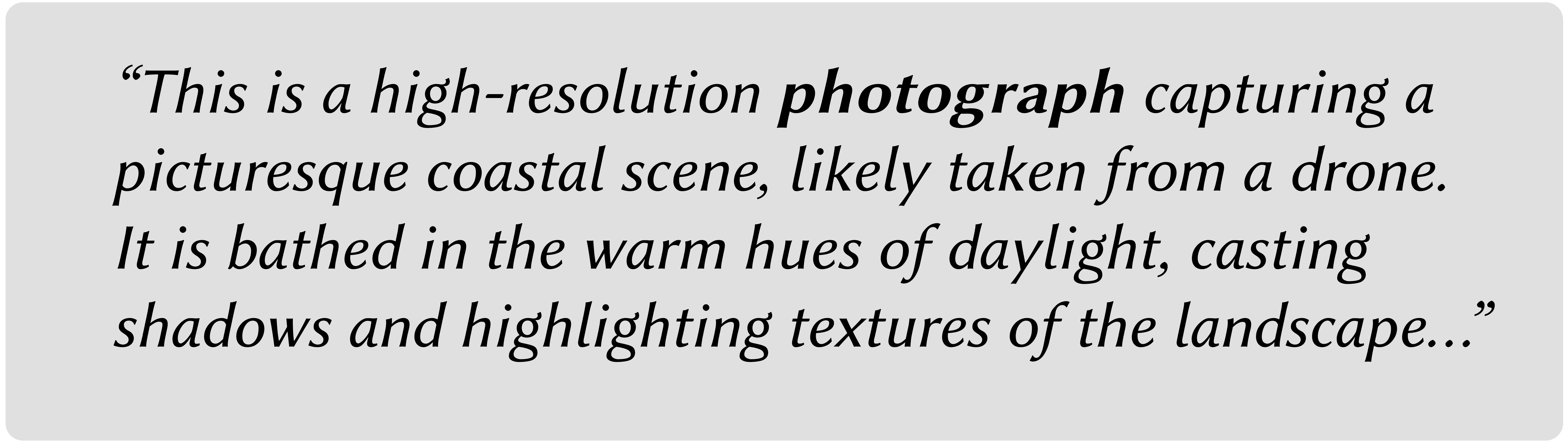}
        \put(28,-5){{The corresponding caption}}
        \end{overpic}\hfill
    \end{minipage}
    \vspace{0.8em}
    \caption{\edit{Example from our \textsf{(RGB, X)} dataset used to train our \xtorgb models: a real video with estimated G-buffers, together with the VLM-produced caption.}}
    \label{fig:x2rgb_data}
\end{figure}

\section{\xtorgb models for realistic generative rendering}
\label{sec:x2rgb}

\begin{figure}[tb]
    \centering
    \includegraphics[width=0.6\linewidth]{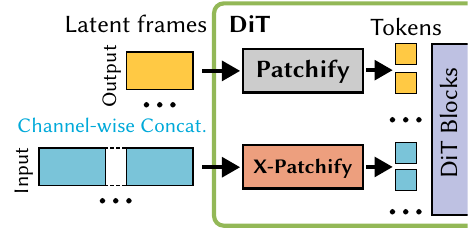}
    \caption{Our \emph{X-patchify} technique uses channel concatenation for a flexible number of input modalities, learning a patchify operator to map all of them into the same tokens, while still keeping outputs as separate tokens.}
    \vspace{-1em}
    \label{fig:x_patchify}
\end{figure}

In generative forward rendering (\xtorgb), our goal is to take one or more G-buffers as inputs, and produce realistic RGB frames as output, generating all information that is not included in the inputs.

\edit{To build such models, we first use our \rgbtox model to prepare a dataset of real videos paired with estimated G-buffers (\cref{sec:dataset}). We then train an \xtorgb model that reuses the conditioning mechanisms introduced in previous sections, while adding a compact way to support multiple G-buffer inputs (\cref{sec:xpatchify}). Finally, we describe inference-time controls for adjusting how strongly the generated RGB sequence follows the provided scene specification, including condition dropout and blurring (\cref{sec:dropout-blurring}) and lighting-related conditioning choices (\cref{sec:lighting}).}

\subsection{\textsf{(RGB, X)} Dataset}
\label{sec:dataset}

In contrast to \rgbtox, it is not sufficient to train \xtorgb on synthetic pairs of G-buffers and RGB images, as the results quickly shift towards a synthetic look; see \cref{fig:x2rgb_cgdata}. Inspired by \citet{zeng2024rgbx}, we use our \rgbtox model to annotate real data with G-buffers, and train the \xtorgb model on the resulting pairs. \edit{We refer to this collection of real RGB videos and corresponding estimated G-buffers as our \textsf{(RGB, X)} dataset.} \edit{We use about 6,622 videos from the free Pexels service, holding out 371 videos for test. For each real video, we sample clips and annotate them with our \rgbtox model to obtain aligned estimated G-buffers.}

Furthermore, we caption the videos with Qwen2.5-VL \cite{qwen2.5-VL}. An example is shown in \cref{fig:x2rgb_data}. Real footage captions will generally contain "photograph, real video" etc., while rendered sequences (some of which are included in Pexels) will contain "video game, rendered" etc. in the captions. \edit{This enables style supervision while preserving the pretrained text prior, so later classifier-free guidance~\cite{ho2022classifier} can use the caption space to steer between realistic and synthetic looks.}

\subsection[\xtorgb model with \xx-patchify]{\edit{\xtorgb model with \xx-patchify}}
\label{sec:xpatchify}

% Similar to the \rgbtox models
\edit{Following the frame-wise formulation in \cref{sec:dit-repurposing} and the improvements in \cref{sec:improved-rgbx}, we provide the conditioning G-buffer frames to the model with the output RGB frames, and we use the QK type embedding and clean token techniques.}
In contrast to \rgbtox, we can now have multiple input signals (e.g. albedo and depth). We introduce the \emph{\xx-patchify} module, which takes multiple input signals and packs them into the same set of input tokens; see \cref{fig:x_patchify}. This is essentially using channel-wise concatenation for all G-buffer inputs to the model.
\edit{\xx-patchify first stacks the selected clean G-buffer latents along the channel dimension, then maps each resulting patch to a single conditioning token:}
\begin{equation}
\edit{\begin{aligned}
\mathbf{x}_{0}^{\xx}
&=
\operatorname*{concat}_{\mathrm{ch}}\!\left(
\{\hat{\mathbf{x}}^{m}_{0}\}_{m\in\mathcal{M}_{\mathrm{in}}}
\right),\\
\mathrm{shape}(\mathbf{x}_{0}^{\xx})
&=
H'\times W'\times C_{\mathrm{cat}}\times F',
\quad C_{\mathrm{cat}}=|\mathcal{M}_{\mathrm{in}}|C,\\
\mathbf{u}_{\ell}^{\xx,\mathrm{in}}
&=
\mathcal{P}_{\xx}\!\left(\mathbf{x}_{0,\ell}^{\xx}\right).
\end{aligned}}
\label{eq:x-patchify}
\end{equation}
\edit{The operator $\operatorname*{concat}_{\mathrm{ch}}$ denotes channel-wise concatenation, and $C_{\mathrm{cat}}$ is the channel count after stacking the selected $C$-channel G-buffer latents. The learned operator $\mathcal{P}_{\xx}$ then patchifies this tensor into conditioning tokens. \xx-patchify is applied only to clean input signals; noisy RGB output tokens remain separate, as in \cref{eq:framewise-concat}.}
We find that this preserves quality and training convergence, unlike mixing inputs and outputs in the same set of tokens.

\edit{During inference, we use classifier-free guidance (CFG)~\cite{ho2022classifier} to bias \xtorgb samples toward real imagery. We define a positive prompt $p^{+}$ that describes the desired realistic imagery, such as ``photograph, real video, high-resolution, high-quality'', and a negative prompt $p^{-}$ that describes the appearance we want to suppress, such as ``synthetic, CG, rendered, video game, unrealistic''. CFG then pushes the predicted velocity from the negative-prompt prediction toward the positive-prompt prediction.}
\edit{We denote the \xtorgb latent collection by $\mathbf{x}_{\mathrm{X}\rightarrow\mathrm{RGB}}(t)$; it contains the clean G-buffer conditions and the noisy RGB latent being denoised:}
\begin{equation}
\edit{\begin{aligned}
\mathbf{x}_{\mathrm{X}\rightarrow\mathrm{RGB}}(t)
&=
\left(
\{\hat{\mathbf{x}}_{0}^{m}\}_{m\in\mathcal{M}_{\mathrm{in}}},
\mathbf{x}_{t}^{\image}
\right),\\
\mathbf{v}^{-} &= \model(\mathbf{x}_{\mathrm{X}\rightarrow\mathrm{RGB}}(t),t,p^{-})^{\image},\\
\mathbf{v}^{+} &= \model(\mathbf{x}_{\mathrm{X}\rightarrow\mathrm{RGB}}(t),t,p^{+})^{\image},\\
\mathbf{v}_{\mathrm{CFG}} &= \mathbf{v}^{-} + w\left(\mathbf{v}^{+}-\mathbf{v}^{-}\right).
\end{aligned}}
\label{eq:x2rgb-cfg}
\end{equation}
\edit{The hatted latents $\hat{\mathbf{x}}_{0}^{m}$ are clean conditioning G-buffer latents, and $\mathbf{x}_{t}^{\image}$ is the noisy RGB latent. The velocities $\mathbf{v}^{+}$ and $\mathbf{v}^{-}$ are conditioned on the positive and negative prompts, and $w$ is the classifier-free guidance weight. The guidance direction preserves the G-buffer condition while biasing the sample away from a synthetic look and toward realism.}

\subsection[Condition dropout and blurring]{\NoCaseChange{\edit{Conditions dropout and blurring}}}
\label{sec:dropout-blurring}

\begin{figure}[tb]
    \centering
    \includegraphics[width=0.75\linewidth]{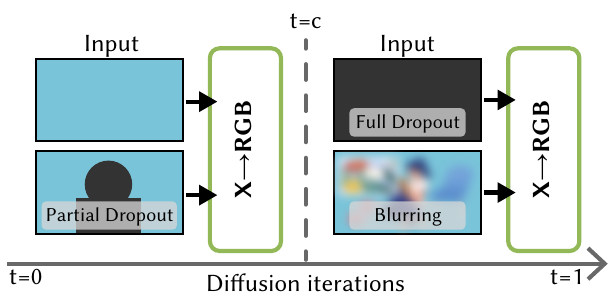}
    \caption{Our models can support full dropout of a G-buffer, partial dropout (where black means no guidance), and blurring. These techniques can be applied on a fraction of diffusion iterations, allowing for control in guidance strength.}
    \vspace{-1em}
    \label{fig:dropout}
\end{figure}

An important goal of generative rendering is to permit
specifying only part of the scene, and let the model creatively fill in the rest.
For this reason, we allow some input G-buffers to be blurred, specified only partially, or completely dropped during the diffusion iterations; see \cref{fig:dropout}.
\edit{Before patchification, these choices transform each clean condition $\hat{\mathbf{x}}^{m}$ into the condition actually given to the model at sampling step $n$. Let $B_{\sigma}$ denote Gaussian blur, $\mathbf{M}^{m}$ a spatial mask whose zero entries mean "no guidance" at those pixels, and $\delta_{m,n}\in\{0,1\}$ a gate indicating whether modality $m$ is active at step $n$. If $n$ indexes the $N$ sampling steps from coarse to fine, a condition used only for the first fraction $c$ of sampling has}
\begin{equation}
\edit{
\delta_{m,n} =
\begin{cases}
1, & n \le cN,\\
0, & n > cN,
\end{cases}
\qquad
\tilde{\mathbf{x}}^{m}_{(n)} =
\delta_{m,n}\,\mathbf{M}^{m}\odot B_{\sigma_m}\!\left(\hat{\mathbf{x}}^{m}\right).}
\label{eq:iteration-dropout}
\end{equation}
\edit{Here $\tilde{\mathbf{x}}^{m}_{(n)}$ is the transformed condition at step $n$, and $\odot$ denotes pointwise multiplication. The full G-buffer is removed whenever $\delta_{m,n}=0$; spatial regions are removed where $\mathbf{M}^{m}=0$; nonzero $\sigma_m$ preserves the condition only at a lower spatial frequency. Setting $c=1$ keeps the condition active throughout sampling, while smaller $c$ applies conditioning during early coarse denoising and removes it during later detail synthesis, reducing guidance strength without changing the trained model.}

\edit{To enable these controls at inference time, we expose the model to the same kinds of missing or weakened conditions during training. For training examples that use controllable G-buffer inputs, we randomly transform the clean conditions before patchification: an entire modality may be dropped, spatial regions may be masked out to indicate no guidance, or the condition may be blurred with a randomly sampled blur amount.} \edit{For blurring, we sample $\sigma$ from $[\sigma_{\min}, \sigma_{\max}]$, with $\sigma_{\min}=1$ and $\sigma_{\max}=10$. For dropout, we remove an entire modality with probability 50\%, or mask out selected segment regions; in both cases, removal is implemented by setting the corresponding latent values to zero before patchification.} \edit{The flow-matching loss is still computed only on the RGB output tokens; these transformations only change the clean conditioning tokens. This teaches the model that absent, partially specified, or blurred G-buffers should be interpreted as weaker guidance rather than as literal black or blurred scene content.}

\subsection[Lighting conditions]{\NoCaseChange{\edit{Lighting conditions}}}
\label{sec:lighting}

In many cases, we would like to control the lighting instead of having it randomly sampled. One option to achieve this is by providing a \emph{diffuse irradiance} $\irra$~\cite{zeng2024rgbx}. Computing diffuse irradiance requires path-tracing the scene, but this is a much simpler path-tracing problem than full rendering, as it does not require complex materials, and the Lambertian shading converges and denoises very easily.

\edit{Beyond irradiance, many other lighting representations are possible, including global IBL lighting \cite{liang2025diffusionrenderer} and image-space grids of local illumination probes \cite{Li2022relighting}. Both of these could be added to our models given appropriate training data. We show a new alternative lighting representation, \emph{albedo-free direct lighting} \direct. Assuming a simple shading model combining one diffuse and one glossy lobe, we compute direct illumination from local as well as environment lights, and avoid multiplication by the diffuse and specular albedos when shading. This representation has the advantage of being even easier to compute than diffuse irradiance, since it does not contain multi-bounce lighting, while preserving the property of being orthogonal to the albedo buffer.}

\edit{Both lighting conditions can be overly constraining when supplied at full resolution throughout the entire denoising process. We therefore use the conditioning dropout and blurring mechanisms from \cref{sec:dropout-blurring} to control the strength of lighting guidance: a lighting buffer can be blurred using a Gaussian with a standard deviation of $\sigma$ pixels, and it can be supplied only for the first fraction $c \in [0,1]$ of diffusion iterations.}

% \paragraph{Summary.} Lighting in \xtorgb models can be controlled in multiple ways, including by a diffuse irradiance buffer or an albedo-free direct lighting buffer. Providing overly detailed lighting buffers can contrain the creativity and realism; decreasing the guidance amount can be achieved through a combination of blurring the lighting buffer, and dropping the lighting guidance for some fraction of late diffusion iterations.

% \zheng{
% \begin{itemize}
%     \item One of our goal in generative rendering (X->RGB) is to support cases where input G-buufers can be partially specified or completely dropped out.
%     \item (why channel dropped out and blurring should be motivated in overview)
%     \item Dropout in spatial domain (pixel maskout) and frequency domain (blurring), dropout in denoising time (during denoising process)
%     \item To achieve that, inspired by RGBX, we randomly drop condition signals during training
%     \item Lighting: blurred irradiance; albedo-free direct lighting instead of irradiance (Marco's experiments)
% \end{itemize}
% }

\section{Streaming models}
\label{sec:streaming}

\begin{figure}[t]
    \centering
    \includegraphics[width=0.9\linewidth]{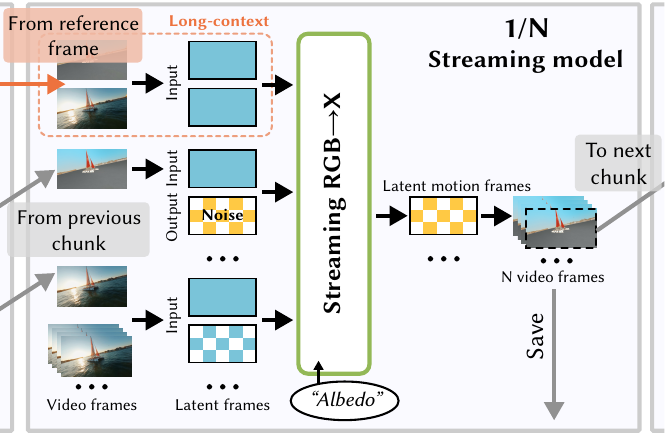}
    \caption{\edit{The overall design of our $1/N$ streaming models. For each chunk, the model receives $N$ current input frames and one stitching frame from the previous chunk, whose input and output are already known, then produces the $N$ new output frames through a diffusion process. Optional long-context tokens provide a clean reference frame for long-term memory. For simplicity, the figure illustrates the \rgbtox case with albedo as \xx.}}
    \label{fig:streaming}
\end{figure}

\edit{The models described so far operate on fixed-length video clips. This length can be increased with more training time and GPU memory, but it remains bounded by the finite context of the base text-to-video DiT model.}
\edit{To be truly useful for generative forward and inverse rendering, the models need to support an unbounded number of frames handled in a \emph{streaming} manner.}

% unless frame resolution is lowered or a stronger compression VAE is used, both of which affect quality.

\paragraph{Teacher forcing} \edit{A straightforward idea to achieve a streaming model is to overlap one frame between chunks. We call this overlapped frame the \emph{stitching frame}. We condition the model on the stitching frame, for which both input and output are known, and generate outputs for $N$ new frames, for which only inputs are provided; we term such a model a $1/N$-streaming model.} \edit{Let $s$ index streaming chunks. Using the sampler notation from \cref{sec:dit-repurposing}, the generated chunk is}
\begin{equation}
\edit{
\hat{\mathbf{x}}^{\mathrm{out}}_{s,1:N}
=
\operatorname{Sample}_{\theta}\!\left(
\mathbf{x}^{\mathrm{in}}_{s,1:N},
\hat{\mathbf{x}}^{\mathrm{out}}_{s-1,N},
p
\right).}
\label{eq:streaming-recursion}
\end{equation}
\edit{The hat marks generated outputs, $\mathbf{x}^{\mathrm{in}}_{s,1:N}$ is the clean input sequence for chunk $s$, $\hat{\mathbf{x}}^{\mathrm{out}}_{s-1,N}$ is the generated stitching frame from the previous chunk, and $p$ is the text prompt used by the diffusion sampler.}

\edit{This simple approach can work for short sequences (100--200 frames), but accumulates error over longer sequences. The practice of training on ground truth while relying on past predictions at inference time is known as \emph{teacher forcing}. As noted by \citet{huang2025selfforcing}, this leads to drift and error accumulation.}
\edit{With $\mathbf{x}^{\mathrm{out},*}$ denoting ground truth, the stitching frame differs between training and inference:}
\begin{equation}
\edit{
\mathbf{x}^{\mathrm{prev}}_{\mathrm{train}} =
\mathbf{x}^{\mathrm{out},*}_{s-1,N},
\qquad
\mathbf{x}^{\mathrm{prev}}_{\mathrm{infer}} =
\hat{\mathbf{x}}^{\mathrm{out}}_{s-1,N}.}
\label{eq:teacher-forcing-mismatch}
\end{equation}
\edit{The star in $\mathbf{x}^{\mathrm{out},*}$ denotes ground truth. Prediction errors can therefore be fed back as the next chunk's stitching frame.}

\paragraph{Augmentation} \edit{To break the ``perfection'' of teacher forcing, we perturb that one stitching frame with: (1) per-channel brightness shifts up to $\pm 2\%$, (2) Gaussian blur or unsharp masking ($\sigma$ up to 1.5), and (3) a VAE round-trip simulating the lossy decode-encode cycle required at inference to extract the stitching frame from decoded outputs.}
\edit{We write the augmented stitching frame as}
\begin{equation}
\edit{
\tilde{\mathbf{x}}^{\mathrm{prev}} =
\mathcal{A}_{\lambda}\!\left(\mathbf{x}^{\mathrm{out},*}_{s-1,N}\right),
\qquad
\lambda\in[0,1].}
\label{eq:streaming-augmentation}
\end{equation}
\edit{The operator $\mathcal{A}_{\lambda}$ applies the brightness, blur or sharpening, and VAE round-trip perturbations, while $\lambda$ controls the augmentation strength.}
While this slightly reduces drifting, it does not resolve it.

\paragraph{Long context tokens.} \edit{Another important issue in streaming designs is lack of long-term memory: even if results are locally temporally stable and do not accumulate error, conditioning only on the stitching frame can result in forgetting details of earlier frames. This can produce issues like temporally shifted albedo in \rgbtox models, or slowly changing lighting in \xtorgb models. Our solution, inspired by FramePack \cite{zhang2025framepack}, is to provide \emph{long-context tokens} to the model. For simplicity, we provide a single reference frame, though multiple reference frames could be supported by the same approach. During training, this is the first frame of the dataset sequence, though during inference this is not required. This reference frame should be representative of overall colors, lighting, etc.}
\edit{Long context appends reference-frame tokens to the per-chunk tokens:}
\begin{equation}
\edit{
\mathcal{U}_{s}^{\mathrm{long}}(t) =
\mathcal{U}_{s}^{\mathrm{chunk}}(t)
\cup
\left\{\mathbf{u}_{\ell}^{m,\mathrm{ref}} : m\in\mathcal{M}_{\mathrm{in}}\cup\mathcal{M}_{\mathrm{out}}\right\}.}
\label{eq:long-context-tokens}
\end{equation}
\edit{The set $\mathcal{U}_{s}^{\mathrm{chunk}}(t)$ contains the regular per-chunk tokens, while $\mathbf{u}_{\ell}^{m,\mathrm{ref}}$ denotes a clean reference-frame token. The reference tokens provide stable sequence-level color, lighting, and material appearance, even when the current chunk is far from the beginning of the video.}
This long-context feature reduces drifting and error accumulation.

\paragraph{Self forcing} \edit{Inspired by \citet{huang2025selfforcing}, we propose to train the model on its own predictions. Specifically, we train the model to predict the flow velocity from its own partially denoised latent as well as the generated stitching frame from the previous chunk. To achieve that, we run inference from pure noise during training time, and add noise to this prediction instead of to the ground truth (and keep the prediction around for the next chunk's training step).}
\edit{Thus the noisy training latent is constructed from the model's sampled output rather than from ground truth:}
\begin{equation}
\edit{\begin{aligned}
\hat{\mathbf{x}}^{\mathrm{out}}_{s,0}
&=
\operatorname{Sample}_{\theta}\!\left(
\mathbf{x}^{\mathrm{in}}_{s,1:N},
\hat{\mathbf{x}}^{\mathrm{out}}_{s-1,N},
p
\right),\\
\hat{\mathbf{x}}^{\mathrm{out}}_{s,t}
&=
(1-t)\hat{\mathbf{x}}^{\mathrm{out}}_{s,0}
+ t\boldsymbol{\epsilon}.
\end{aligned}}
\label{eq:self-forcing}
\end{equation}
\edit{The latent $\hat{\mathbf{x}}^{\mathrm{out}}_{s,0}$ is a sampled output for chunk $s$, and $\hat{\mathbf{x}}^{\mathrm{out}}_{s,t}$ is its noisy version used for training. The model is then trained on latents closer to those it will produce and reuse during streaming inference.}

\edit{We find that self forcing by itself is not immediately better than teacher forcing: it can struggle to converge and may replace drift with flickering between chunks. A hybrid strategy works better: first train a teacher forcing model with the above augmentations, then continue from that checkpoint with an interleaved teacher forcing and self forcing stage. In this second stage, each training example uses teacher forcing or self forcing with equal probability, and we use a 10$\times$ smaller learning rate to stabilize training. This strategy helps because the model needs to be already reasonably good to produce valid samples during self forcing training. Note that unlike \citet{huang2025selfforcing}, we do not currently use DMD distillation~\cite{yin2024dmd,yin2024dmd2} or adversarial losses. Another difference is that their model is an $L/1$ model, where $L$ is up to 81; that is, conditioning on all previous frames in an 81-frame generation.}

\section{Unified model}
\label{sec:unified}

\begin{figure}[t]
    \begin{overpic}[width=1\linewidth]{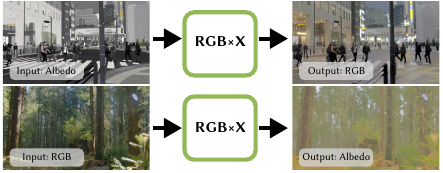}
    \end{overpic}
    \caption{Our proof-of-concept unified model, RGB$\times$X, supports forward and inverse rendering within the same model weights. The input and output tokens are marked using embeddings, just like modalities, letting the model infer the desired task.}
    \label{fig:unified}
\end{figure}

The techniques introduced in previous sections (frame-wise concatenation, QK type embedding, and clean tokens) generalize to a single unified \edit{\rgbxx} model, that is, a single model capable of generative rendering and inverse rendering. We simply concatenate all input sequences and all desired output sequences as frames, mark which is which through embeddings, and train the model with the desired combinations.
\edit{For each training example, choosing $\mathcal{M}_{\mathrm{in}}$ and $\mathcal{M}_{\mathrm{out}}$ specifies both the latent collection and the internal token layout:}
\begin{equation}
\edit{\begin{aligned}
\mathbf{x}_{\mathrm{uni}}(t;\mathcal{M}_{\mathrm{in}},\mathcal{M}_{\mathrm{out}})
&=
\left(
\{\mathbf{x}_{0}^{m}\}_{m\in\mathcal{M}_{\mathrm{in}}},
\{\mathbf{x}_{t}^{m}\}_{m\in\mathcal{M}_{\mathrm{out}}}
\right),\\
\mathcal{U}_{\mathrm{uni}}(t;\mathcal{M}_{\mathrm{in}},\mathcal{M}_{\mathrm{out}})
&=
\left\{\mathbf{u}_{\ell}^{m,\mathrm{in}} : m\in\mathcal{M}_{\mathrm{in}}\right\}
\cup
\left\{\mathbf{u}_{\ell,t}^{m,\mathrm{out}} : m\in\mathcal{M}_{\mathrm{out}}\right\}.
\end{aligned}}
\label{eq:unified-token-set}
\end{equation}
\edit{The collection $\mathbf{x}_{\mathrm{uni}}$ is the external latent input to the model, while $\mathcal{U}_{\mathrm{uni}}$ is the internal token layout after patchification. A task is therefore a choice of which modalities are clean conditions and which modalities are noisy outputs to denoise.}
\edit{As in the frame-wise model, training ignores input tokens and evaluates flow matching only on output modalities:}
\begin{equation}
\edit{
\mathcal{L}_{\mathrm{uni}} =
\mathbb{E}\!\left[
\sum_{m\in\mathcal{M}_{\mathrm{out}}}
\left\|\model(\mathbf{x}_{\mathrm{uni}}(t),t,p)^{m}
-(\boldsymbol{\epsilon}^{m}-\mathbf{x}^{m}_{0})\right\|_2^2
\right].}
\label{eq:unified-loss}
\end{equation}
\edit{The loss $\mathcal{L}_{\mathrm{uni}}$ is the output-only flow-matching loss for the unified model, and $\mathbf{x}_{\mathrm{uni}}(t)$ denotes the latent collection for the sampled task. The type embeddings identify role and modality, allowing the same weights to express \rgbtox, \xtorgb, or multi-input/multi-output configurations.}

One difference is that we can no longer use the \xx-patchify module introduced in \cref{sec:xpatchify} to patchify multiple output signals into one single set of tokens; when these tokens are noisy tokens, the model struggles to converge. Therefore, we need a separate set of tokens per modality. As shown in \cref{fig:unified}, our unified \rgbxx model can achieve both forward and inverse rendering, and support multiple inputs and outputs. In summary, techniques introduced in previous sections also generalize to a unified model that supports inverse and forward generative rendering, though we leave a full exploration of model unification for future work.

\section{Results}
\subsection{Implementation details}
\edit{The base model used in our experiments is the Wan 2.1 14B video model \cite{wan}. Unless stated otherwise, we finetune all DiT parameters of Wan 2.1 while keeping the VAE frozen. We use AdamW with a constant learning-rate schedule, learning rate $10^{-5}$ for image/video finetuning and $10^{-6}$ for streaming finetuning. Training uses a global batch size of 128 across 32 A100 80GB GPUs. All reported inference timings are measured on a single A100 80GB GPU.}

\paragraph{\rgbtox image and video models.} \edit{We train our \rgbtox models on the internal synthetic dataset described in \cref{sec:improved-rgbx}. We use a two-step training strategy. We first finetune the base model on still images rendered at $1920\times1088$ for 200 steps; each image step takes 70 seconds. We then finetune the resulting image model on 17-frame videos at $1280\times720$ for 500 steps; each video step takes 200 seconds.}

\paragraph{\xtorgb image and video models.} \edit{We train several \xtorgb models on the \textsf{(RGB, X)} dataset described in \cref{sec:dataset}, each with a different set of conditioning G-buffers. The specialized models shown in \cref{fig:x2rgb_ours} each use one of the following conditioning sets: albedo, normal, depth, albedo with irradiance, or albedo with direct lighting. We also train a generalized \xtorgb model conditioned on albedo, normal, irradiance, and material inputs with modality dropout for comparison with \rgbx~\cite{zeng2024rgbx}. For all \xtorgb training, we downsample real videos to $1280\times720$, sample 17-frame clips, and extract G-buffers using our \rgbtox model. We use a two-step strategy: first finetuning on still frames for 100 steps, then finetuning on 17-frame videos for 200 steps. The generalized model instead uses 100 image steps followed by 700 video steps to improve convergence.}

\paragraph{Streaming models.} \edit{In our experiments, we instantiate the $1/N$ streaming formulation as a $1/16$ streaming model. Both \rgbtox and \xtorgb streaming models are finetuned from their corresponding 17-frame video models. Streaming finetuning has two stages. In the first stage, we train with augmented teacher forcing; each step has a similar cost to the corresponding video-model training step. The augmentation intensity in \cref{eq:streaming-augmentation} ramps linearly over 100 training steps, with 20\% of samples left unaugmented. We train this first stage for 80 steps for \rgbtox and 40 steps for \xtorgb. In the second stage, we train on 113-frame sequences with hybrid teacher forcing and self forcing examples; each training step operates on a 17-frame chunk and takes 1000 seconds. We interleave teacher forcing and self forcing with equal probability, and train this hybrid stage for another 80 steps for \rgbtox and 40 steps for \xtorgb.}

\paragraph{Inference.} \edit{All results are generated using 20 diffusion sampling steps. For 17-frame generation, \rgbtox takes approximately 200 seconds to synthesize a 17-frame video segment with $\mathrm{CFG}=1$, whereas \xtorgb takes approximately 400 seconds with $\mathrm{CFG}=3$. For the streaming variant, the total inference time is approximately the above inference time multiplied by the number of chunks.}

\subsection{\rgbtox results}

\begin{figure*}[t]
    \vspace{1mm}
    \footnotesize
    \centering
    \begin{minipage}{1.0\linewidth}
        \begin{overpic}[width=0.24\linewidth]{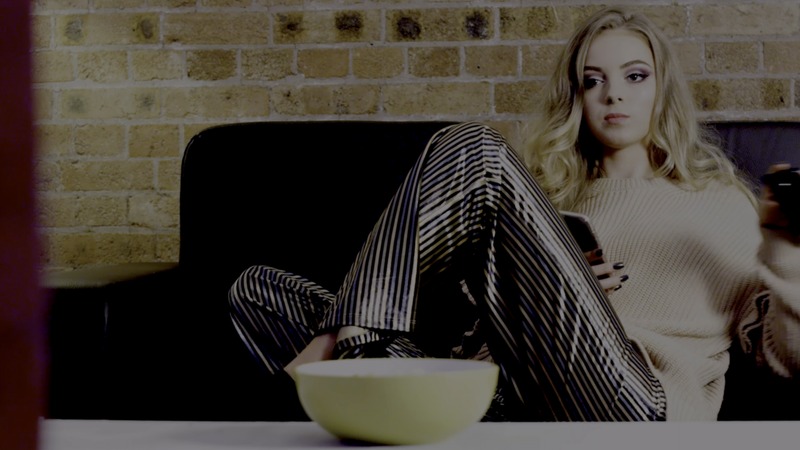}
        \put(33,58){Input RGB}
        \end{overpic}\hspace{1em}
        \begin{overpic}[width=0.24\linewidth]{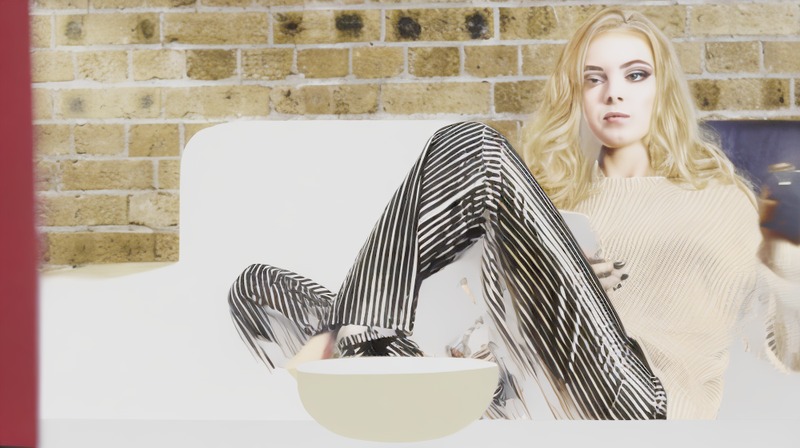}
        \put(19,58){\rgbx \cite{zeng2024rgbx}}
        \put(-7,24){\rotatebox[origin=c]{90}{Albedo}}
        \end{overpic}\hfill
        \begin{overpic}[width=0.24\linewidth]{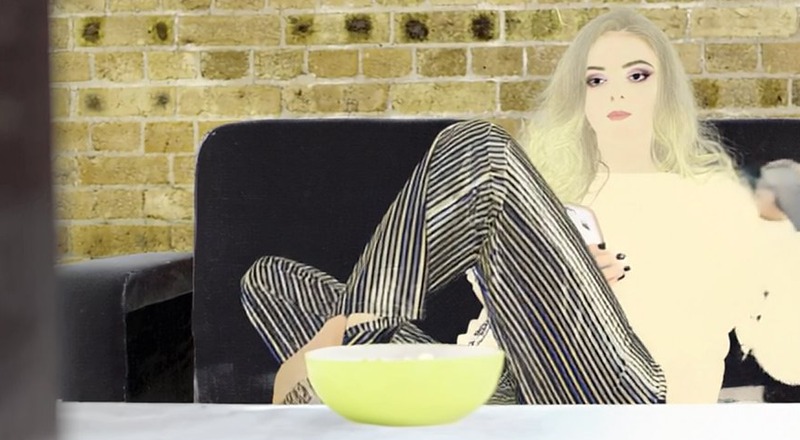}
        \put(28,58){DiffusionRenderer}
        \end{overpic}\hfill
        \begin{overpic}[width=0.24\linewidth]{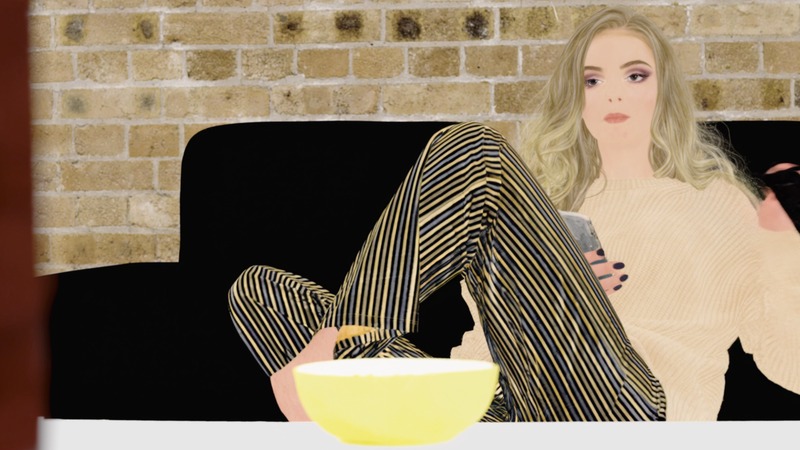}
        \put(40,58){\textbf{Ours}}
        \end{overpic}
    \end{minipage}\par
    \begin{minipage}{1.0\linewidth}
        \begin{overpic}[width=0.24\linewidth]{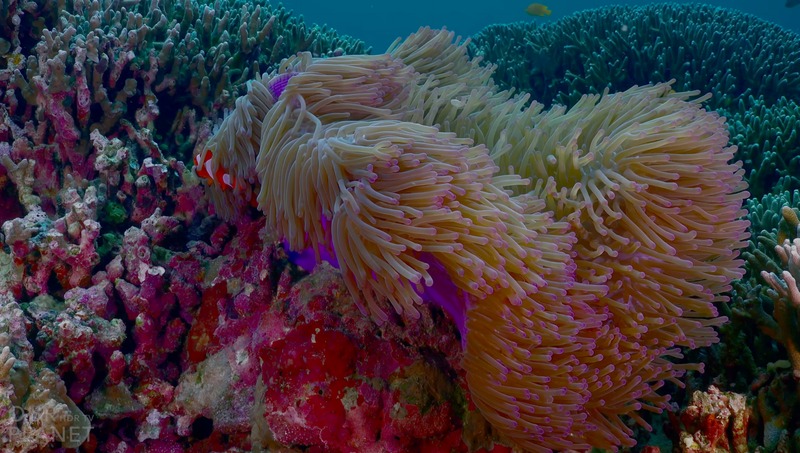}
        \end{overpic}\hspace{1em}
        \begin{overpic}[width=0.24\linewidth]{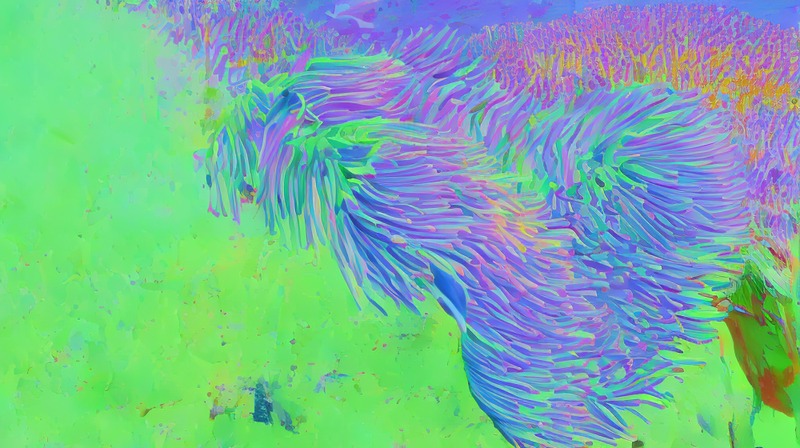}
        \put(-7,24){\rotatebox[origin=c]{90}{Normal}}
        \end{overpic}\hfill
        \begin{overpic}[width=0.24\linewidth]{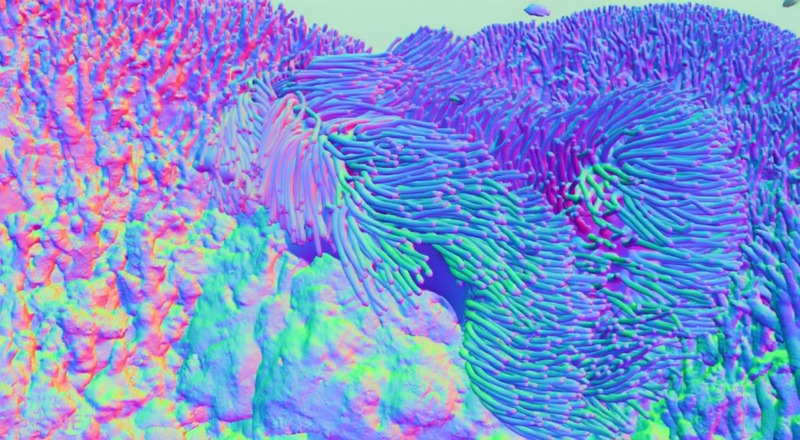}
        \end{overpic}\hfill
        \begin{overpic}[width=0.24\linewidth]{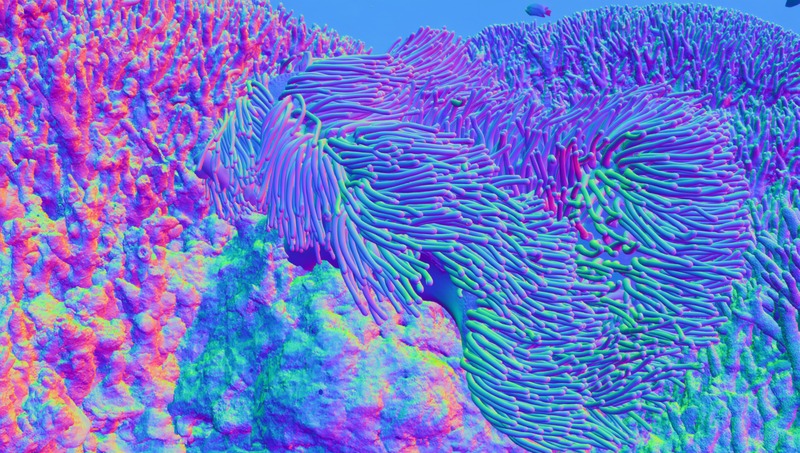}
        \end{overpic}
    \end{minipage}\par
    \begin{minipage}{1.0\linewidth}
        \begin{overpic}[width=0.24\linewidth]{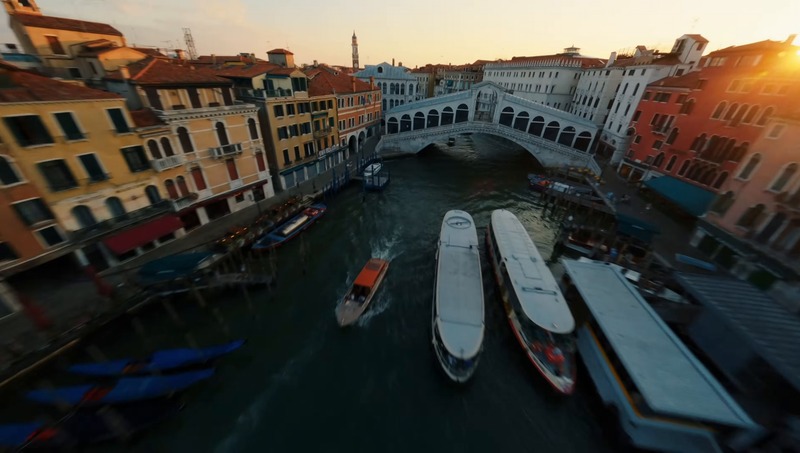}
        \end{overpic}\hspace{1em}
        \begin{overpic}[width=0.24\linewidth]{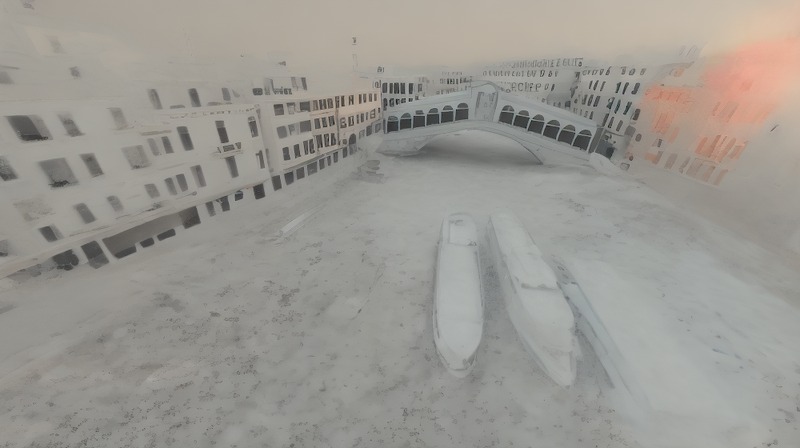}
        \put(-7,23){\rotatebox[origin=c]{90}{Irradiance}}
        \end{overpic}\hfill
        \begin{overpic}[width=0.24\linewidth]{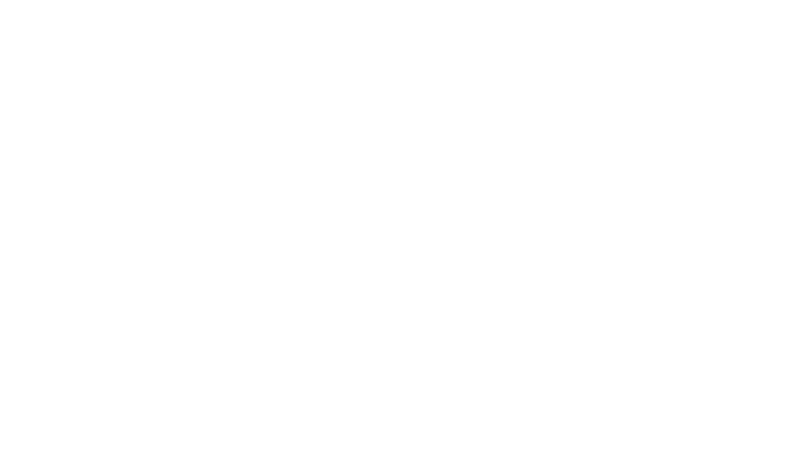}
        \put(42,25){N/A}
        \end{overpic}\hfill
        \begin{overpic}[width=0.24\linewidth]{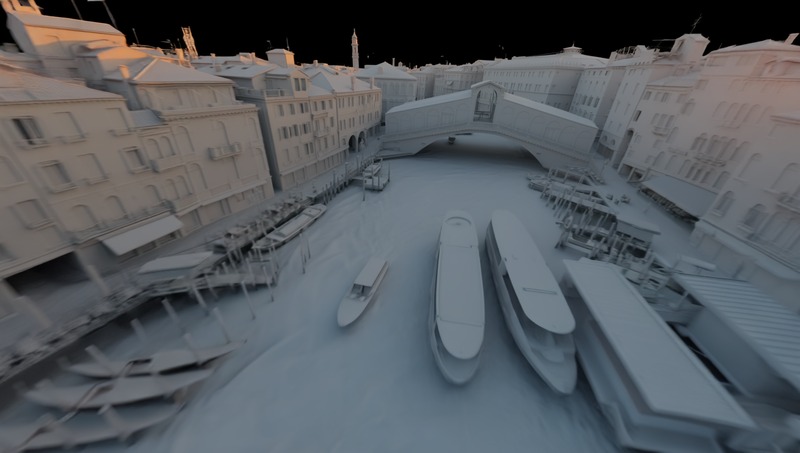}
        \end{overpic}
    \end{minipage}\par
    \begin{minipage}{1.0\linewidth}
        \begin{overpic}[width=0.24\linewidth]{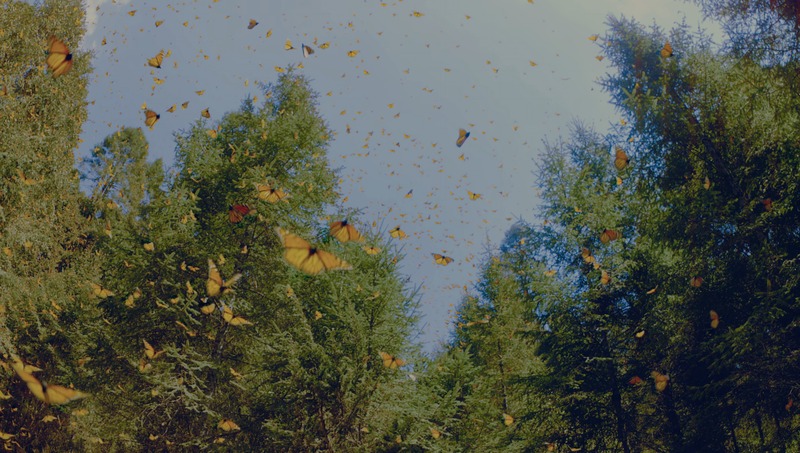}
        \end{overpic}\hspace{1em}
        \begin{overpic}[width=0.24\linewidth]{figures/white_placeholder.jpg}
        \put(42,25){N/A}
        \put(-7,24){\rotatebox[origin=c]{90}{Depth}}
        \end{overpic}\hfill
        \begin{overpic}[width=0.24\linewidth]{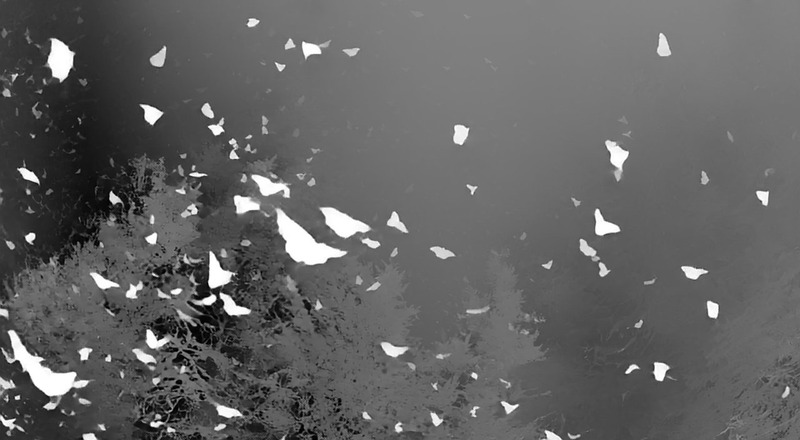}
        \end{overpic}\hfill
        \begin{overpic}[width=0.24\linewidth]{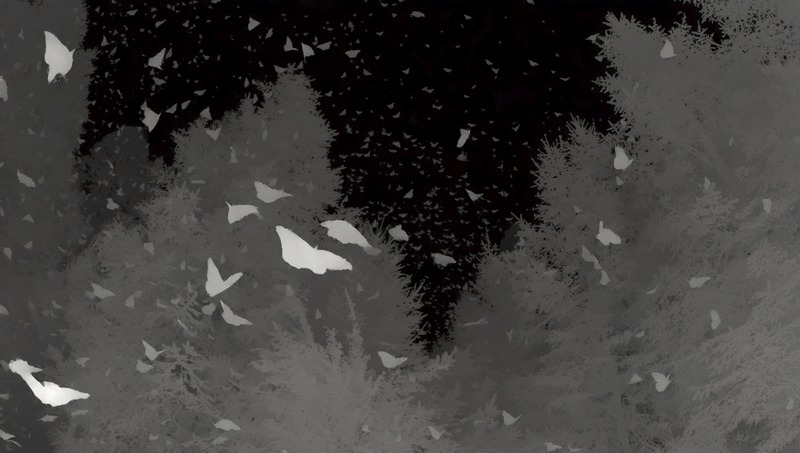}
        \end{overpic}
    \end{minipage}\par
    \begin{minipage}{1.0\linewidth}
        \begin{overpic}[width=0.24\linewidth]{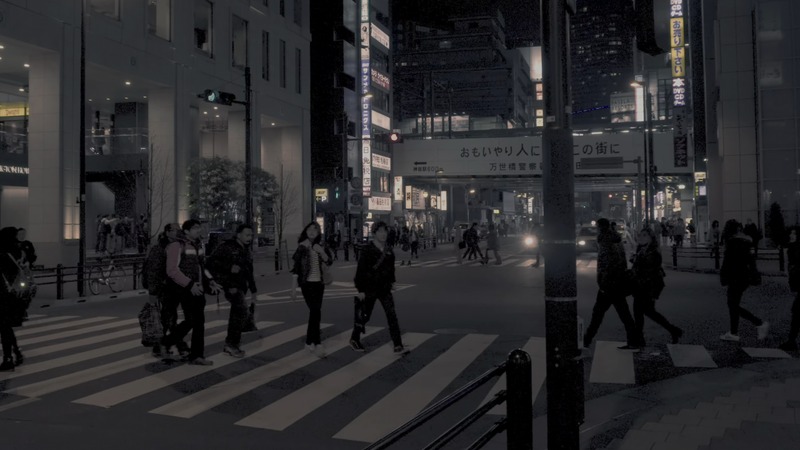}
        \end{overpic}\hspace{1em}
        \begin{overpic}[width=0.24\linewidth]{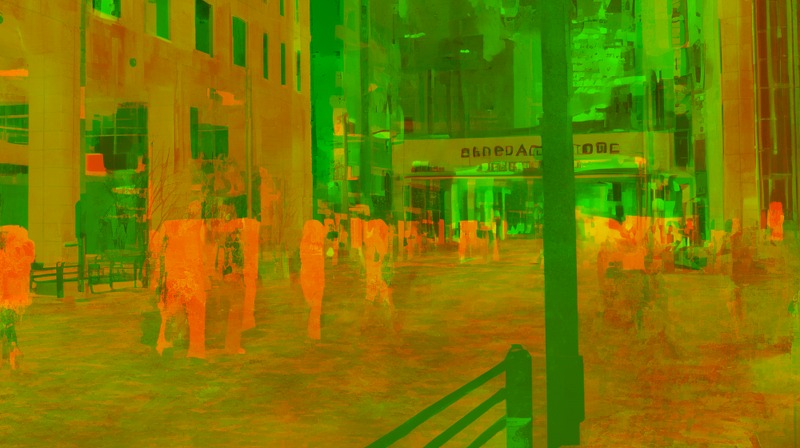}
        \put(-7,24){\rotatebox[origin=c]{90}{Material}}
        \end{overpic}\hfill
        \begin{overpic}[width=0.24\linewidth]{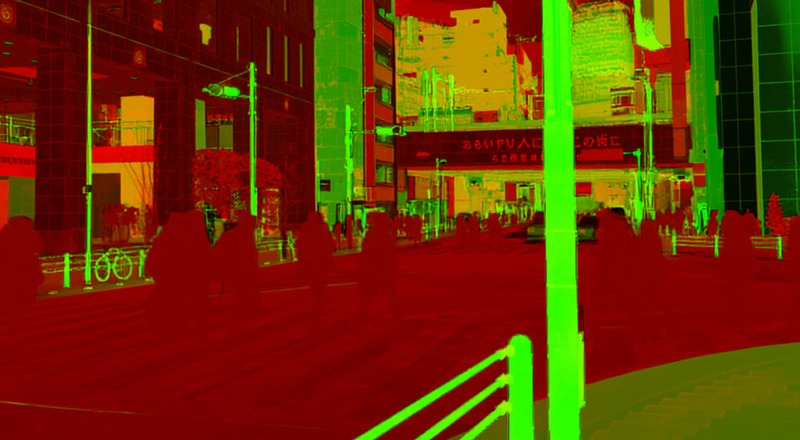}
        \end{overpic}\hfill
        \begin{overpic}[width=0.24\linewidth]{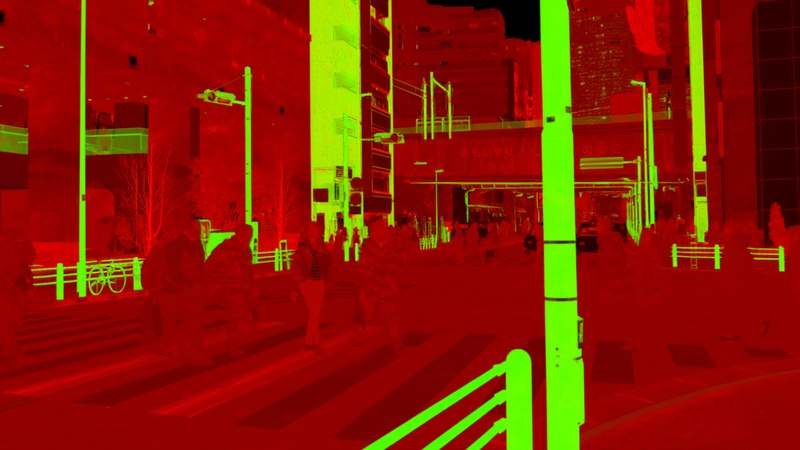}
        \end{overpic}
    \end{minipage}
    \caption{\edit{Qualitative \rgbtox comparison on \textsf{(RGB, X)} test set. Columns show the input RGB, \rgbx~\cite{zeng2024rgbx}, DiffusionRenderer~\cite{liang2025diffusionrenderer} (specifically its newer version based on Cosmos \cite{agarwal2025cosmos}), and ours. Our model produces flatter albedo with less residual shading, supports both depth and irradiance, and avoids the sky-depth failure shown by DiffusionRenderer; DiffusionRenderer produces normals of similar visual quality.}}
    \label{fig:rgb2x_baseline}
\end{figure*}

\begin{figure*}[t]
    \vspace{1mm}
    \footnotesize
    \begin{minipage}{1.0\linewidth}
        \begin{overpic}[width=0.159\linewidth]{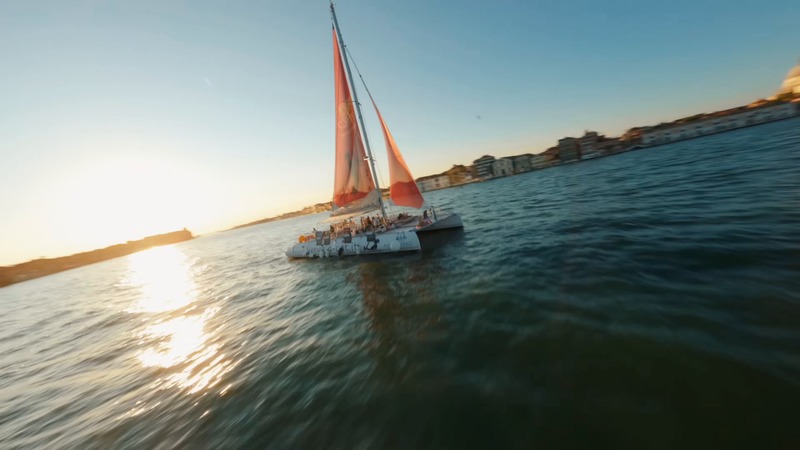}
        \put(30,59){Input RGB}
        \end{overpic}\hspace{1em}
        \begin{overpic}[width=0.159\linewidth]{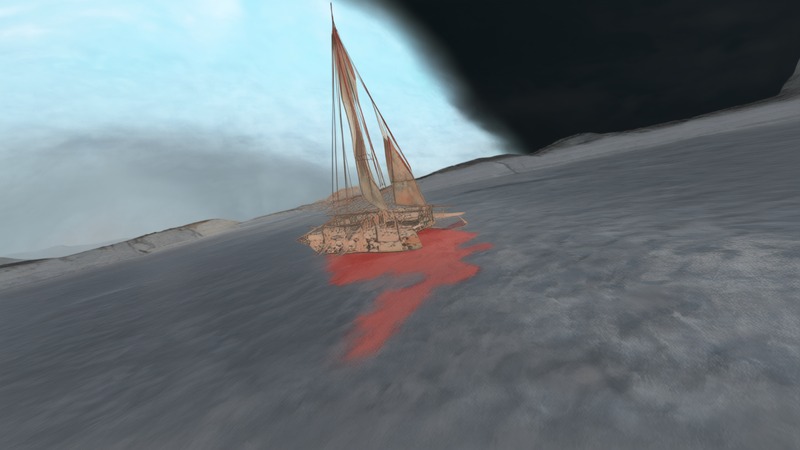}
        \put(-10,24){\rotatebox[origin=c]{90}{Albedo}}
        \put(10,59){Channel-wise Concat.}
        \end{overpic}\hfill
        \begin{overpic}[width=0.159\linewidth]{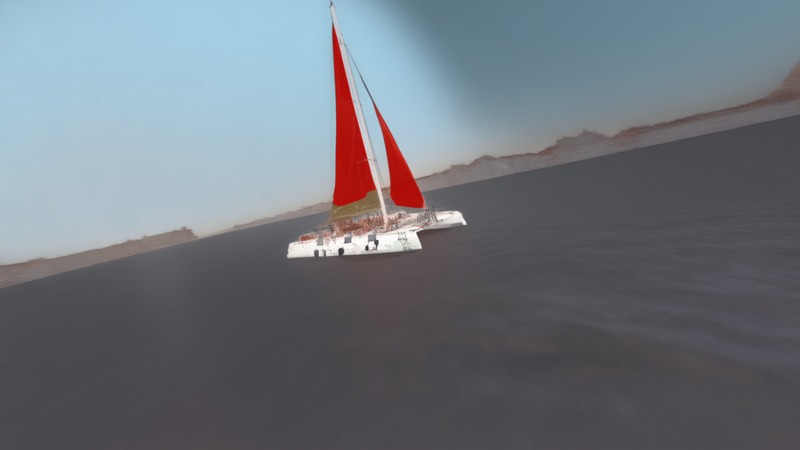}
        \put(10,59){VACE~\cite{jiang2025vace}}
        \end{overpic}\hfill
        \begin{overpic}[width=0.159\linewidth]{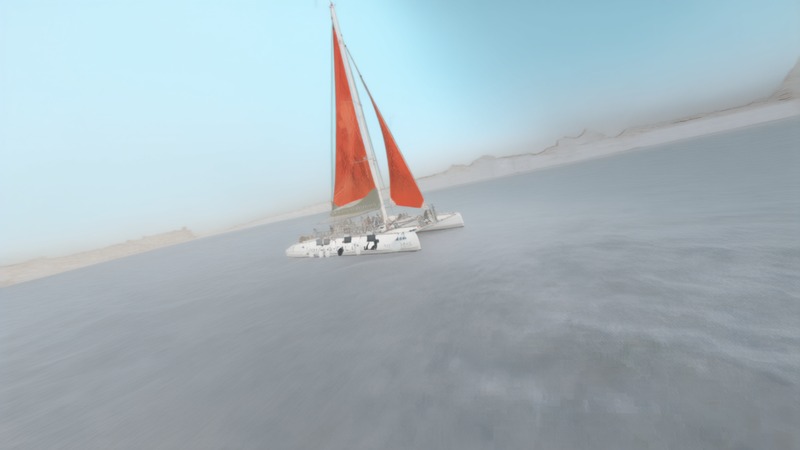}
        \put(15,59){Frame-wise Concat.}
        \end{overpic}\hfill
        \begin{overpic}[width=0.159\linewidth]{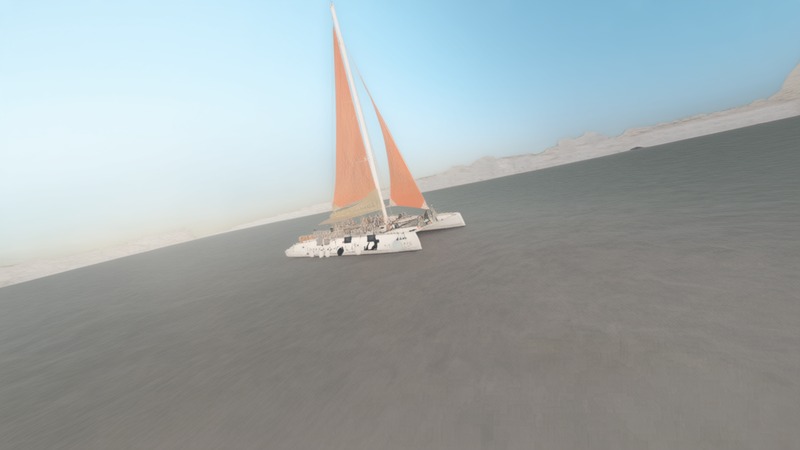}
        \put(4,59){\textbf{$+$ QK Type Embedding}}
        \end{overpic}\hfill
        \begin{overpic}[width=0.159\linewidth]{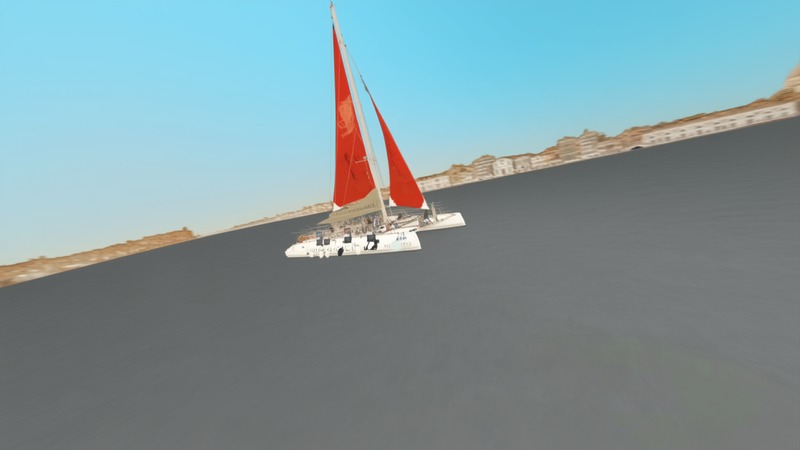}
        \put(4,59){\textbf{$+$ Clean Input Tokens}}
        \end{overpic}
    \end{minipage}\par
    \begin{minipage}{1.0\linewidth}
        \begin{overpic}[width=0.159\linewidth]{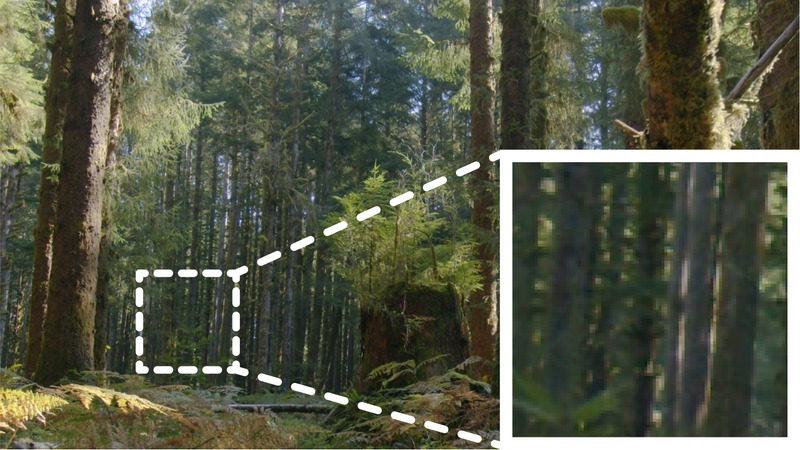}
        \end{overpic}\hspace{1em}
        \begin{overpic}[width=0.159\linewidth]{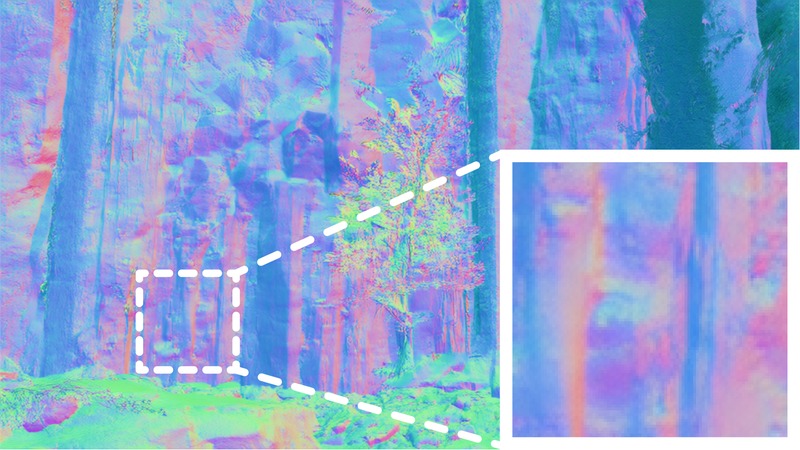}
        \put(-10,24){\rotatebox[origin=c]{90}{Normal}}
        \end{overpic}\hfill
        \begin{overpic}[width=0.159\linewidth]{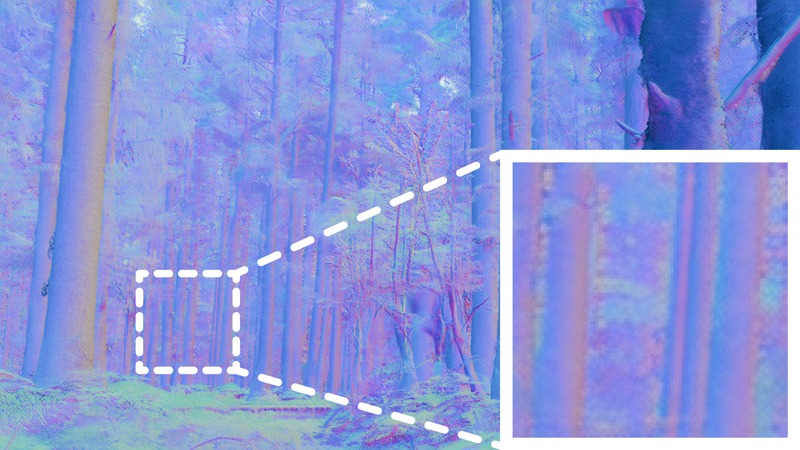}
        \end{overpic}\hfill
        \begin{overpic}[width=0.159\linewidth]{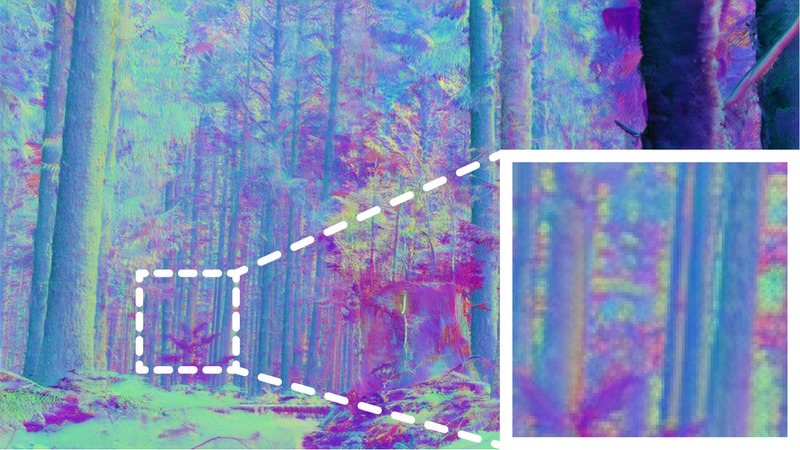}
        \end{overpic}\hfill
        \begin{overpic}[width=0.159\linewidth]{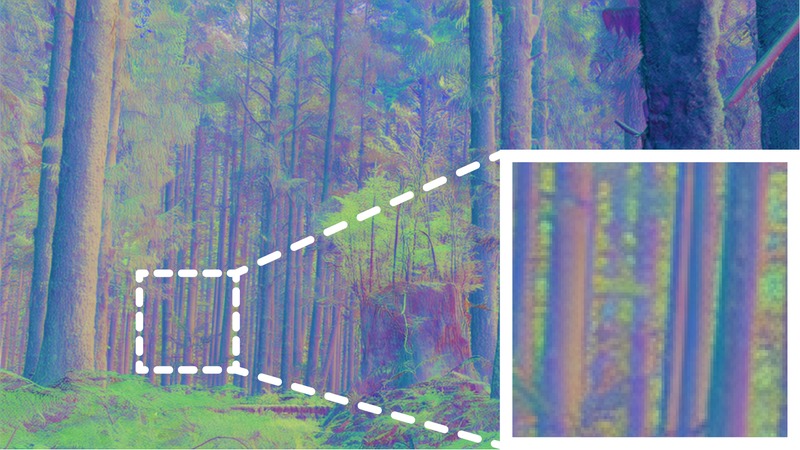}
        \end{overpic}\hfill
        \begin{overpic}[width=0.159\linewidth]{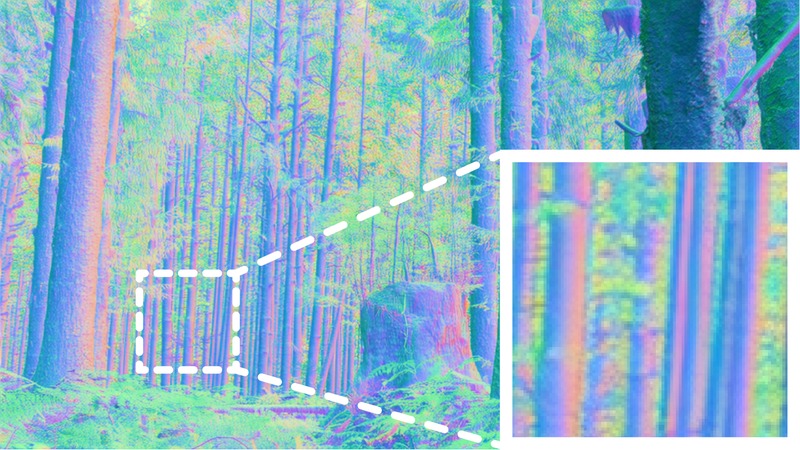}
        \end{overpic}
    \end{minipage}\par
    \begin{minipage}{1.0\linewidth}
        \begin{overpic}[width=0.159\linewidth]{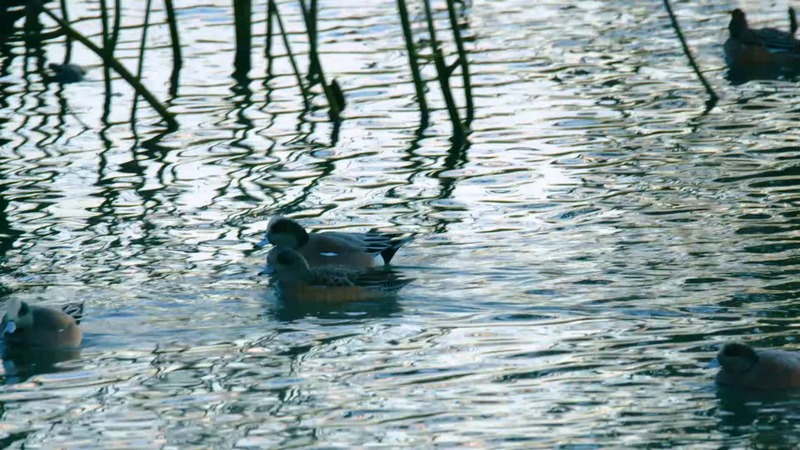}
        \end{overpic}\hspace{1em}
        \begin{overpic}[width=0.159\linewidth]{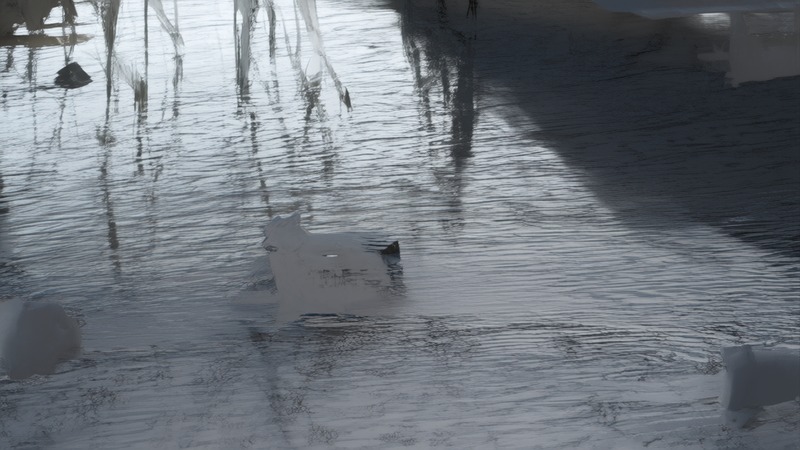}
        \put(-10,23){\rotatebox[origin=c]{90}{Irradiance}}
        \end{overpic}\hfill
        \begin{overpic}[width=0.159\linewidth]{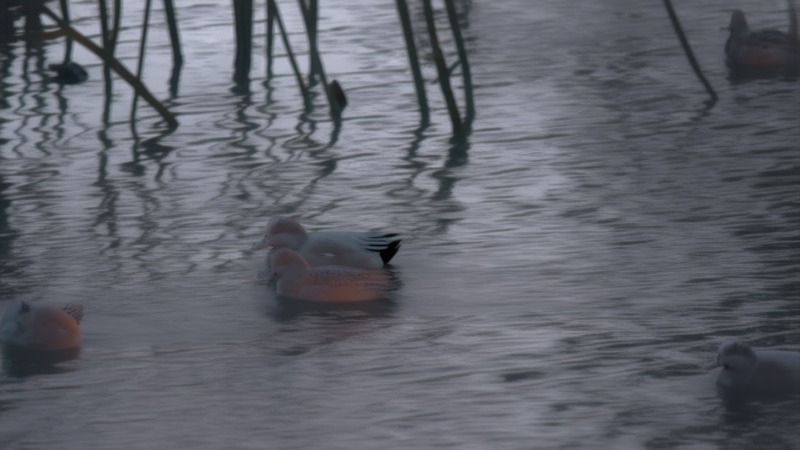}
        \end{overpic}\hfill
        \begin{overpic}[width=0.159\linewidth]{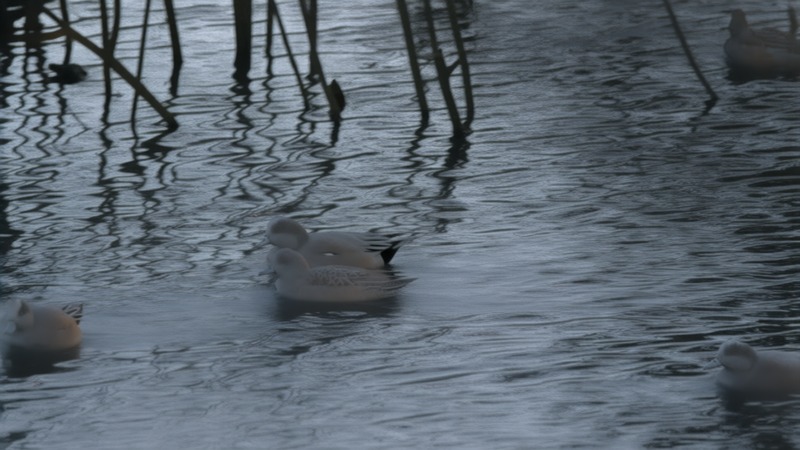}
        \end{overpic}\hfill
        \begin{overpic}[width=0.159\linewidth]{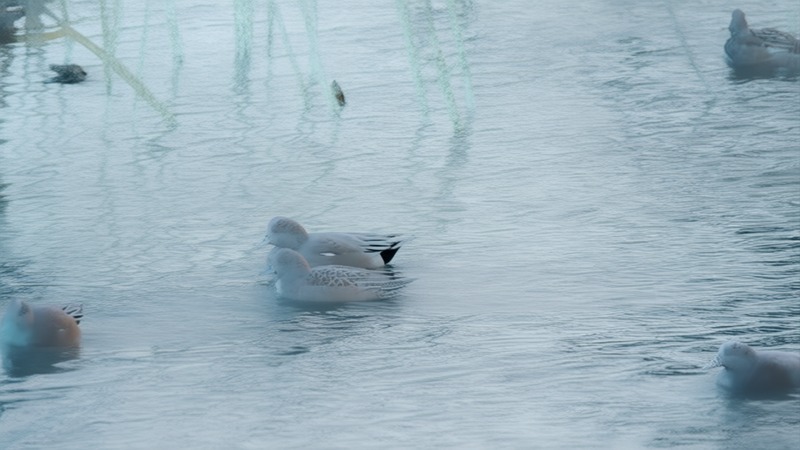}
        \end{overpic}\hfill
        \begin{overpic}[width=0.159\linewidth]{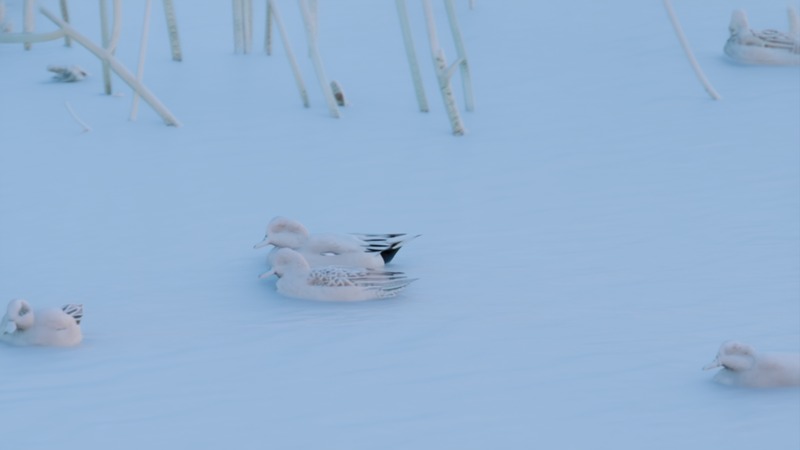}
        \end{overpic}
    \end{minipage}\par
    \begin{minipage}{1.0\linewidth}
        \begin{overpic}[width=0.159\linewidth]{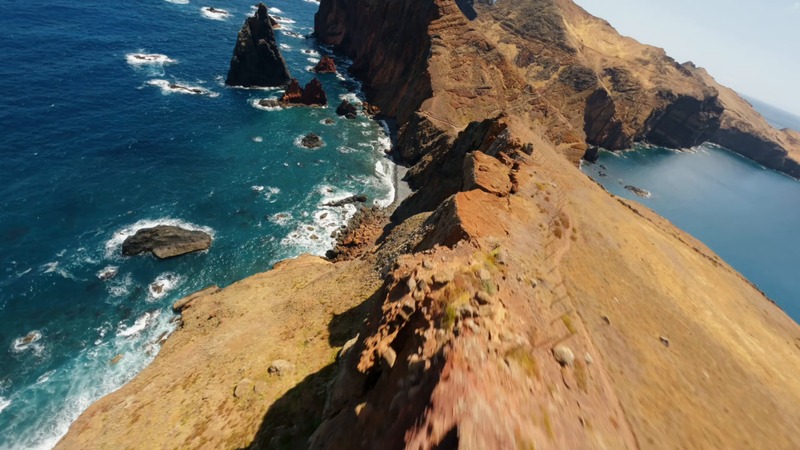}
        \end{overpic}\hspace{1em}
        \begin{overpic}[width=0.159\linewidth]{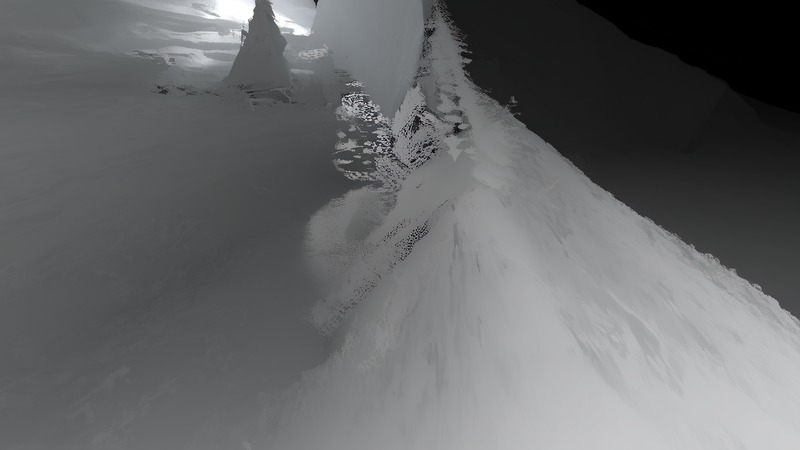}
        \put(-10,24){\rotatebox[origin=c]{90}{Depth}}
        \end{overpic}\hfill
        \begin{overpic}[width=0.159\linewidth]{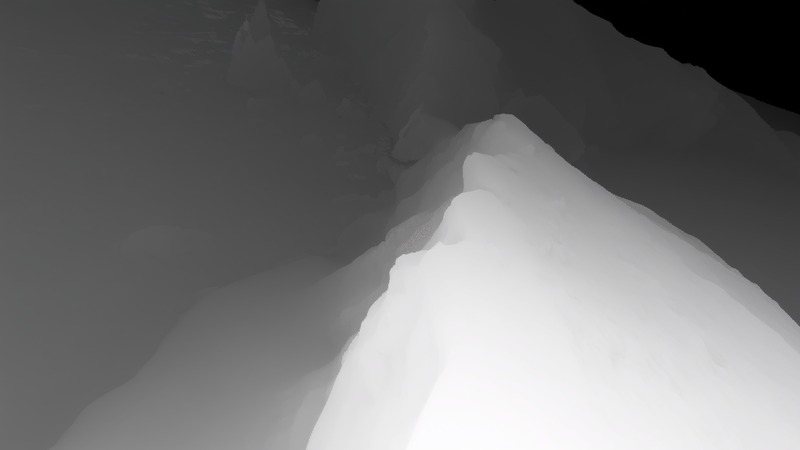}
        \end{overpic}\hfill
        \begin{overpic}[width=0.159\linewidth]{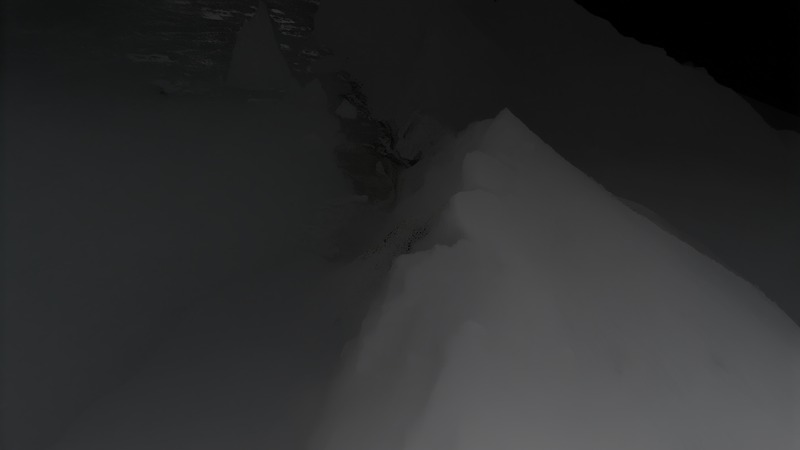}
        \end{overpic}\hfill
        \begin{overpic}[width=0.159\linewidth]{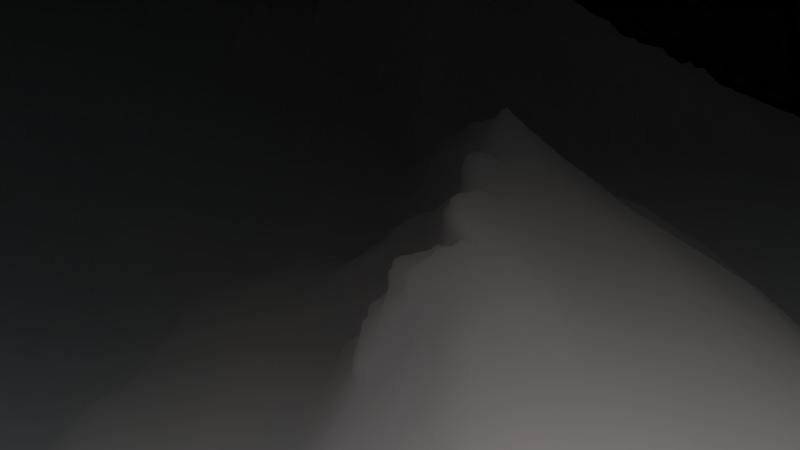}
        \end{overpic}\hfill
        \begin{overpic}[width=0.159\linewidth]{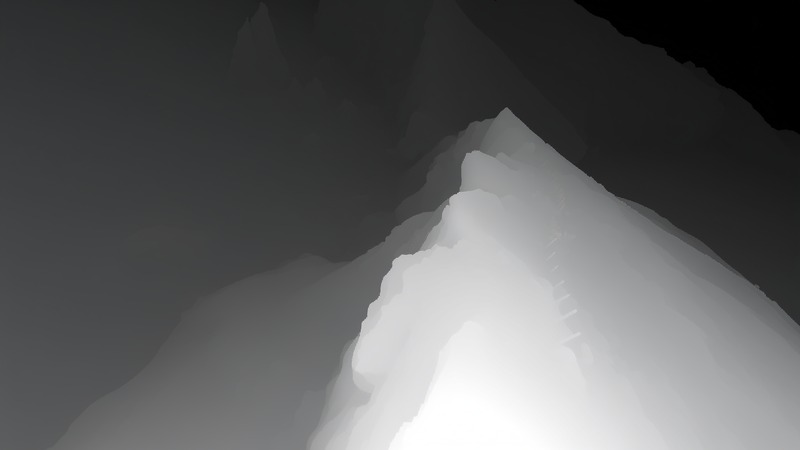}
        \end{overpic}
    \end{minipage}\par
    \begin{minipage}{1.0\linewidth}
        \begin{overpic}[width=0.159\linewidth]{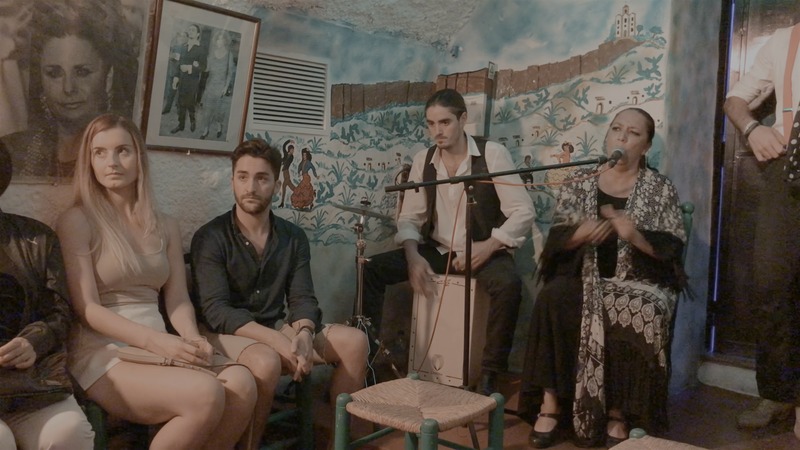}
        \end{overpic}\hspace{1em}
        \begin{overpic}[width=0.159\linewidth]{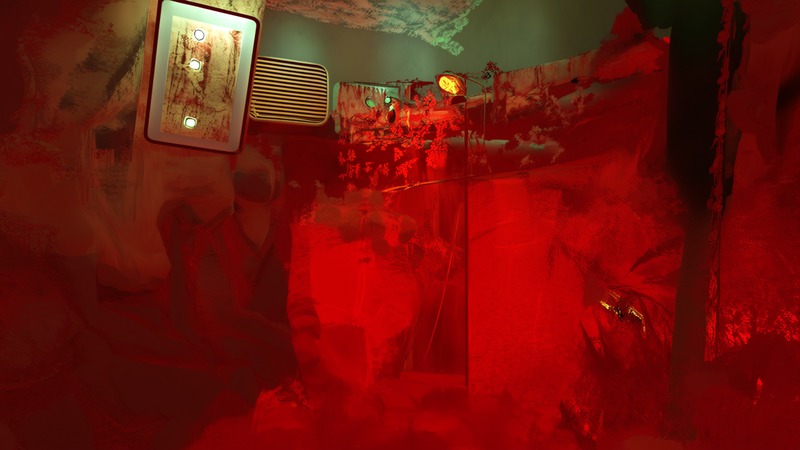}
        \put(-10,24){\rotatebox[origin=c]{90}{Material}}
        \end{overpic}\hfill
        \begin{overpic}[width=0.159\linewidth]{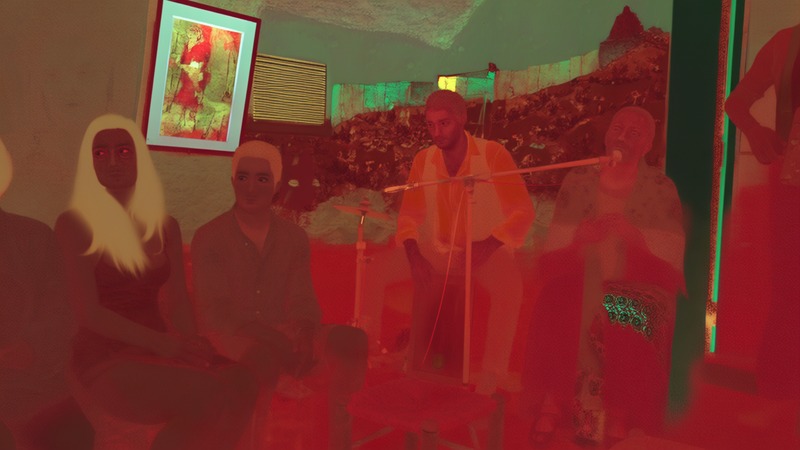}
        \end{overpic}\hfill
        \begin{overpic}[width=0.159\linewidth]{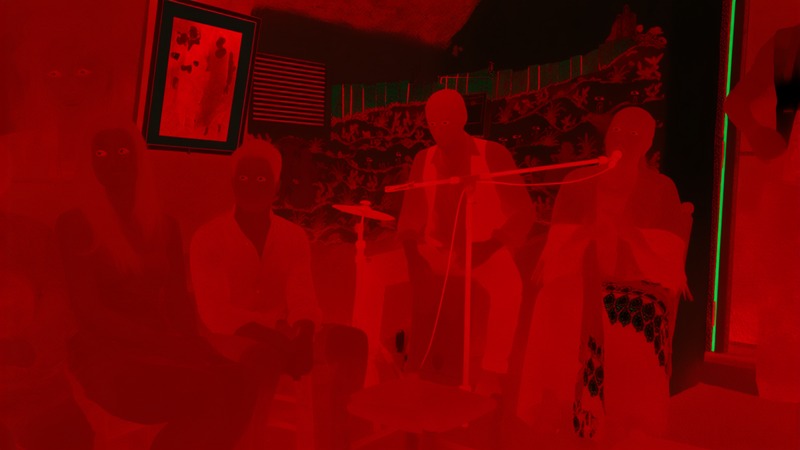}
        \end{overpic}\hfill
        \begin{overpic}[width=0.159\linewidth]{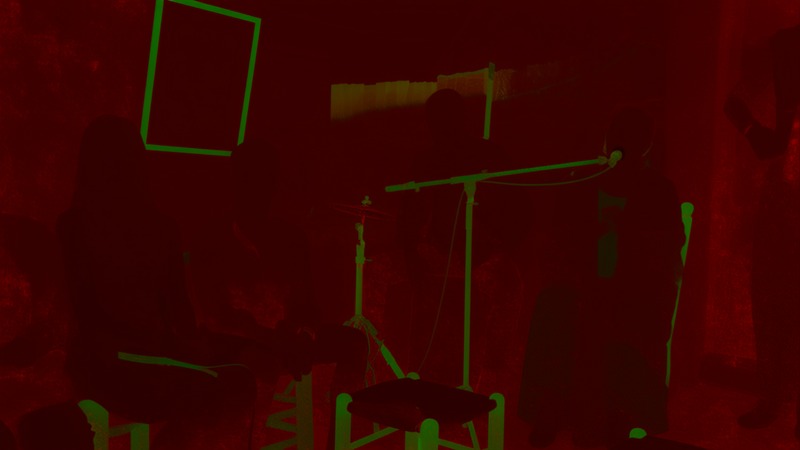}
        \end{overpic}\hfill
        \begin{overpic}[width=0.159\linewidth]{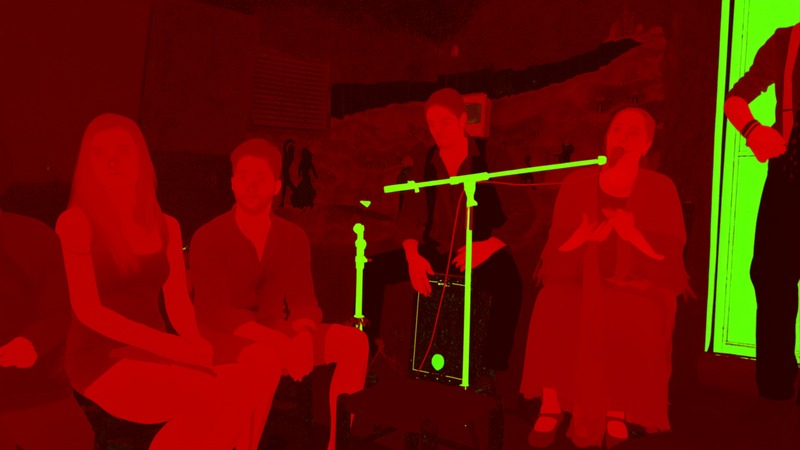}
        \end{overpic}
    \end{minipage}
    \caption{\edit{Ablation of \rgbtox conditioning choices and our proposed techniques. We compare channel-wise concatenation and VACE~\cite{jiang2025vace} with the preliminary frame-wise model, then add QK type embeddings and clean input tokens. All variants are trained for the same duration.}}
    \label{fig:rgb2x_ablation}
\end{figure*}

\begin{table}
\centering
\begin{tabular}{l|cccc}
\bf PSNR & Albedo & Normal & Irradiance & Depth \\
  \hline
\rgbx & 18.39 & 19.25 & 11.66 & N/A \\
\footnotesize DiffusionRenderer & 15.29 & \bf 21.54 & N/A & 17.72 \\
Ours & \bf 20.17 & 21.22 & \bf 25.19 & \bf 29.47 \\
  \hline
\bf SSIM & Albedo & Normal & Irradiance & Depth \\
  \hline
\rgbx & 0.7252 & 0.6901 & 0.7162 & N/A \\
\footnotesize DiffusionRenderer  & 0.7425 & 0.7372 & N/A & 0.6570 \\
Ours & \bf 0.8219 & \bf 0.7638 & \bf 0.8589 & \bf 0.9032 \\
  \hline
\bf LPIPS & Albedo & Normal & Irradiance & Depth \\
  \hline
\rgbx  & 0.1920 & 0.2225 & \bf 0.2673 & N/A \\
\footnotesize DiffusionRenderer & 0.2675 & 0.2275 & N/A & 0.3309 \\
Ours & \bf 0.1420 & \bf 0.1986 & 0.3535 & \bf 0.1166 \\
\end{tabular}
\caption{Quantitative comparison of our \rgbtox results to \rgbx~\cite{zeng2024rgbx} and DiffusionRenderer~\cite{liang2025diffusionrenderer} on the test set of the Hypersim dataset. Our result is best across all three metrics for depth and albedo, and in two out of three metrics on normal and irradiance.}
\label{tab:quantitative}
\end{table}

\edit{We first evaluate our \rgbtox model. The supplementary video shows temporally coherent albedo estimates (0:10--1:09) and irradiance estimates (1:12--2:12), both central targets of intrinsic decomposition. Normals, depth, and material estimates are shown in 2:18--2:44.}

\paragraph{Comparisons} \edit{In \cref{fig:rgb2x_baseline}, we visually compare our G-buffer estimates to \rgbx \cite{zeng2024rgbx} and the Cosmos-based version of DiffusionRenderer \cite{liang2025diffusionrenderer,agarwal2025cosmos}, using the official checkpoints released by the authors. On held-out \textsf{(RGB, X)} test set, \rgbx produces less accurate estimates, consistent with its older Stable Diffusion 2.1 backbone and indoor-only training data. DiffusionRenderer produces normals of similar visual quality, but its albedo estimates retain more residual shading and its depth prediction fails on sky regions. The methods also differ in modality coverage: \rgbx does not support depth, while DiffusionRenderer does not support irradiance. Our model therefore enables irradiance estimation on real video footage in this comparison; see the supplementary video (1:11--2:12), with additional visual comparisons at 2:51--3:32. In \cref{fig:rgb2x_ablation}, we also compare conditioning mechanisms against channel-wise concatenation and VACE \cite{jiang2025vace}. Channel-wise concatenation, used in \rgbx and DiffusionRenderer, extends patchification to two concatenated latent frames: a clean input and a noisy output. This uses fewer tokens, but convergence is significantly worse because the model must quickly repurpose attention patterns learned between video frames into attention between input and output signals. VACE instead injects input signals through control tokens.}

\paragraph{Quantitative comparisons} \edit{In \cref{tab:quantitative}, we quantitatively compare our \rgbtox results to previous methods on the Hypersim test set \cite{hypersim}. Our model is best across all three metrics for albedo and depth, and best in two out of three metrics for normal and irradiance. Since no publicly available dataset provides reliable ground-truth roughness and metallicity buffers, we do not evaluate material properties quantitatively. Note that \rgbx \cite{zeng2024rgbx} was trained on Hypersim whereas our model was not. Hypersim is also synthetic, so improvements on real inputs are better judged visually.}

\paragraph{Ablations} \edit{In \cref{fig:rgb2x_ablation}, we also ablate the two key techniques introduced in \cref{sec:improved-rgbx}. Starting from the preliminary frame-wise \rgbtox model described in \cref{sec:initial-rgbx}, adding QK type embeddings improves the estimates across modalities, and adding clean input tokens further improves quality and convergence.}

\subsection{\xtorgb results}

\begin{figure*}[t]
    \vspace{1mm}
    \footnotesize
    \centering
    \begin{minipage}{1.0\linewidth}
        \begin{overpic}[width=0.16\linewidth]{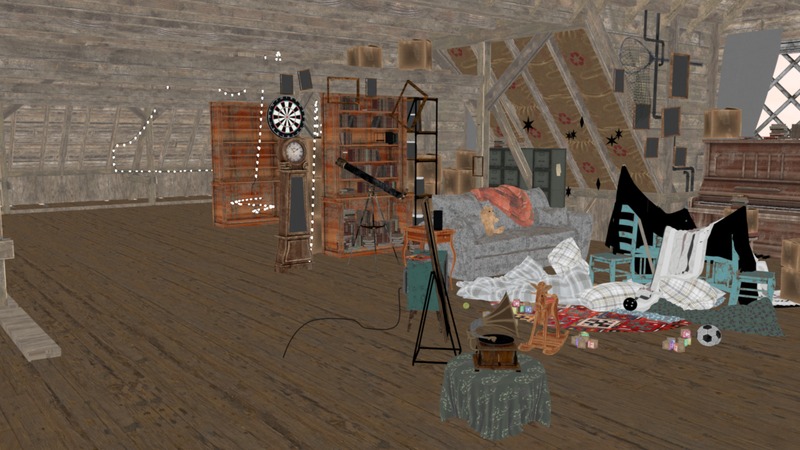}
        \put(20,58){Input G-buffers}
        \put(0,2){
        \begin{tikzpicture} \node[fill=black!10,fill opacity=0.75,rounded corners=1ex, text width=0.7cm,align=left] {Albedo}; \end{tikzpicture}
        }
        \end{overpic}\hfill
        \begin{overpic}[width=0.16\linewidth]{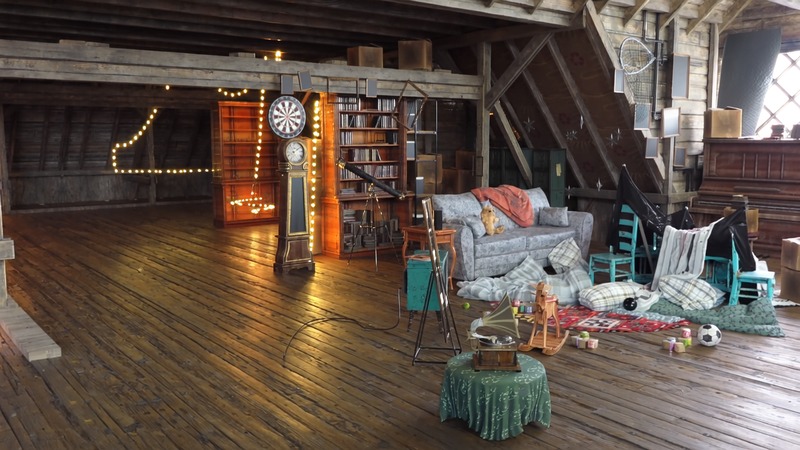}
        \put(38,58){\textbf{Ours}}
        \end{overpic}\hspace{0.2em}
        \begin{overpic}[width=0.16\linewidth]{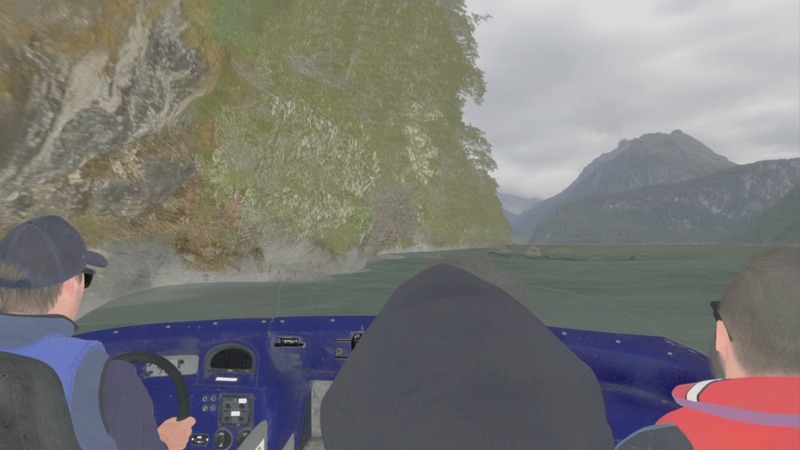}
        \put(20,58){Input G-buffers}
        \put(0,2){
        \begin{tikzpicture} \node[fill=black!10,fill opacity=0.75,rounded corners=1ex, text width=0.7cm,align=left] {Albedo}; \end{tikzpicture}
        }
        \end{overpic}\hfill
        \begin{overpic}[width=0.16\linewidth]{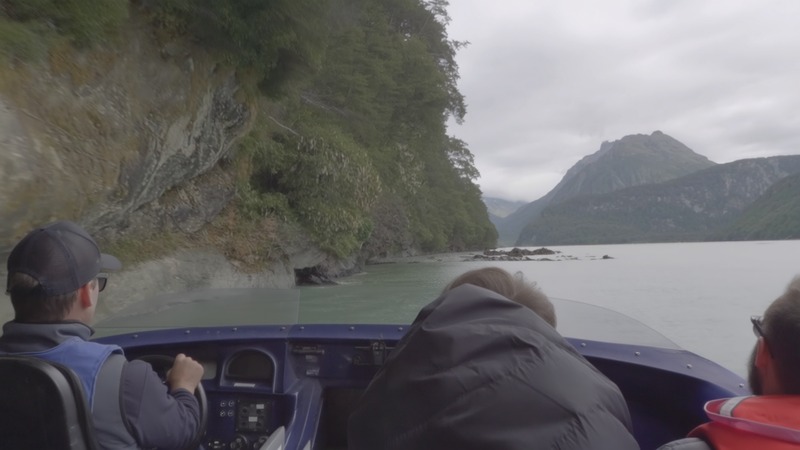}
        \put(38,58){\textbf{Ours}}
        \end{overpic}\hspace{0.2em}
        \begin{overpic}[width=0.16\linewidth]{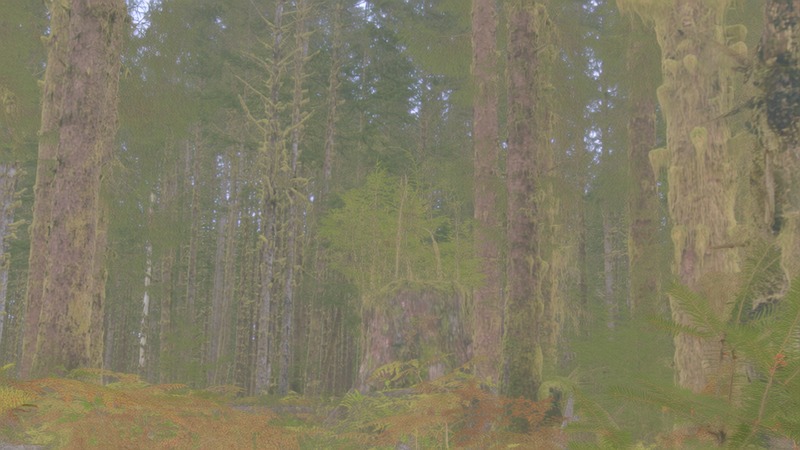}
        \put(20,58){Input G-buffers}
        \put(0,2){
        \begin{tikzpicture} \node[fill=black!10,fill opacity=0.75,rounded corners=1ex, text width=0.7cm,align=left] {Albedo}; \end{tikzpicture}
        }
        \end{overpic}\hfill
        \begin{overpic}[width=0.16\linewidth]{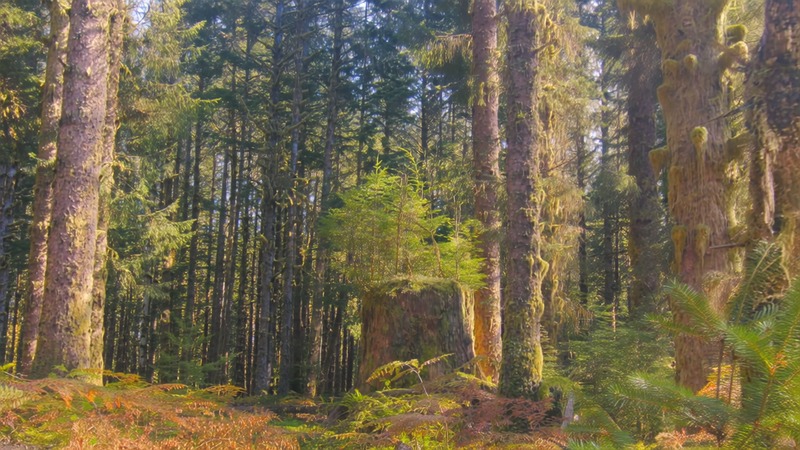}
        \put(38,58){\textbf{Ours}}
        \end{overpic}
    \end{minipage}\par\smallskip
    \begin{minipage}{1.0\linewidth}
        \begin{overpic}[width=0.16\linewidth]{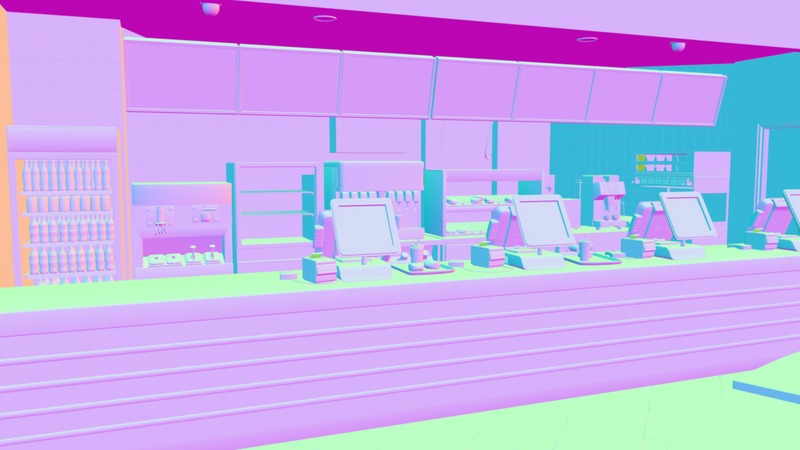}
        \put(0,2){
        \begin{tikzpicture} \node[fill=black!10,fill opacity=0.75,rounded corners=1ex, text width=0.8cm,align=left] {Normal}; \end{tikzpicture}
        }
        \end{overpic}\hfill
        \begin{overpic}[width=0.16\linewidth]{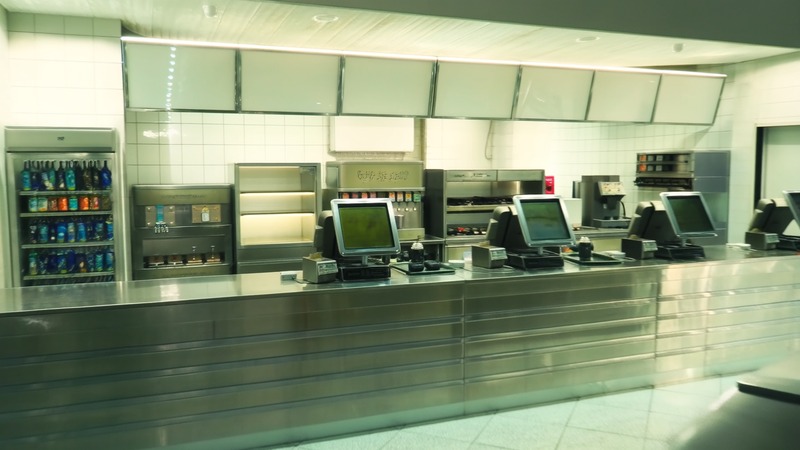}
        \end{overpic}\hspace{0.2em}
        \begin{overpic}[width=0.16\linewidth]{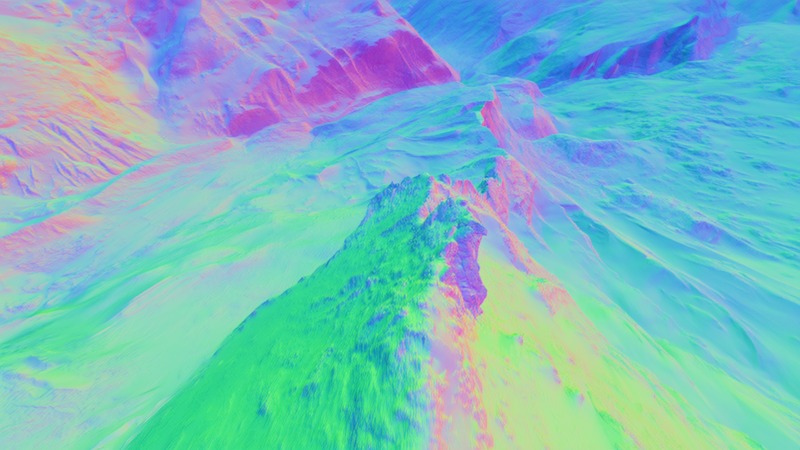}
        \put(0,2){
        \begin{tikzpicture} \node[fill=black!10,fill opacity=0.75,rounded corners=1ex, text width=0.8cm,align=left] {Normal}; \end{tikzpicture}
        }
        \end{overpic}\hfill
        \begin{overpic}[width=0.16\linewidth]{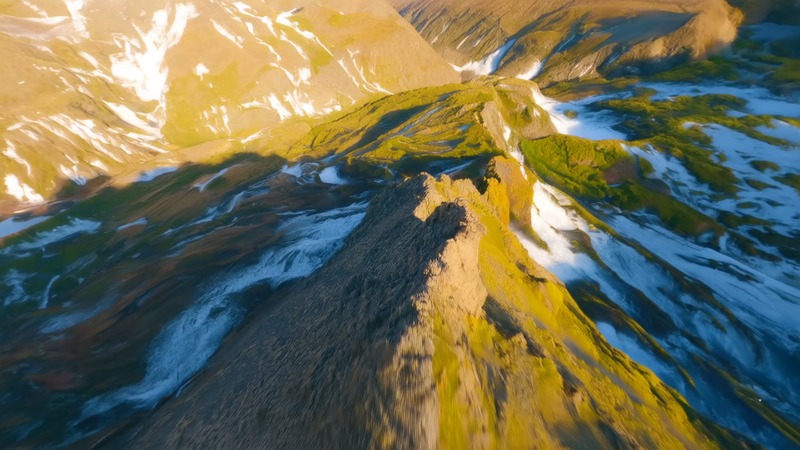}
        \end{overpic}\hspace{0.2em}
        \begin{overpic}[width=0.16\linewidth]{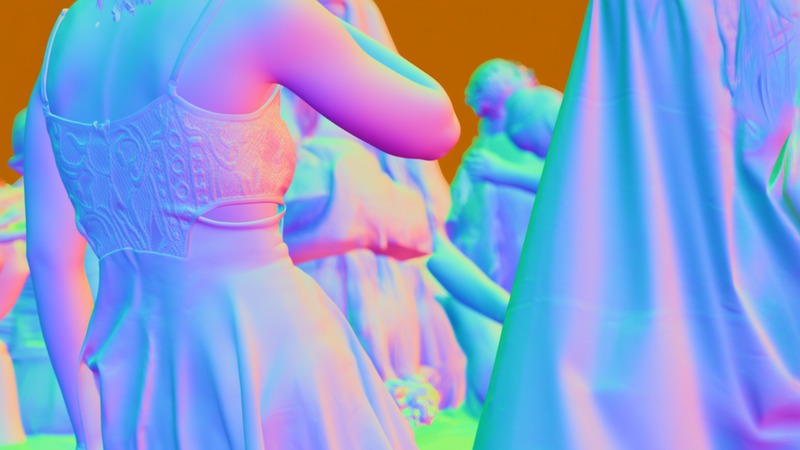}
        \put(0,2){
        \begin{tikzpicture} \node[fill=black!10,fill opacity=0.75,rounded corners=1ex, text width=0.8cm,align=left] {Normal}; \end{tikzpicture}
        }
        \end{overpic}\hfill
        \begin{overpic}[width=0.16\linewidth]{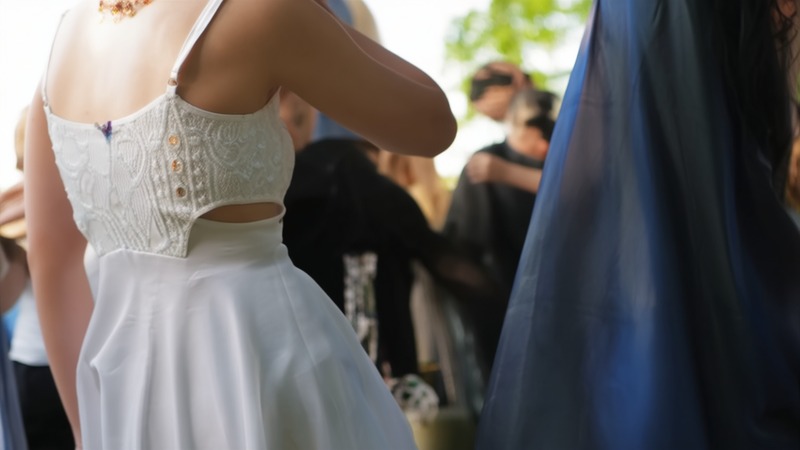}
        \end{overpic}
    \end{minipage}\par\smallskip
    \begin{minipage}{1.0\linewidth}
        \begin{overpic}[width=0.16\linewidth]{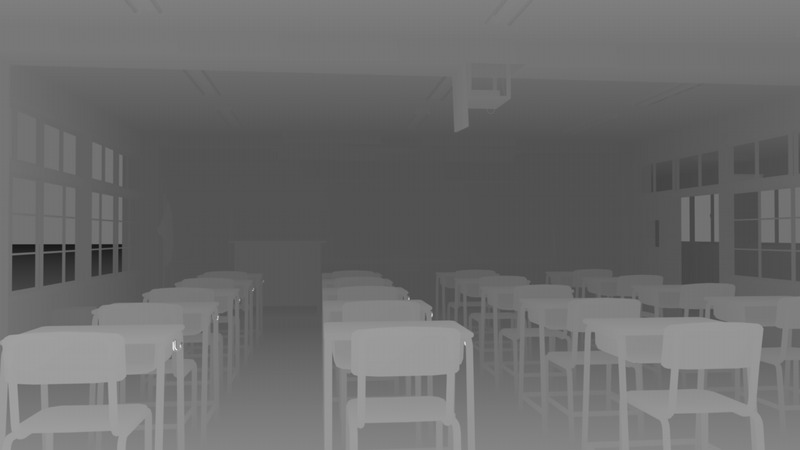}
        \put(0,2){
        \begin{tikzpicture} \node[fill=black!10,fill opacity=0.75,rounded corners=1ex, text width=0.6cm,align=left] {Depth}; \end{tikzpicture}
        }
        \end{overpic}\hfill
        \begin{overpic}[width=0.16\linewidth]{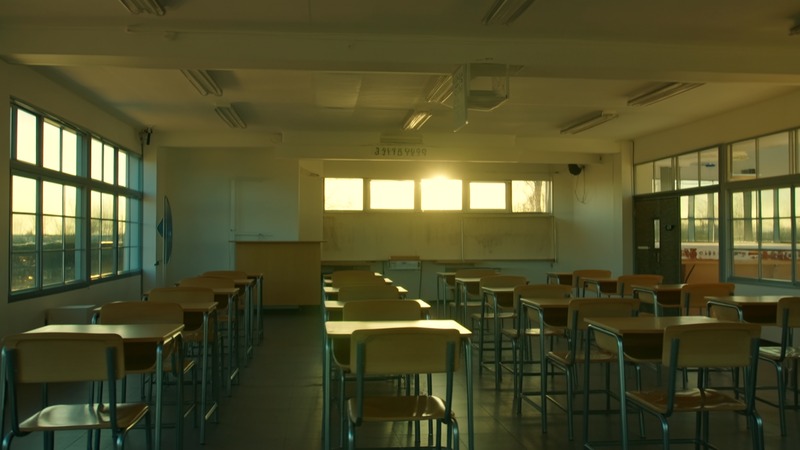}
        \end{overpic}\hspace{0.2em}
        \begin{overpic}[width=0.16\linewidth]{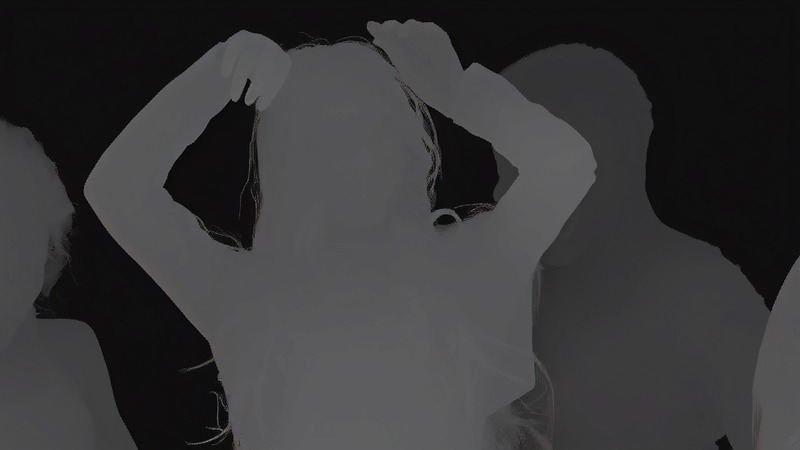}
        \put(0,2){
        \begin{tikzpicture} \node[fill=black!10,fill opacity=0.75,rounded corners=1ex, text width=0.6cm,align=left] {Depth}; \end{tikzpicture}
        }
        \end{overpic}\hfill
        \begin{overpic}[width=0.16\linewidth]{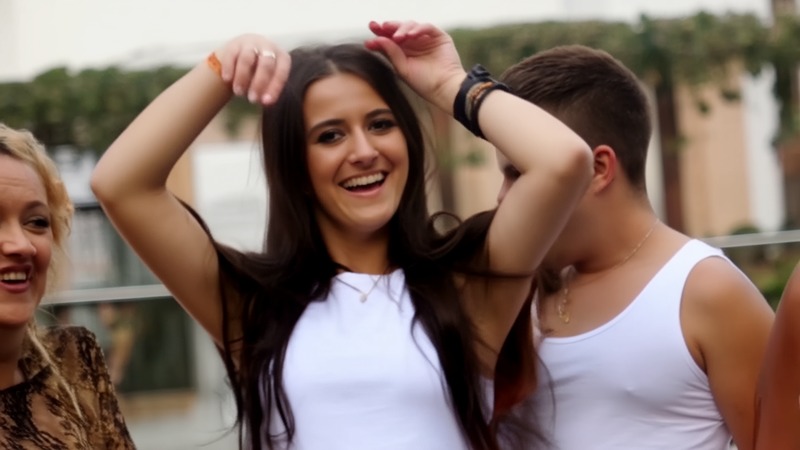}
        \end{overpic}\hspace{0.2em}
        \begin{overpic}[width=0.16\linewidth]{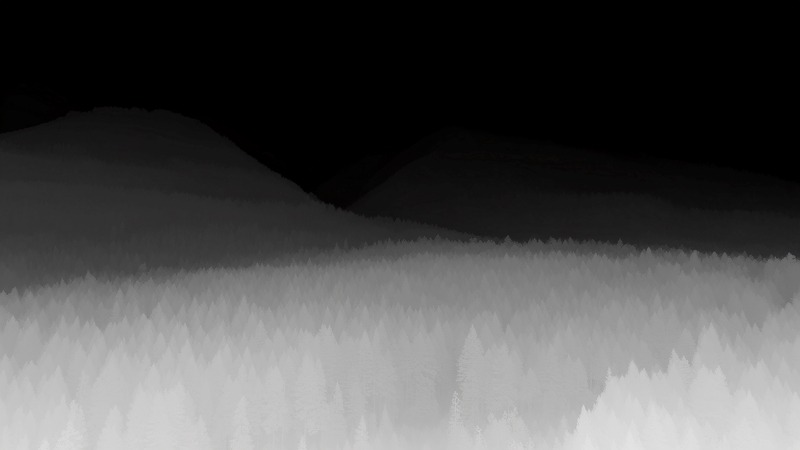}
        \put(0,2){
        \begin{tikzpicture} \node[fill=black!10,fill opacity=0.75,rounded corners=1ex, text width=0.8cm,align=left] {Depth}; \end{tikzpicture}
        }
        \end{overpic}\hfill
        \begin{overpic}[width=0.16\linewidth]{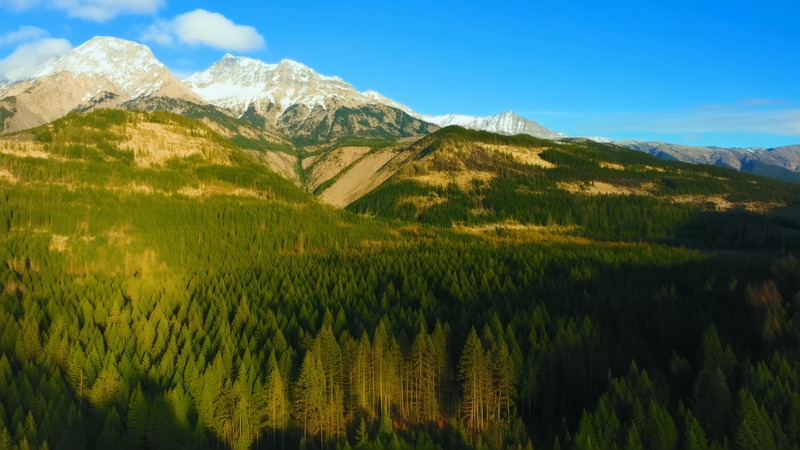}
        \end{overpic}
    \end{minipage}\par\smallskip
    \begin{minipage}{1.0\linewidth}
        \begin{overpic}[width=0.16\linewidth]{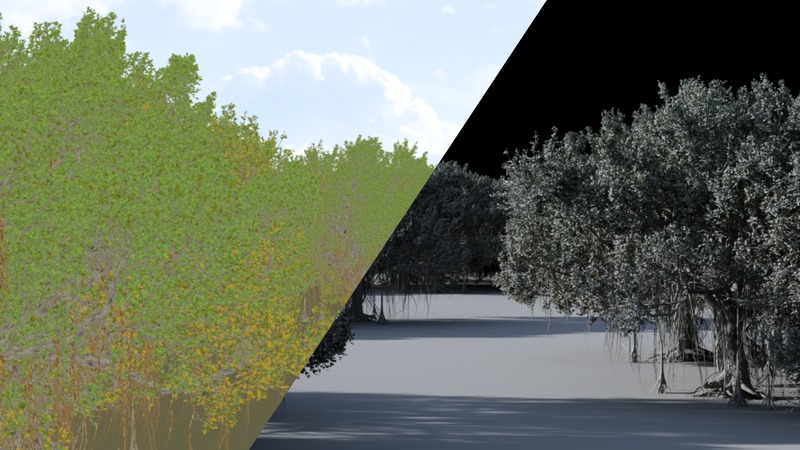}
        \put(0,2){
        \begin{tikzpicture} \node[fill=black!10,fill opacity=0.75,rounded corners=1ex, text width=2.1cm,align=left] {Albedo / Irradiance}; \end{tikzpicture}
        }
        \end{overpic}\hfill
        \begin{overpic}[width=0.16\linewidth]{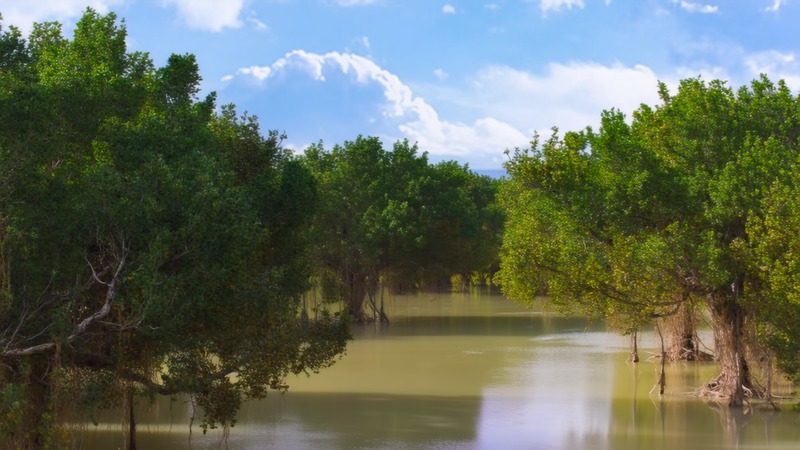}
        \end{overpic}\hspace{0.2em}
        \begin{overpic}[width=0.16\linewidth]{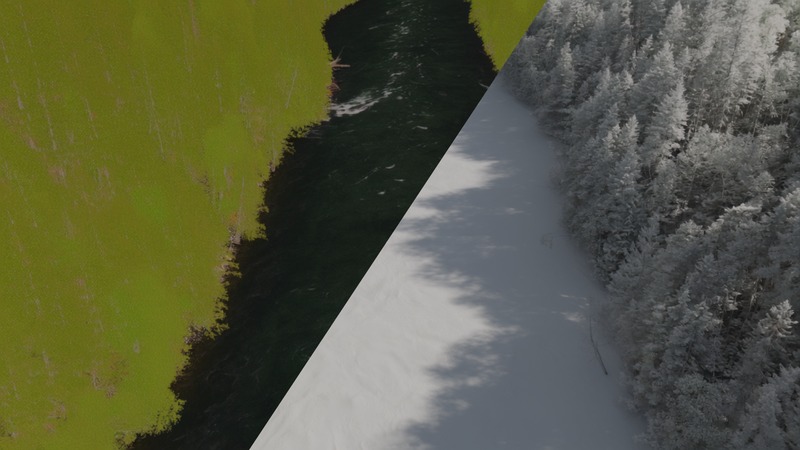}
        \put(0,2){
        \begin{tikzpicture} \node[fill=black!10,fill opacity=0.75,rounded corners=1ex, text width=2.1cm,align=left] {Albedo / Irradiance}; \end{tikzpicture}
        }
        \end{overpic}\hfill
        \begin{overpic}[width=0.16\linewidth]{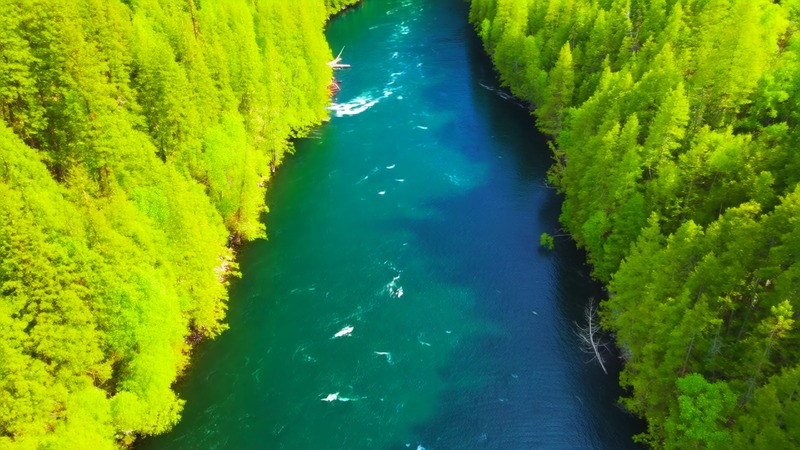}
        \end{overpic}\hspace{0.2em}
        \begin{overpic}[width=0.16\linewidth]{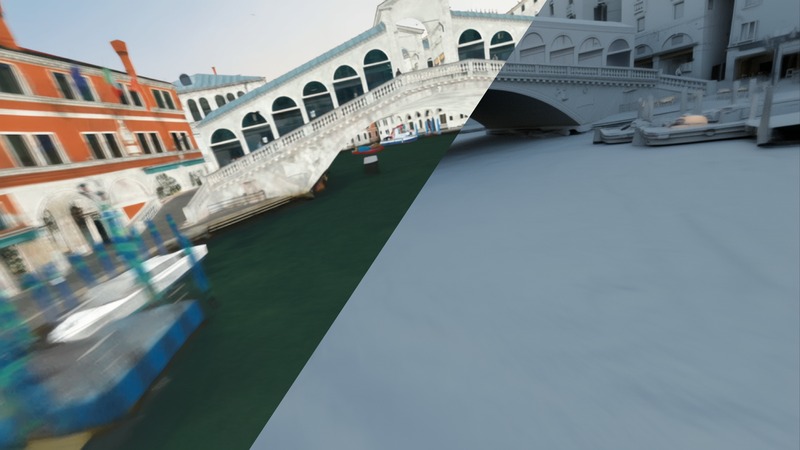}
        \put(0,2){
        \begin{tikzpicture} \node[fill=black!10,fill opacity=0.75,rounded corners=1ex, text width=2.1cm,align=left] {Albedo / Irradiance}; \end{tikzpicture}
        }
        \end{overpic}\hfill
        \begin{overpic}[width=0.16\linewidth]{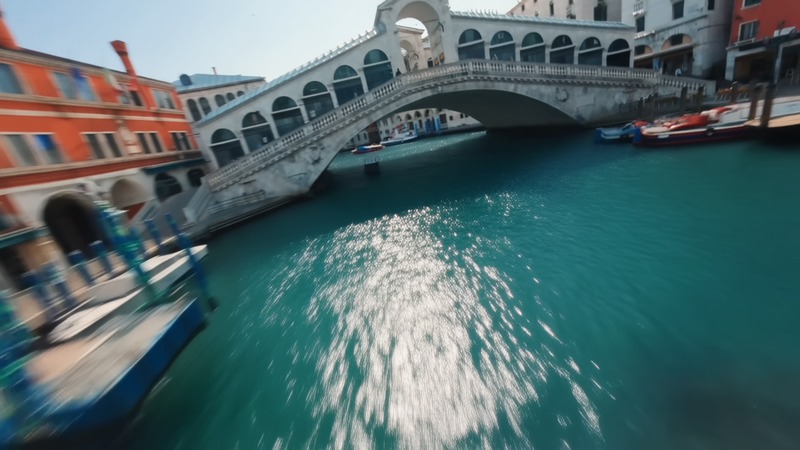}
        \end{overpic}
    \end{minipage}\par\smallskip
    \begin{minipage}{1.0\linewidth}
        \begin{overpic}[width=0.16\linewidth]{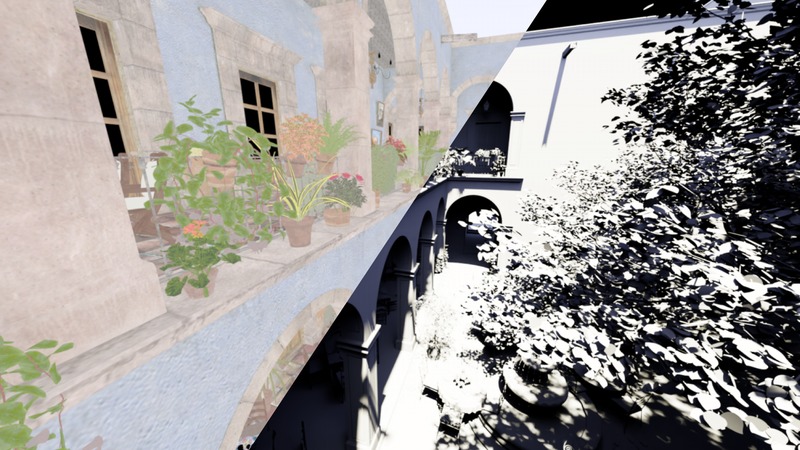}
        \put(0,2){
        \begin{tikzpicture} \node[fill=black!10,fill opacity=0.75,rounded corners=1ex, text width=2.5cm,align=left] {Albedo / Direct lighting}; \end{tikzpicture}
        }
        \end{overpic}\hfill
        \begin{overpic}[width=0.16\linewidth]{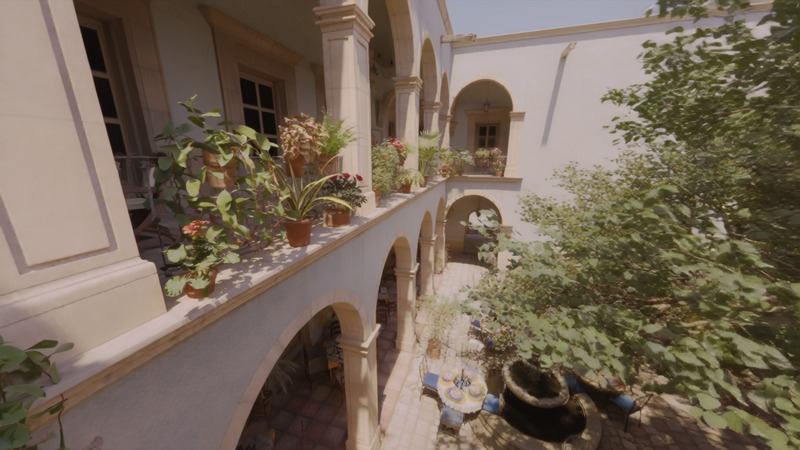}
        \end{overpic}\hspace{0.2em}
        \begin{overpic}[width=0.16\linewidth]{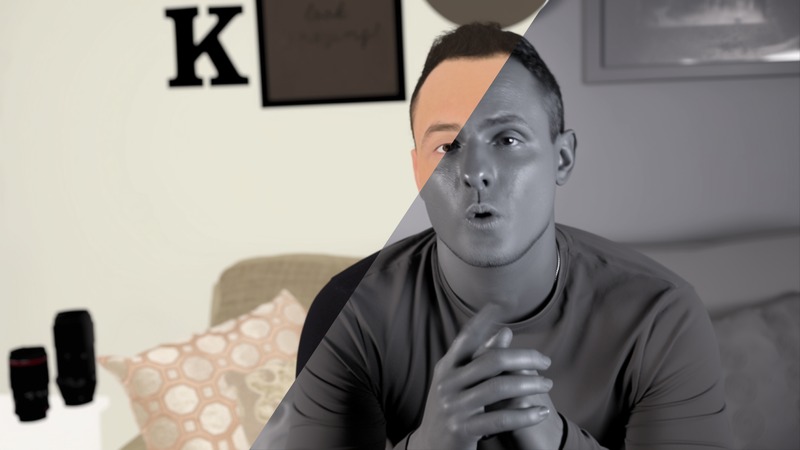}
        \put(0,2){
        \begin{tikzpicture} \node[fill=black!10,fill opacity=0.75,rounded corners=1ex, text width=2.5cm,align=left] {Albedo / Direct lighting}; \end{tikzpicture}
        }
        \end{overpic}\hfill
        \begin{overpic}[width=0.16\linewidth]{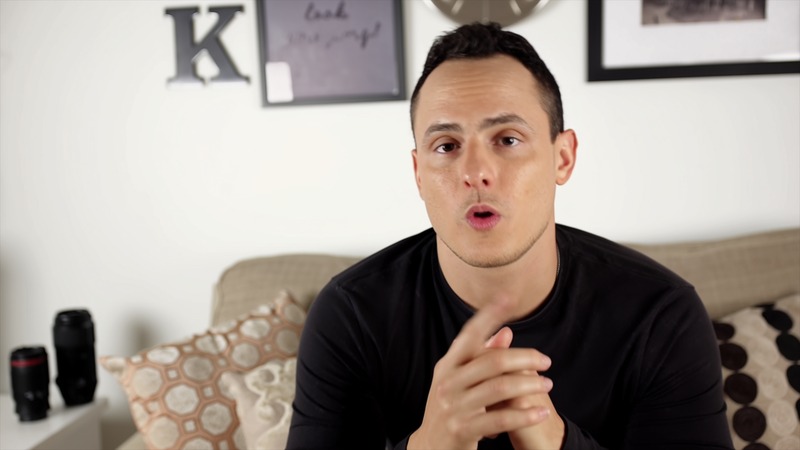}
        \end{overpic}\hspace{0.2em}
        \begin{overpic}[width=0.16\linewidth]{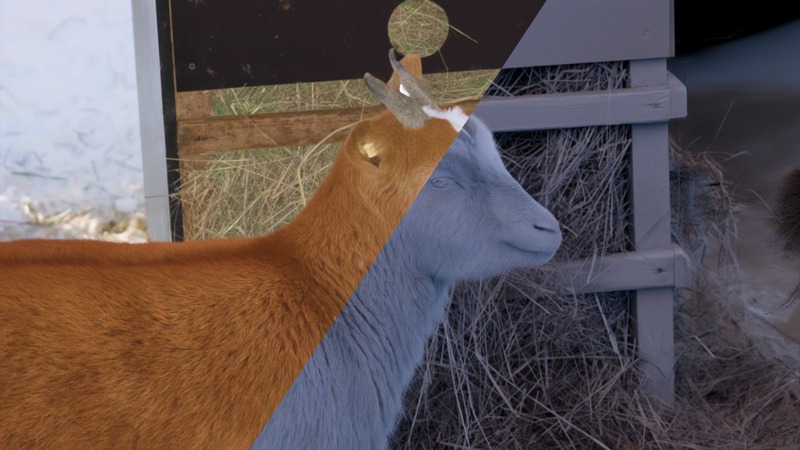}
        \put(0,2){
        \begin{tikzpicture} \node[fill=black!10,fill opacity=0.75,rounded corners=1ex, text width=2.5cm,align=left] {Albedo / Direct lighting}; \end{tikzpicture}
        }
        \end{overpic}\hfill
        \begin{overpic}[width=0.16\linewidth]{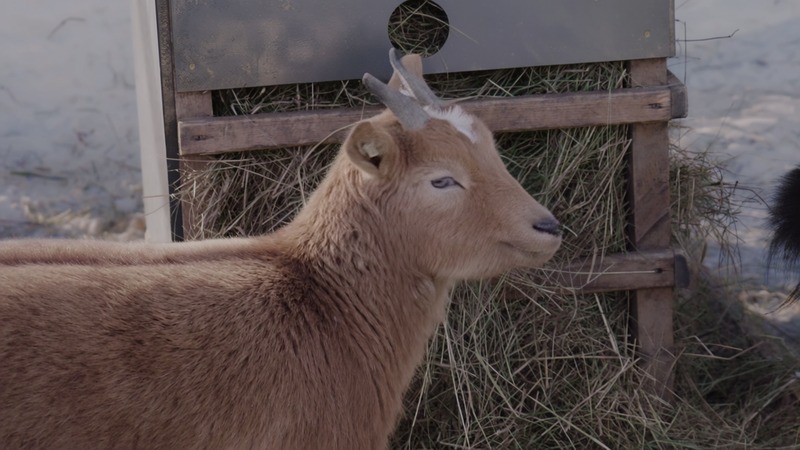}
        \end{overpic}
    \end{minipage}\par
    \caption{Results from various \xtorgb models, conditioned on albedo, normal, depth, and combinations of albedo with irradiance (blurred and \edit{dropped after the first 20\% diffusion iterations}), and albedo with albedo-free direct lighting. The first column uses synthetic G-buffers rendered from 3D scenes; the remaining columns use G-buffers estimated from real videos. All models support video; please see supplementary video for many more video results.}
    \label{fig:x2rgb_ours}
\end{figure*}
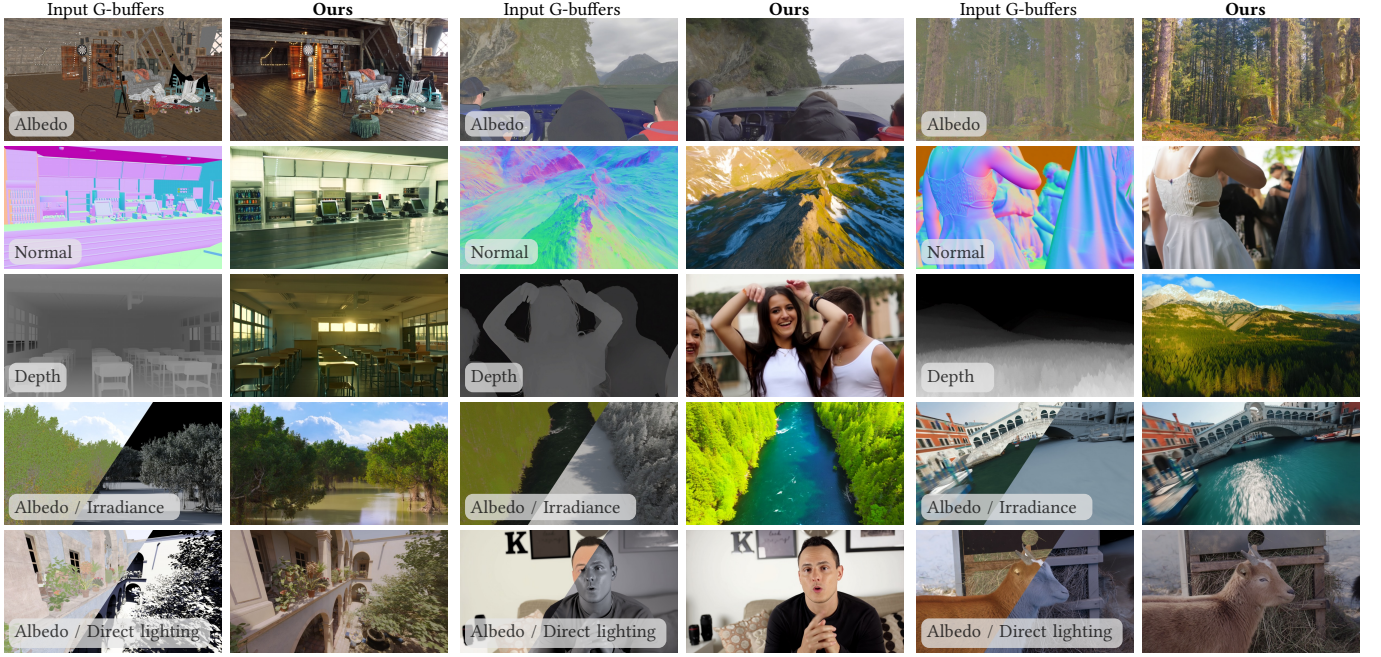

\edit{In \cref{fig:x2rgb_ours}, we illustrate specialized \xtorgb models trained to condition on different G-buffer inputs. The first three rows use separate models trained to condition on albedo, normal, and depth. The last two rows show lighting-guided generation from models trained to condition on albedo with irradiance, where irradiance is blurred and dropped after the first $20\%$ of denoising iterations, and albedo with direct lighting. The first column uses synthetic G-buffers rendered from 3D scenes, while the remaining examples use G-buffers estimated from real videos by our \rgbtox model. Static images only partially capture the model capabilities, so we refer readers to the supplementary video for additional results: 3:36--7:11 for albedo guidance, including multiple seeds and text-prompt comparisons; 7:16--7:33 for depth guidance; 7:43--8:11 for normal guidance; 8:18--8:40 for albedo with irradiance; and 8:49--9:20 for albedo with direct illumination.}

\begin{figure}[t]
    \vspace{1mm}
    \footnotesize
    \begin{minipage}{1\linewidth}
        \begin{overpic}[width=0.3279\linewidth]{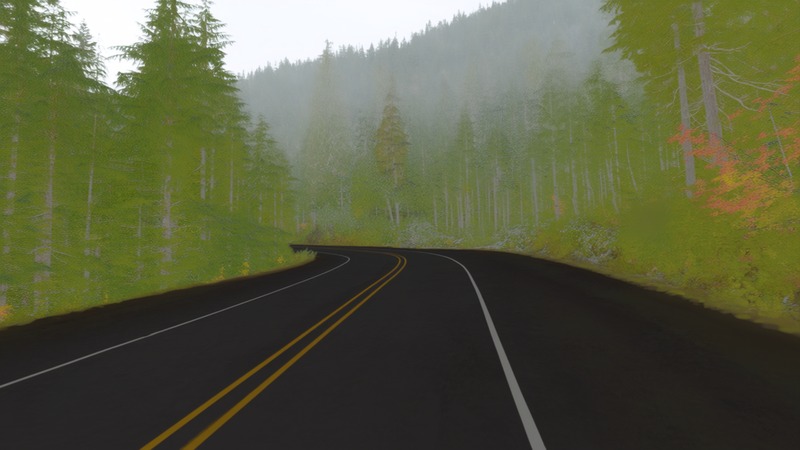}
        \put(20,59){Input G-buffers}
        \put(0,2){
        \begin{tikzpicture} \node[fill=black!10,fill opacity=0.75,rounded corners=1ex, text width=0.7cm,align=left] {Albedo}; \end{tikzpicture}
        }
        \end{overpic}\hfill
        \begin{overpic}[width=0.3279\linewidth]{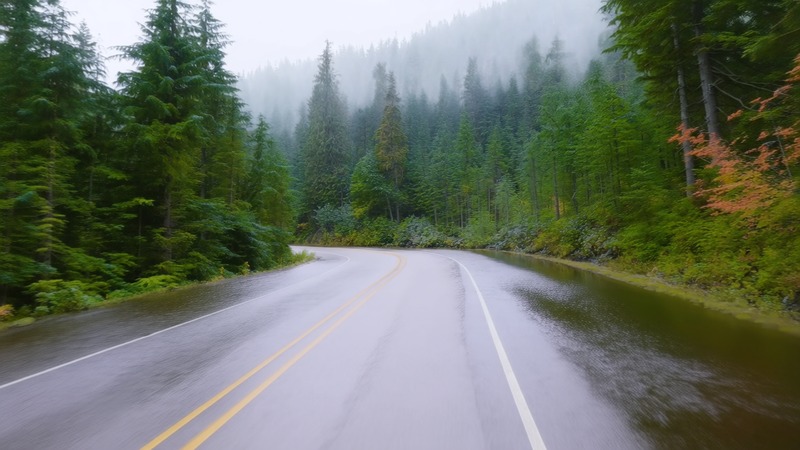}
        \put(20,59){Ask for realistic}
        \end{overpic}\hfill
        \begin{overpic}[width=0.3279\linewidth]{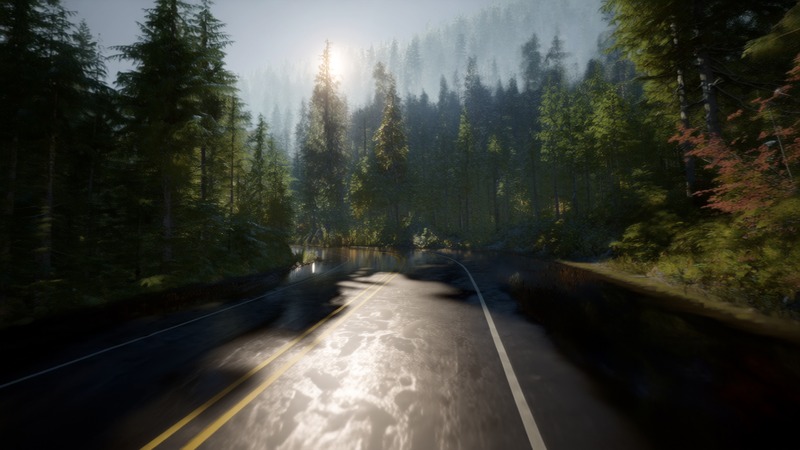}
        \put(20,59){Ask for synthetic}
        \end{overpic}
    \end{minipage}\par\smallskip
    \begin{minipage}{1\linewidth}
        \begin{overpic}[width=0.3279\linewidth]{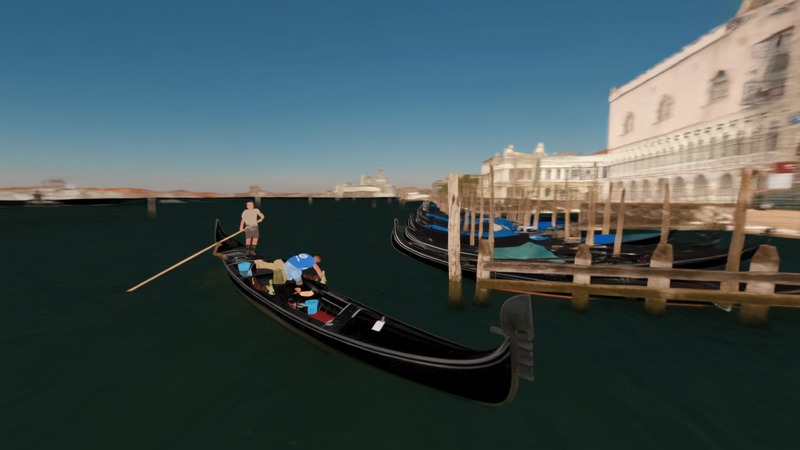}
        \put(0,2){
        \begin{tikzpicture} \node[fill=black!10,fill opacity=0.75,rounded corners=1ex, text width=0.7cm,align=left] {Albedo}; \end{tikzpicture}
        }
        \end{overpic}\hfill
        \begin{overpic}[width=0.3279\linewidth]{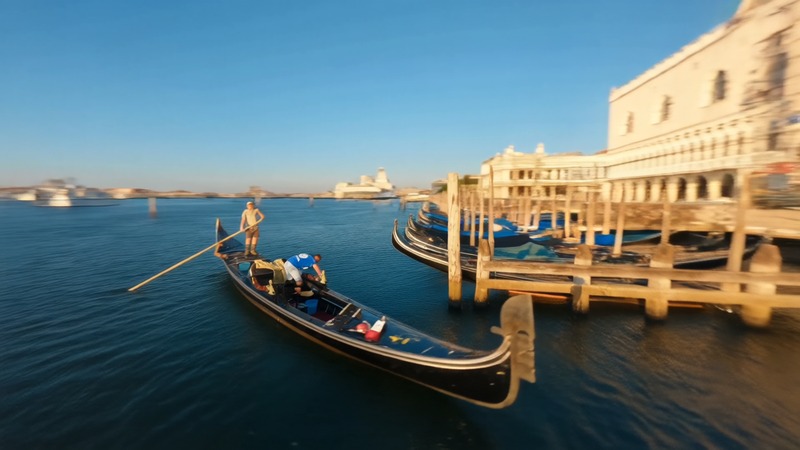}
        \end{overpic}\hfill
        \begin{overpic}[width=0.3279\linewidth]{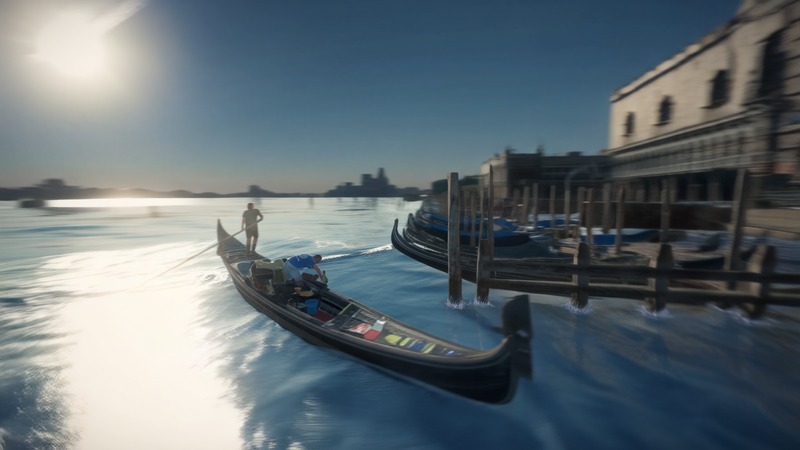}
        \end{overpic}
    \end{minipage}\par\smallskip
    \begin{minipage}{1\linewidth}
        \begin{overpic}[width=0.3279\linewidth]{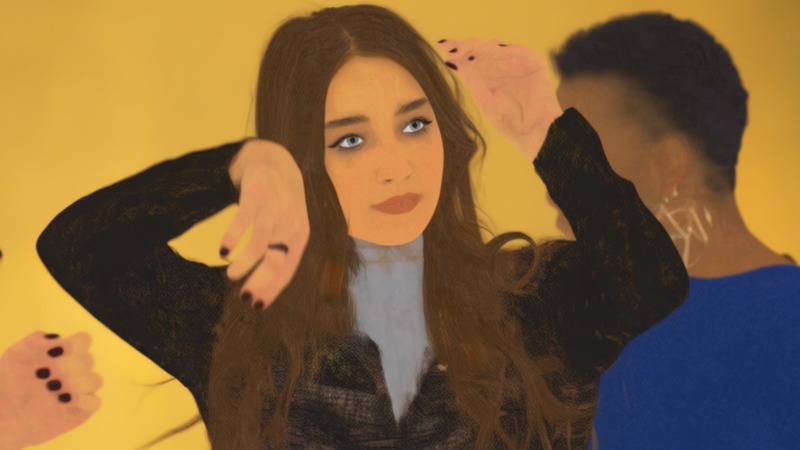}
        \put(0,2){
        \begin{tikzpicture} \node[fill=black!10,fill opacity=0.75,rounded corners=1ex, text width=0.7cm,align=left] {Albedo}; \end{tikzpicture}
        }
        \end{overpic}\hfill
        \begin{overpic}[width=0.3279\linewidth]{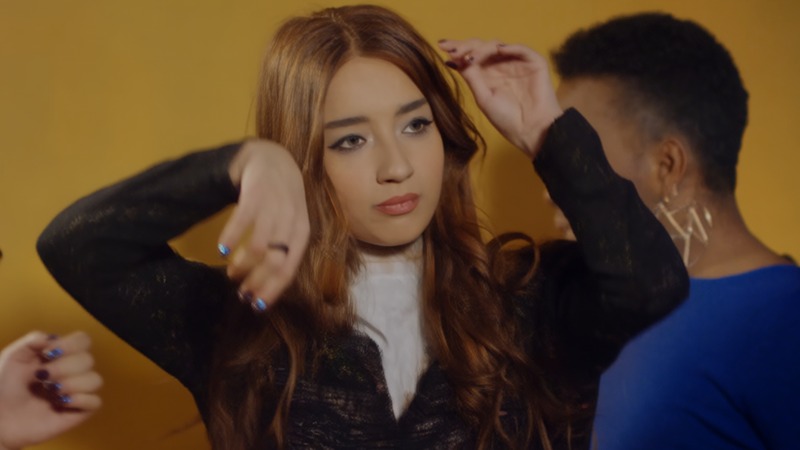}
        \end{overpic}\hfill
        \begin{overpic}[width=0.3279\linewidth]{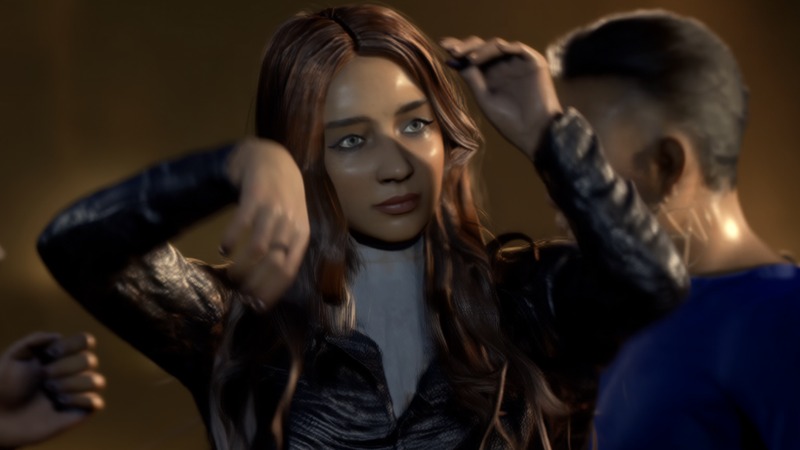}
        \end{overpic}
    \end{minipage}\par\smallskip
    \begin{minipage}{1\linewidth}
        \begin{overpic}[width=0.3279\linewidth]{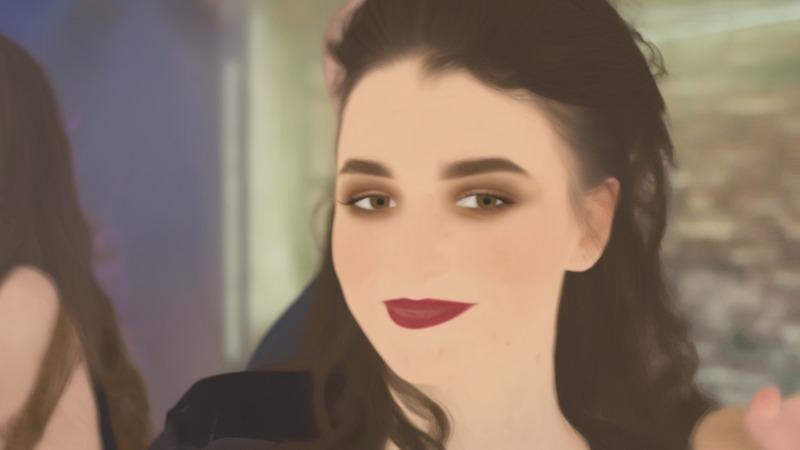}
        \put(0,2){
        \begin{tikzpicture} \node[fill=black!10,fill opacity=0.75,rounded corners=1ex, text width=0.7cm,align=left] {Albedo}; \end{tikzpicture}
        }
        \end{overpic}\hfill
        \begin{overpic}[width=0.3279\linewidth]{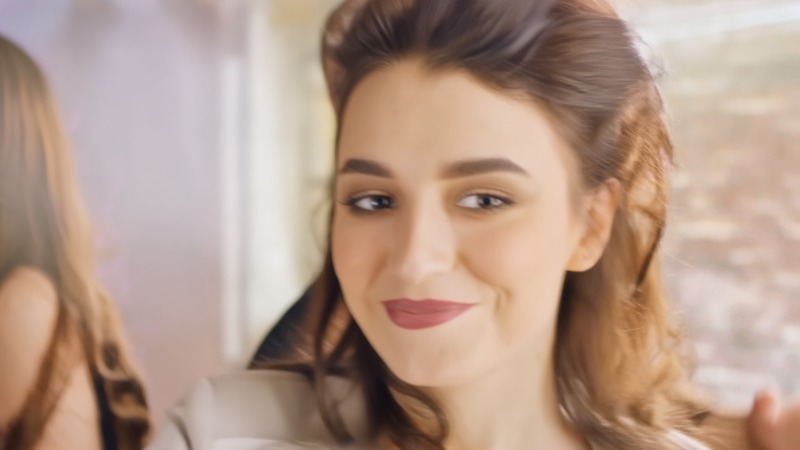}
        \end{overpic}\hfill
        \begin{overpic}[width=0.3279\linewidth]{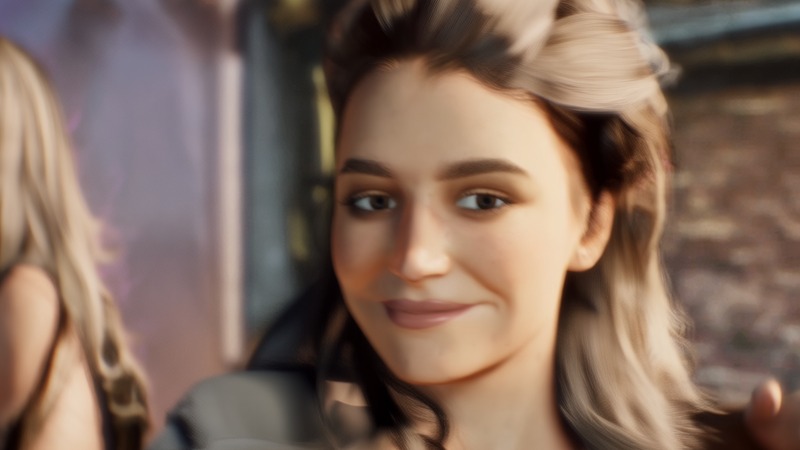}
        \end{overpic}
    \end{minipage}
    \caption{\edit{Realism control with text guidance. We show input albedo buffers (left), CFG with a positive prompt for realistic imagery and a negative prompt for synthetic imagery, pushing towards realism (middle), and the same prompts flipped, producing a more synthetic appearance (right).}}
    \label{fig:x2rgb_prompts}
\end{figure}
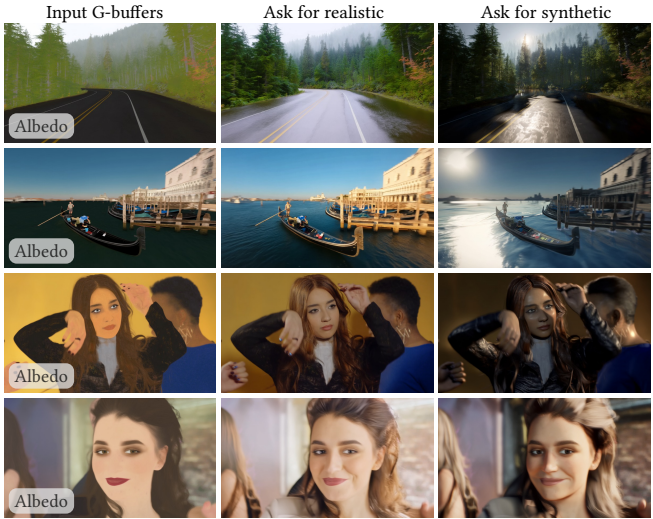

\paragraph{Realism control} \edit{In \cref{fig:x2rgb_prompts}, we show how the positive and negative prompts described in \cref{sec:xpatchify} control the realism of \xtorgb outputs. With a positive prompt asking for realistic imagery and a negative prompt describing synthetic or rendered appearance, CFG pushes the samples toward photographic lighting, textures, and camera-like image statistics. Flipping the two prompts produces the opposite effect: the geometry and colors still follow the input albedo buffers, but the generated images become more synthetic and stylized.}

\begin{figure*}[t]
    \vspace{1mm}
    \footnotesize
    \centering
    \begin{minipage}{1.0\linewidth}
        \begin{overpic}[width=0.162\linewidth]{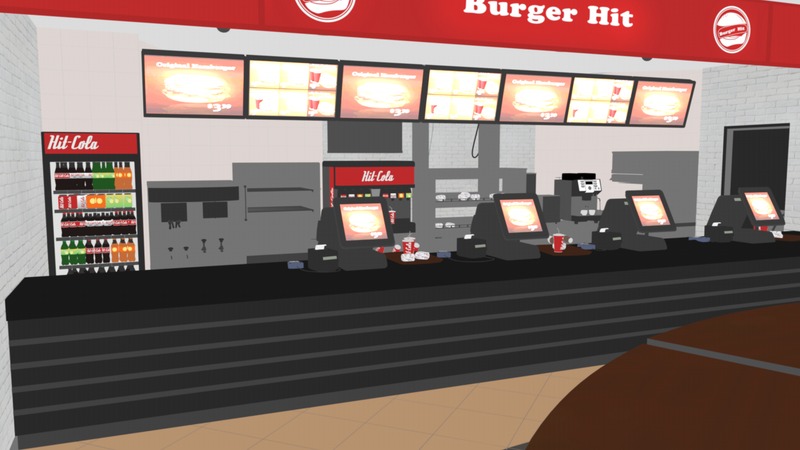}
        \put(20,58){Input G-buffers}
        \put(0,2){
        \begin{tikzpicture} \node[fill=black!10,fill opacity=0.75,rounded corners=1ex, text width=0.7cm,align=left] {Albedo}; \end{tikzpicture}
        }
        \end{overpic}\hfill
        \begin{overpic}[width=0.162\linewidth]{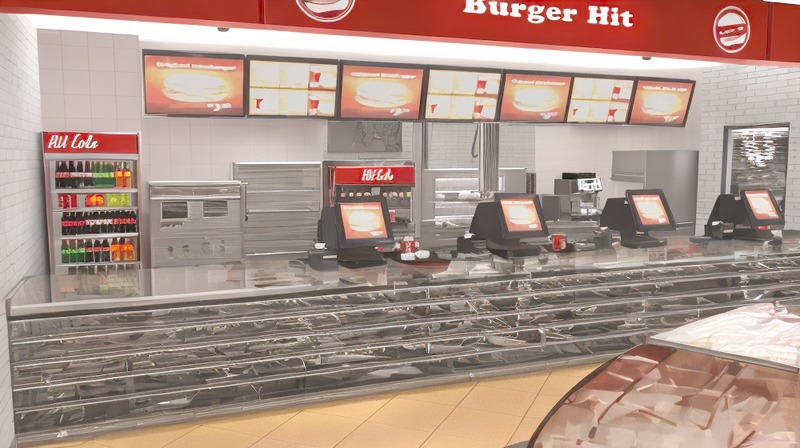}
        \put(2,58){\footnotesize \rgbx~\cite{zeng2024rgbx}}
        \end{overpic}\hfill
        \begin{overpic}[width=0.162\linewidth]{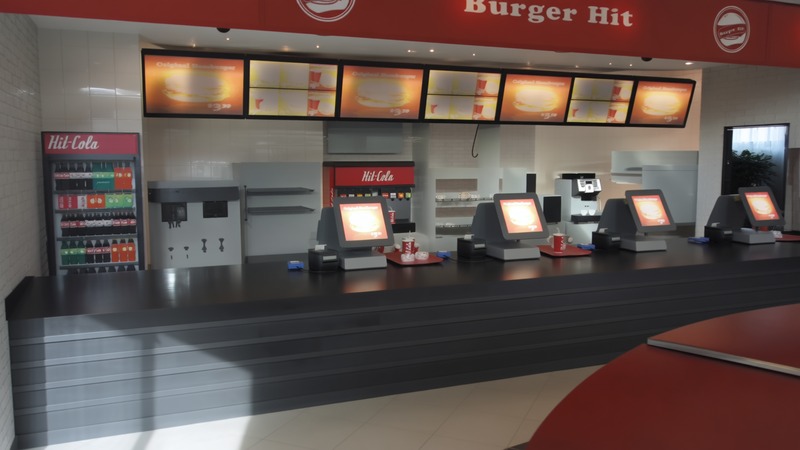}
        \put(38,58){\textbf{Ours}}
        \end{overpic}\hfill\hspace{0.25em}
        \begin{overpic}[width=0.162\linewidth]{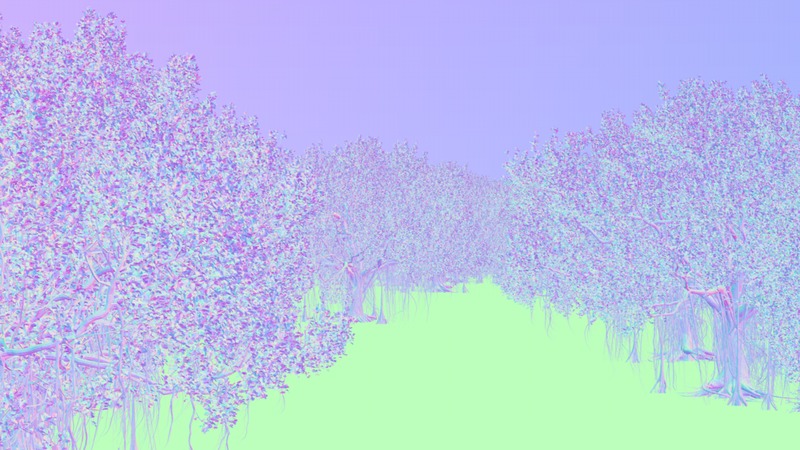}
        \put(20,58){Input G-buffers}
        \put(0,2){
        \begin{tikzpicture} \node[fill=black!10,fill opacity=0.75,rounded corners=1ex, text width=0.8cm,align=left] {Normal}; \end{tikzpicture}
        }
        \end{overpic}\hfill
        \begin{overpic}[width=0.162\linewidth]{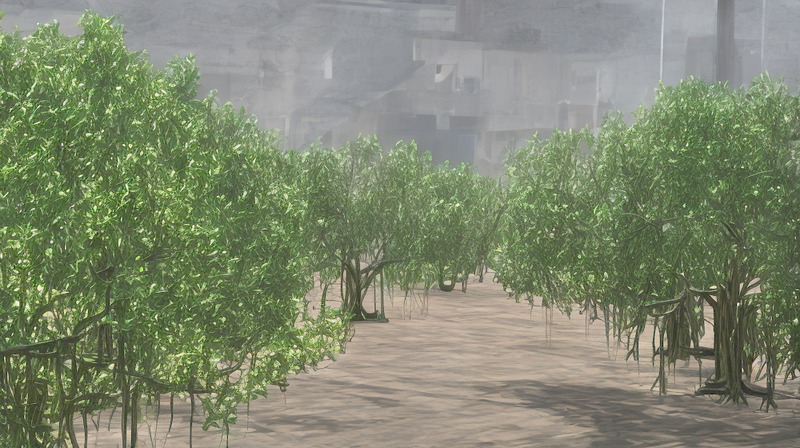}
        \put(2,58){\footnotesize \rgbx~\cite{zeng2024rgbx}}
        \end{overpic}\hfill
        \begin{overpic}[width=0.162\linewidth]{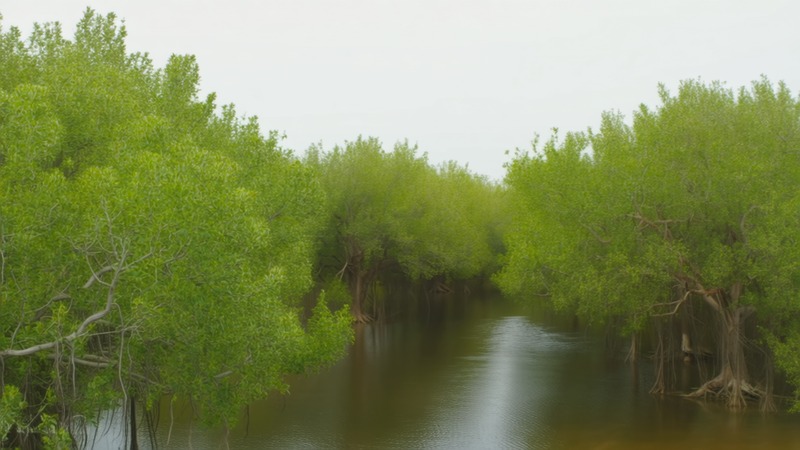}
        \put(35,58){\textbf{Ours}}
        \end{overpic}
    \end{minipage}\par\smallskip
    \begin{minipage}{1.0\linewidth}
        \begin{overpic}[width=0.162\linewidth]{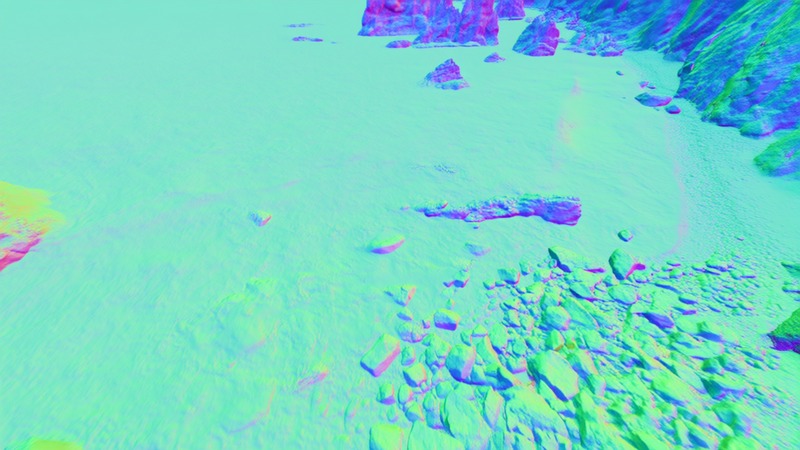}
        \put(0,2){
        \begin{tikzpicture} \node[fill=black!10,fill opacity=0.75,rounded corners=1ex, text width=0.8cm,align=left] {Normal}; \end{tikzpicture}
        }
        \end{overpic}\hfill
        \begin{overpic}[width=0.162\linewidth]{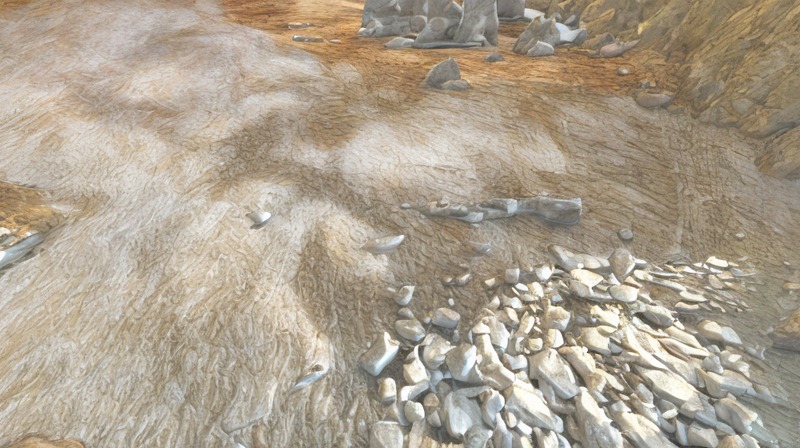}
        \end{overpic}\hfill
        \begin{overpic}[width=0.162\linewidth]{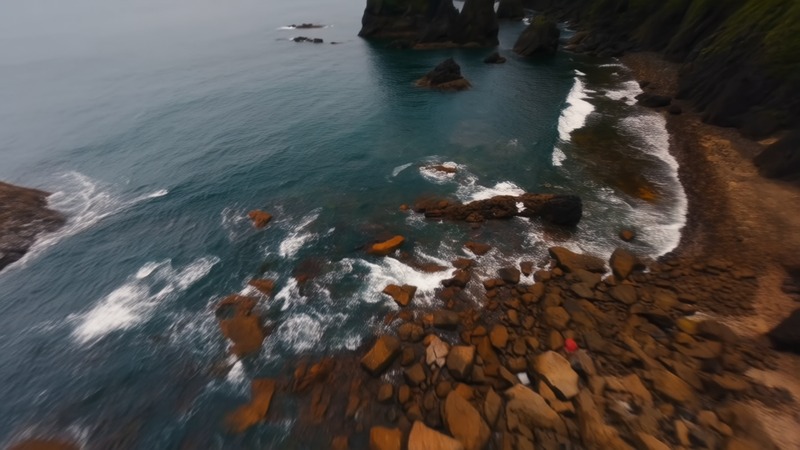}
        \end{overpic}\hfill\hspace{0.25em}
        \begin{overpic}[width=0.162\linewidth]{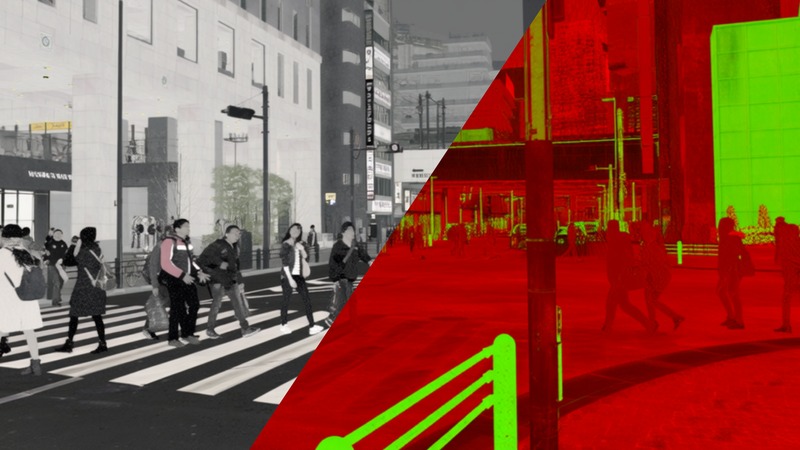}
        \put(0,2){
        \begin{tikzpicture} \node[fill=black!10,fill opacity=0.75,rounded corners=1ex, text width=1.9cm,align=left] {Albedo / Material}; \end{tikzpicture}
        }
        \end{overpic}\hfill
        \begin{overpic}[width=0.162\linewidth]{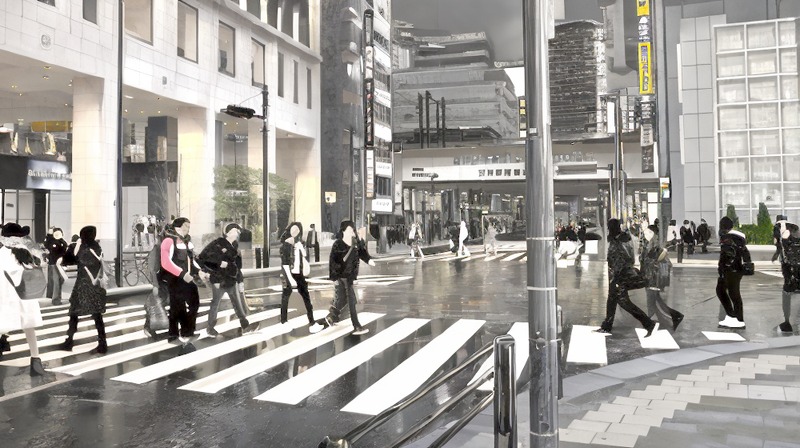}
        \end{overpic}\hfill
        \begin{overpic}[width=0.162\linewidth]{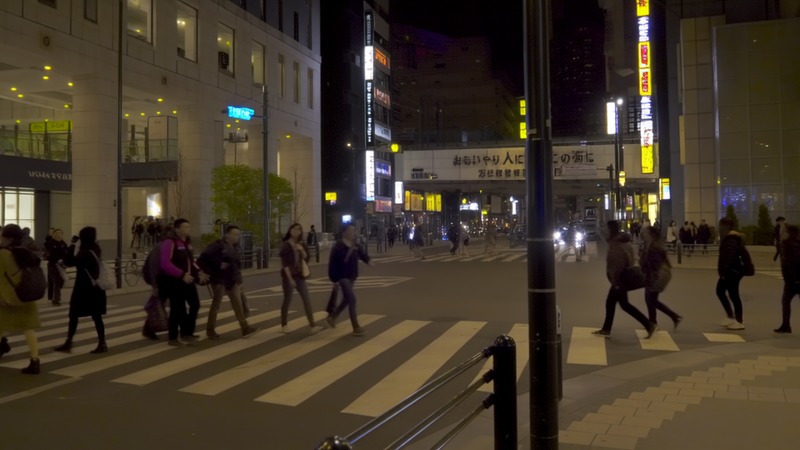}
        \end{overpic}
    \end{minipage}\par\smallskip
    \begin{minipage}{1.0\linewidth}
        \begin{overpic}[width=0.162\linewidth]{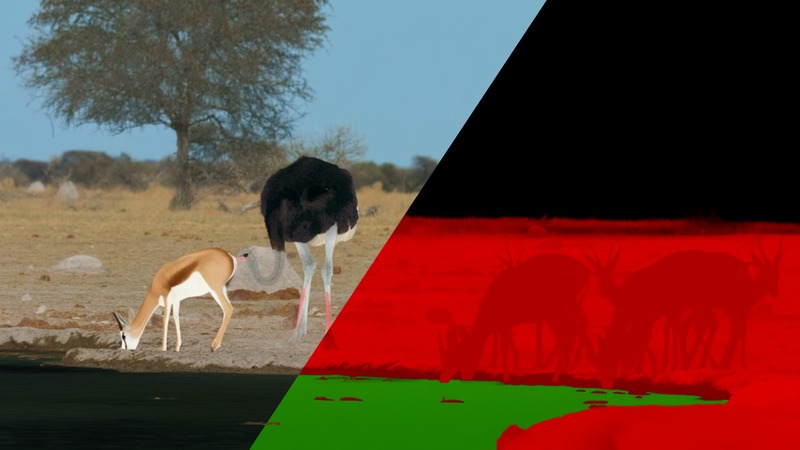}
        \put(0,2){
        \begin{tikzpicture} \node[fill=black!10,fill opacity=0.75,rounded corners=1ex, text width=1.9cm,align=left] {Albedo / Material}; \end{tikzpicture}
        }
        \end{overpic}\hfill
        \begin{overpic}[width=0.162\linewidth]{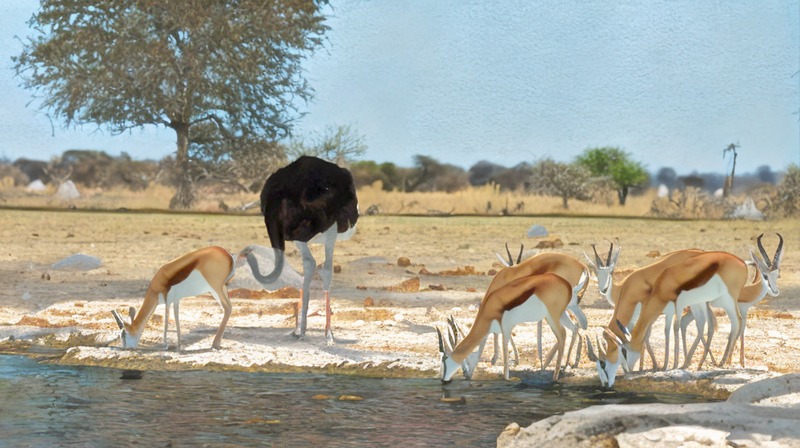}
        \end{overpic}\hfill
        \begin{overpic}[width=0.162\linewidth]{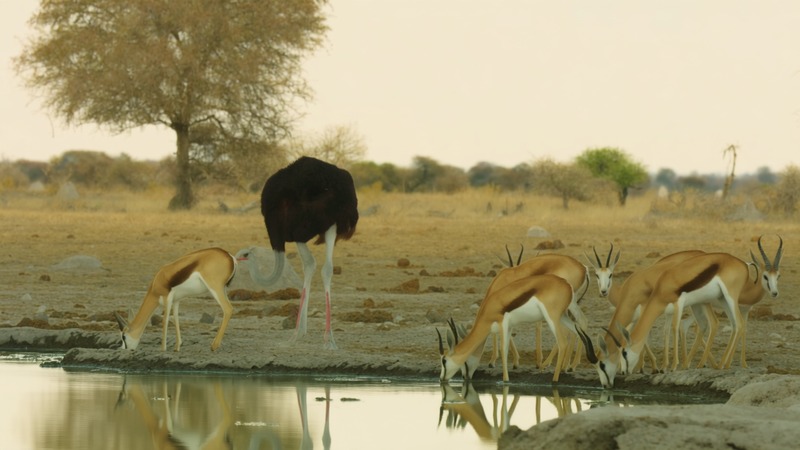}
        \end{overpic}\hfill\hspace{0.25em}
        \begin{overpic}[width=0.162\linewidth]{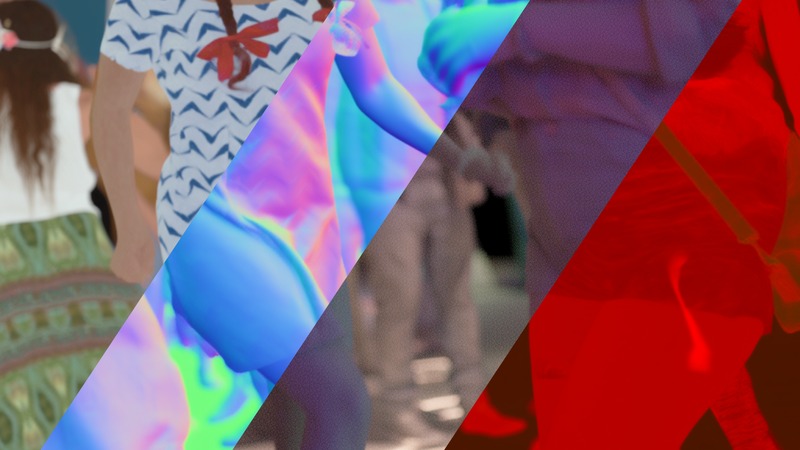}
        \put(0,2){
        \begin{tikzpicture} \node[fill=black!10,fill opacity=0.75,rounded corners=1ex, text width=2.2cm,align=left] {Albedo / Normal \\ {Irradiance / Material}}; \end{tikzpicture}
        }
        \end{overpic}\hfill
        \begin{overpic}[width=0.162\linewidth]{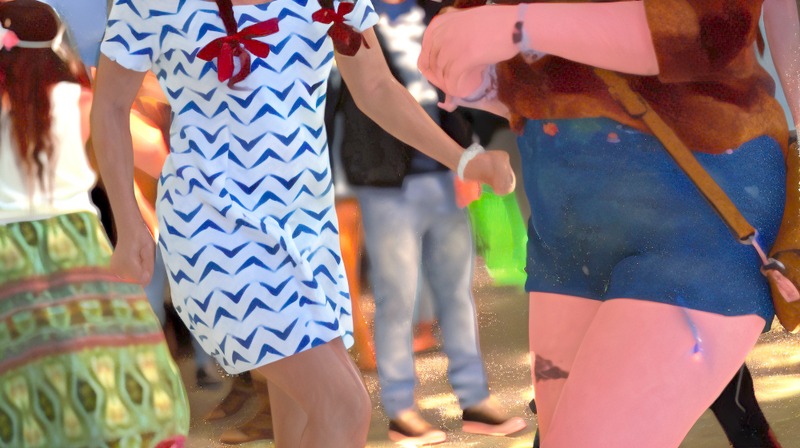}
        \end{overpic}\hfill
        \begin{overpic}[width=0.162\linewidth]{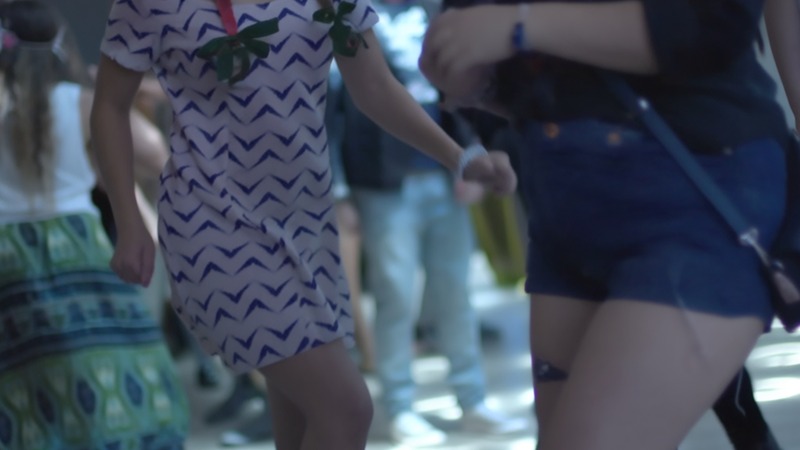}
        \end{overpic}
    \end{minipage}
    \caption{\edit{Qualitative \xtorgb comparison with \rgbx~\cite{zeng2024rgbx}. The first row uses synthetic G-buffers rendered from 3D scenes; the remaining rows use G-buffers estimated from real videos. Compared with \rgbx, our results appear less synthetic. Since \rgbx~\cite{zeng2024rgbx} is an image model, its outputs also exhibit temporal inconsistencies; please refer to the supplementary video.}}
    \label{fig:x2rgb_baseline}
\end{figure*}

\begin{figure*}[t]
    \vspace{1mm}
    \footnotesize
    \centering
    \begin{minipage}{1.0\linewidth}
        \begin{overpic}[width=0.195\linewidth]{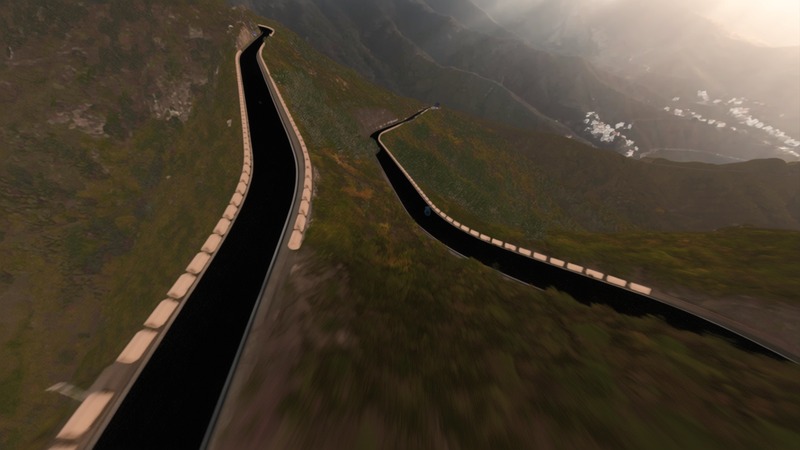}
        \put(28,59){Input Albedo}
        \end{overpic}\hfill
        \begin{overpic}[width=0.195\linewidth]{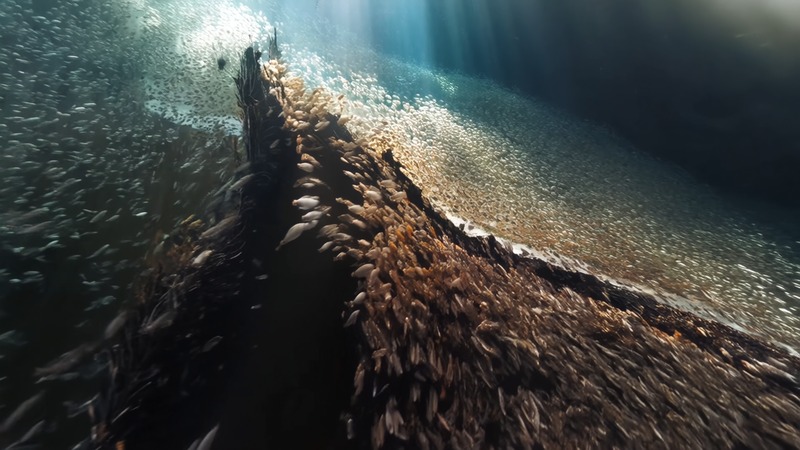}
        \put(16,59){Channel-wise Concat.}
        \end{overpic}\hfill
        \begin{overpic}[width=0.195\linewidth]{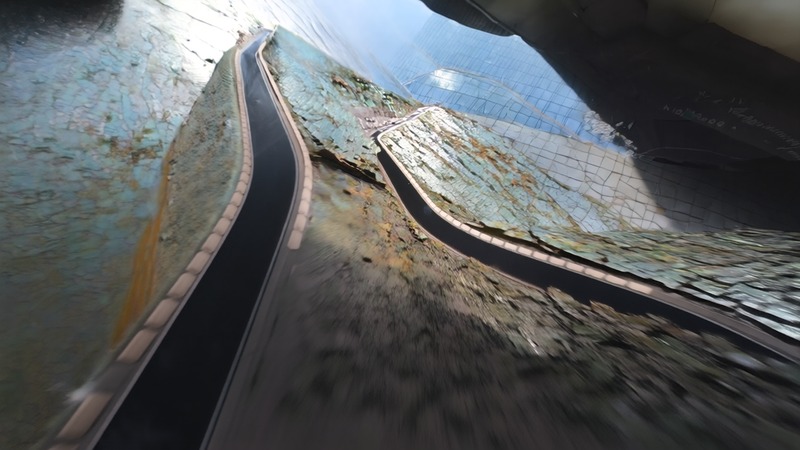}
        \put(15,59){VACE~\cite{jiang2025vace}}
        \end{overpic}\hfill
        \begin{overpic}[width=0.195\linewidth]{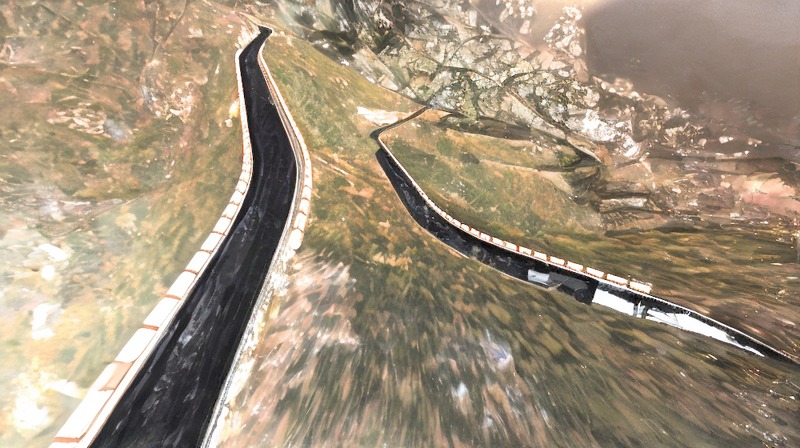}
        \put(15,59){\rgbx~\cite{zeng2024rgbx}}
        \end{overpic}\hfill
        \begin{overpic}[width=0.195\linewidth]{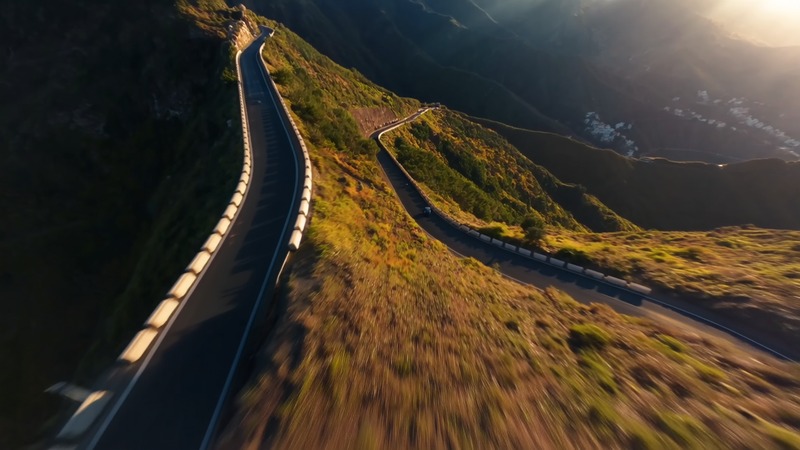}
        \put(39,59){\textbf{Ours}}
        \end{overpic}
    \end{minipage}\par
    \begin{minipage}{1.0\linewidth}
        \begin{overpic}[width=0.195\linewidth]{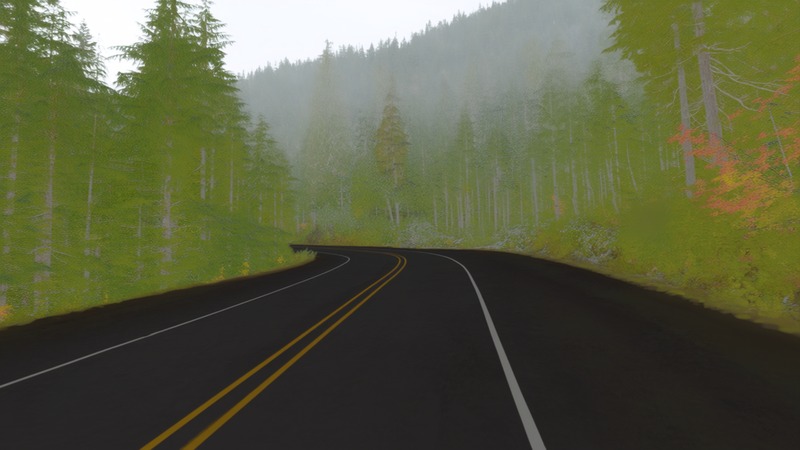}
        \end{overpic}\hfill
        \begin{overpic}[width=0.195\linewidth]{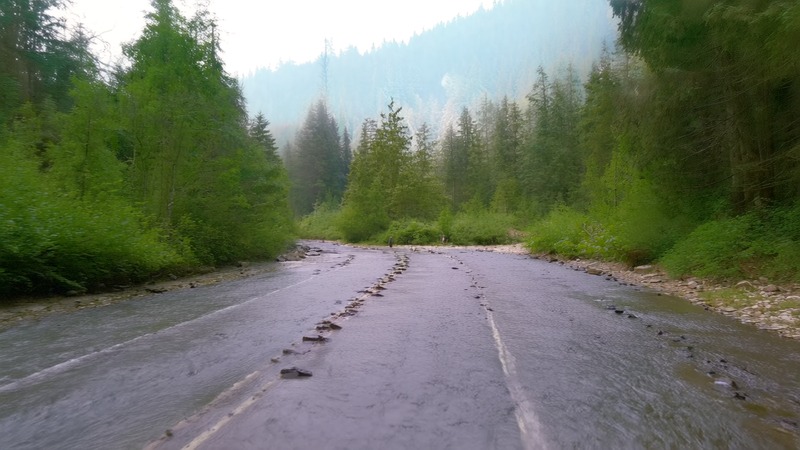}
        \end{overpic}\hfill
        \begin{overpic}[width=0.195\linewidth]{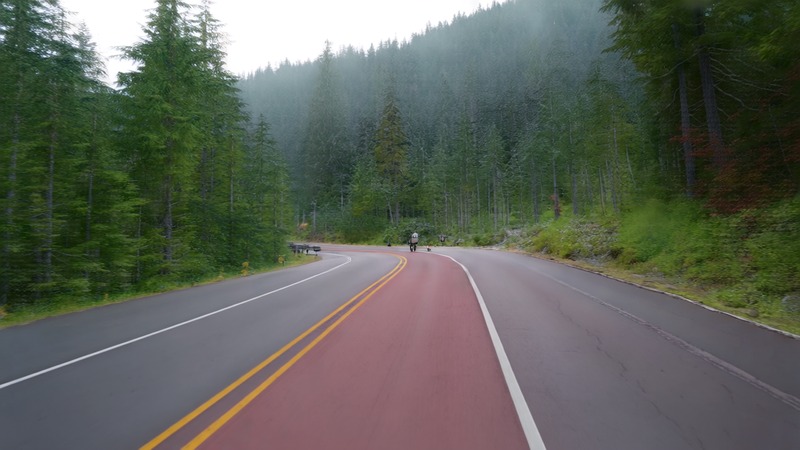}
        \end{overpic}\hfill
        \begin{overpic}[width=0.195\linewidth]{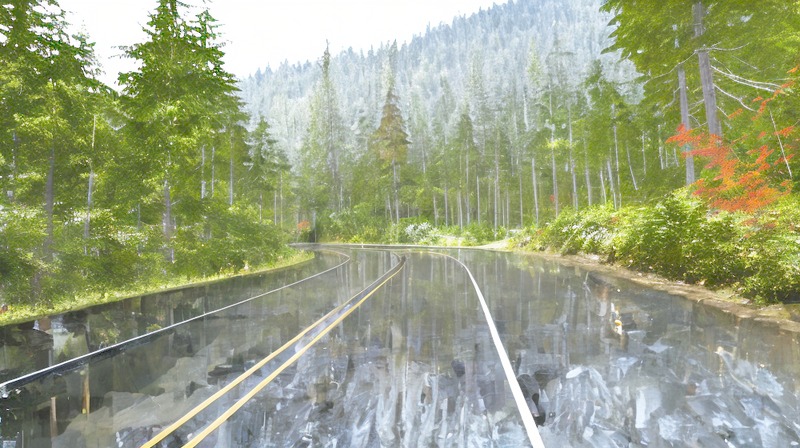}
        \end{overpic}\hfill
        \begin{overpic}[width=0.195\linewidth]{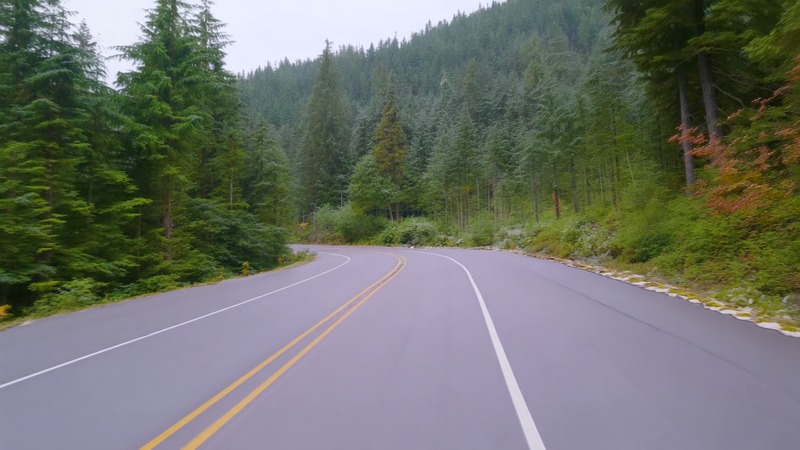}
        \end{overpic}
    \end{minipage}\par
    \begin{minipage}{1.0\linewidth}
        \begin{overpic}[width=0.195\linewidth]{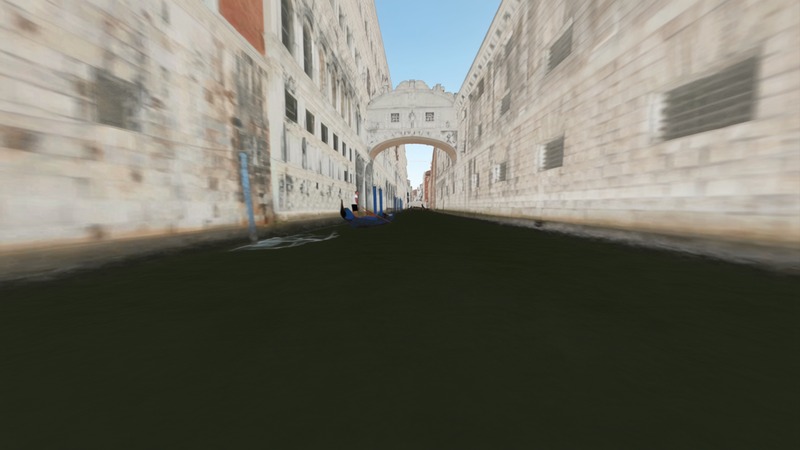}
        \end{overpic}\hfill
        \begin{overpic}[width=0.195\linewidth]{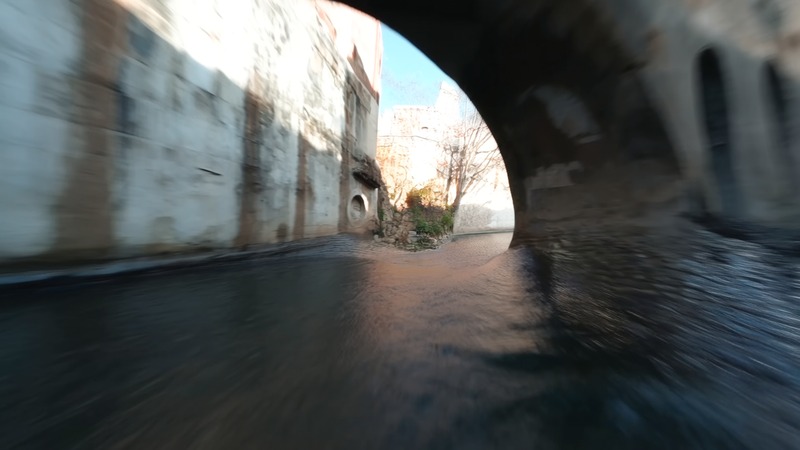}
        \end{overpic}\hfill
        \begin{overpic}[width=0.195\linewidth]{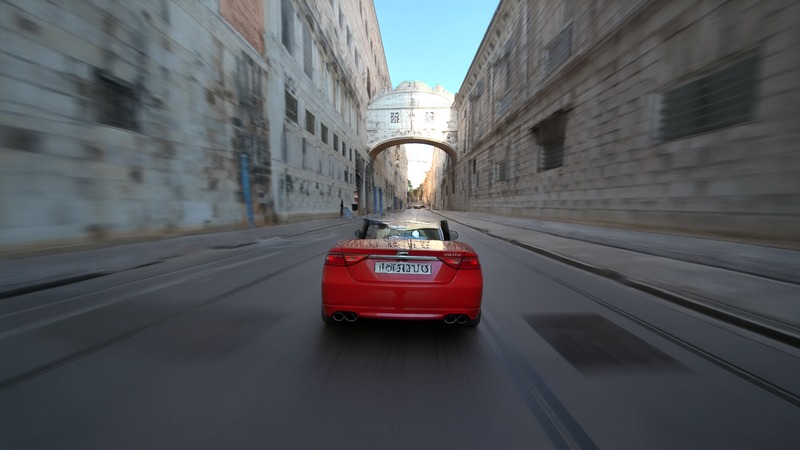}
        \end{overpic}\hfill
        \begin{overpic}[width=0.195\linewidth]{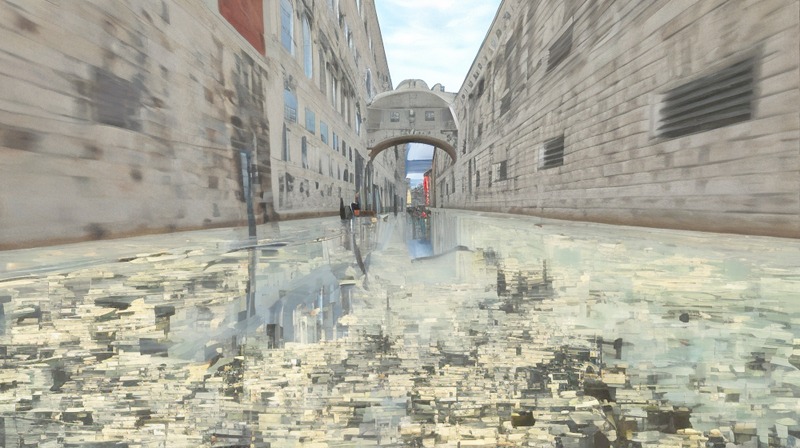}
        \end{overpic}\hfill
        \begin{overpic}[width=0.195\linewidth]{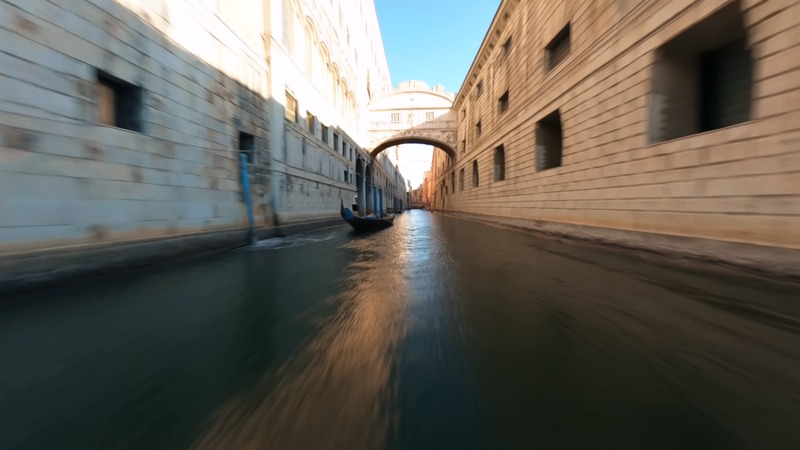}
        \end{overpic}
    \end{minipage}\par
    \begin{minipage}{1.0\linewidth}
        \begin{overpic}[width=0.195\linewidth]{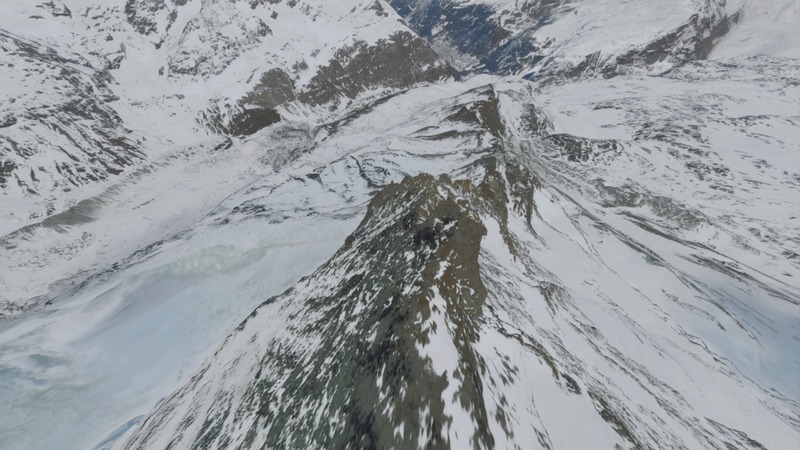}
        \end{overpic}\hfill
        \begin{overpic}[width=0.195\linewidth]{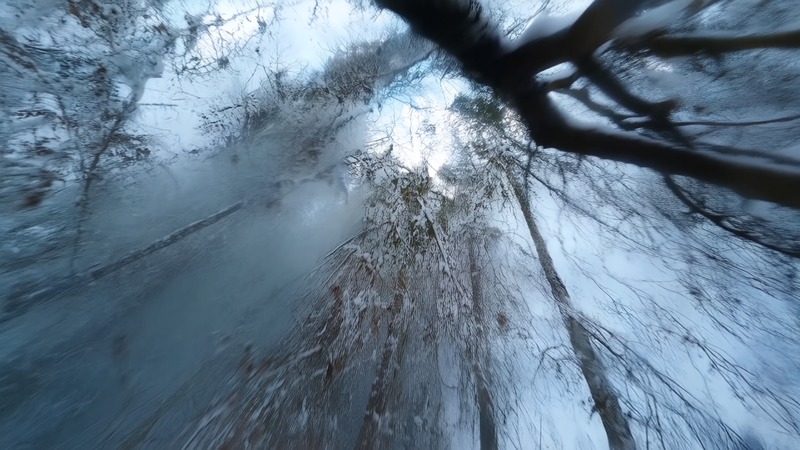}
        \end{overpic}\hfill
        \begin{overpic}[width=0.195\linewidth]{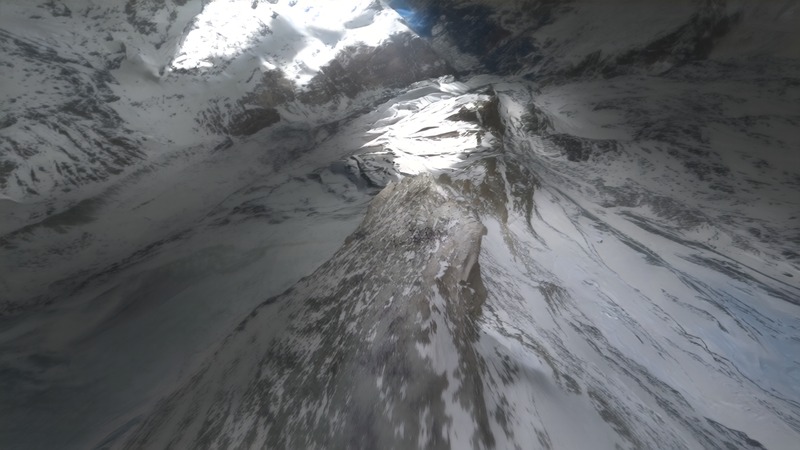}
        \end{overpic}\hfill
        \begin{overpic}[width=0.195\linewidth]{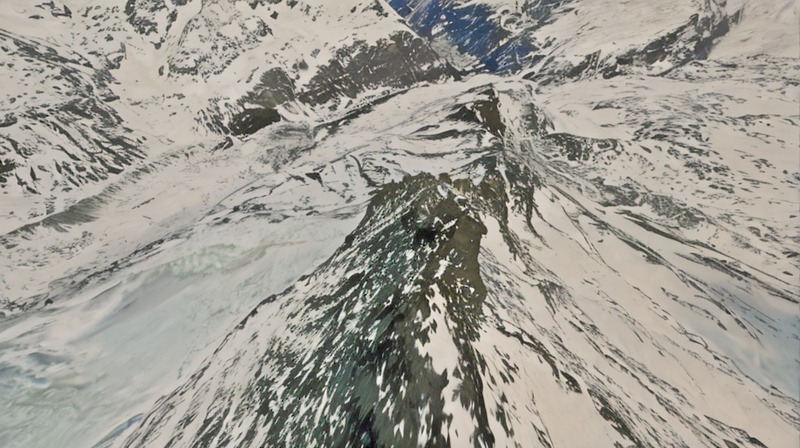}
        \end{overpic}\hfill
        \begin{overpic}[width=0.195\linewidth]{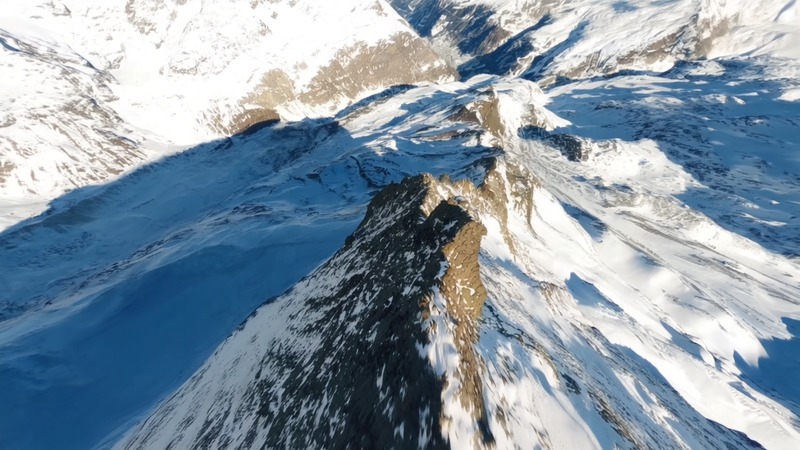}
        \end{overpic}
    \end{minipage}\par
    \begin{minipage}{1.0\linewidth}
        \begin{overpic}[width=0.195\linewidth]{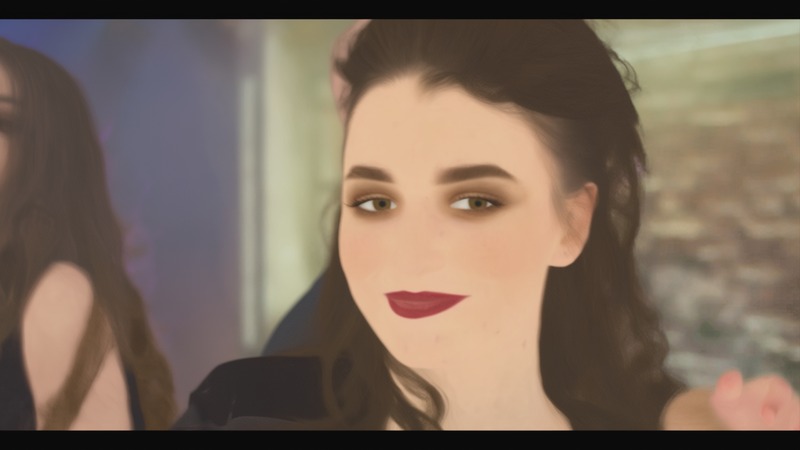}
        \end{overpic}\hfill
        \begin{overpic}[width=0.195\linewidth]{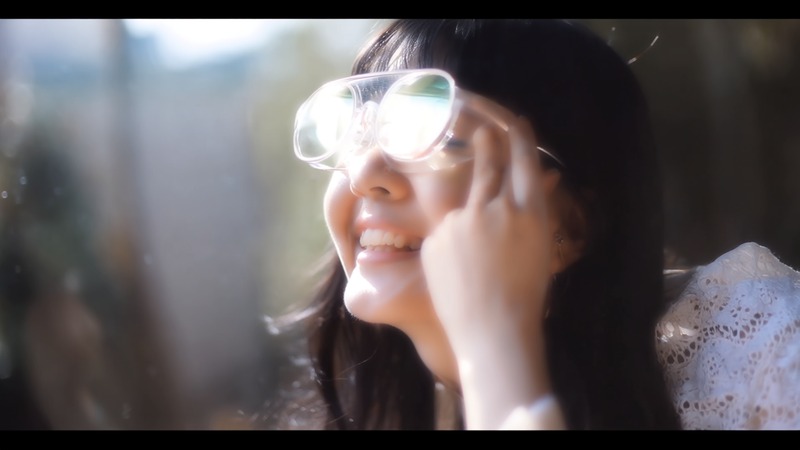}
        \end{overpic}\hfill
        \begin{overpic}[width=0.195\linewidth]{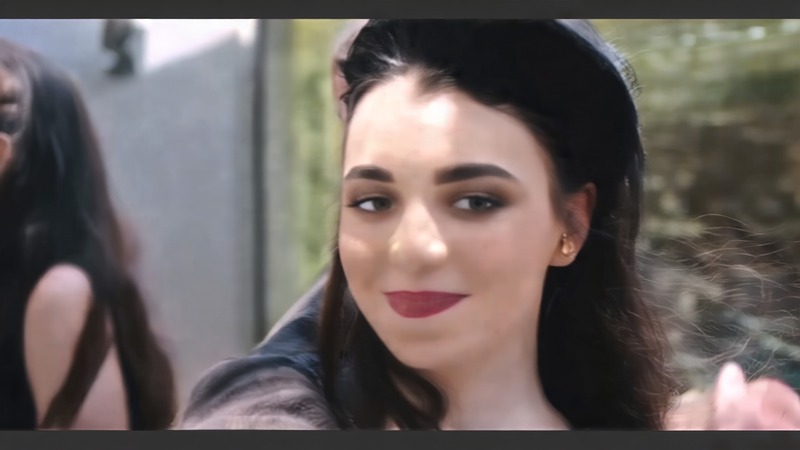}
        \end{overpic}\hfill
        \begin{overpic}[width=0.195\linewidth]{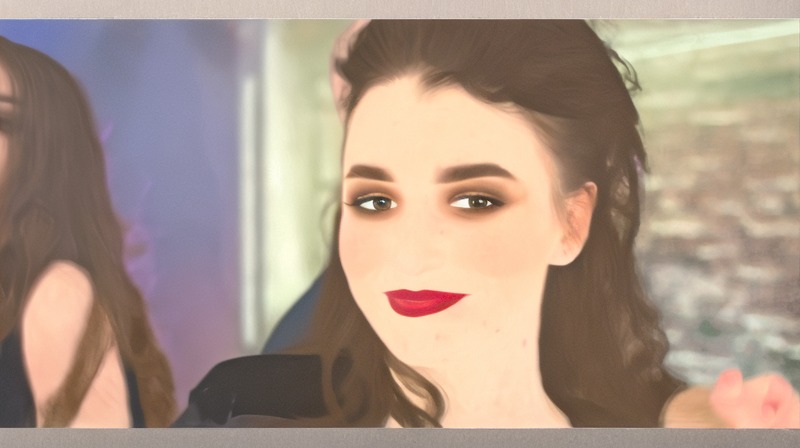}
        \end{overpic}\hfill
        \begin{overpic}[width=0.195\linewidth]{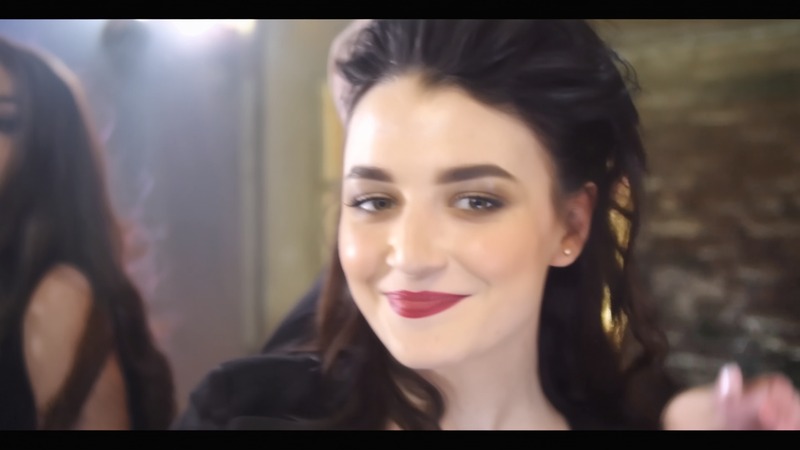}
        \end{overpic}
    \end{minipage}\par
    \begin{minipage}{1.0\linewidth}
        \begin{overpic}[width=0.195\linewidth]{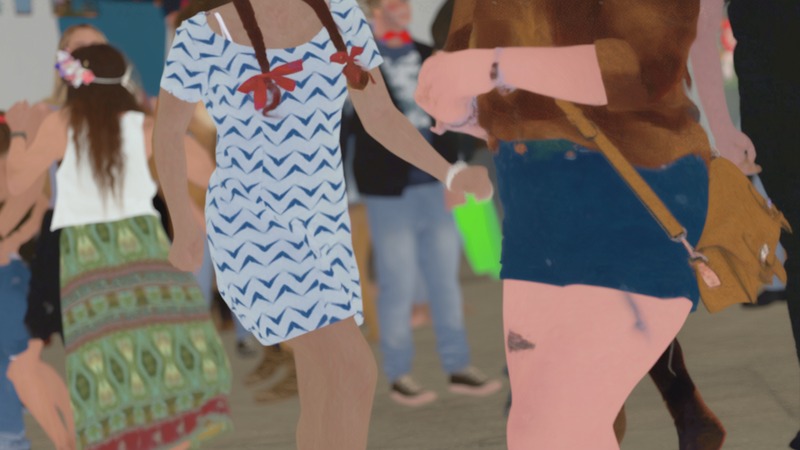}
        \end{overpic}\hfill
        \begin{overpic}[width=0.195\linewidth]{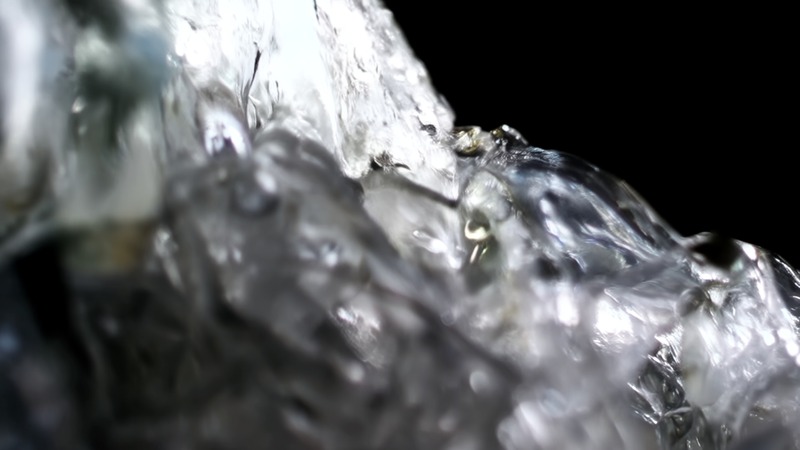}
        \end{overpic}\hfill
        \begin{overpic}[width=0.195\linewidth]{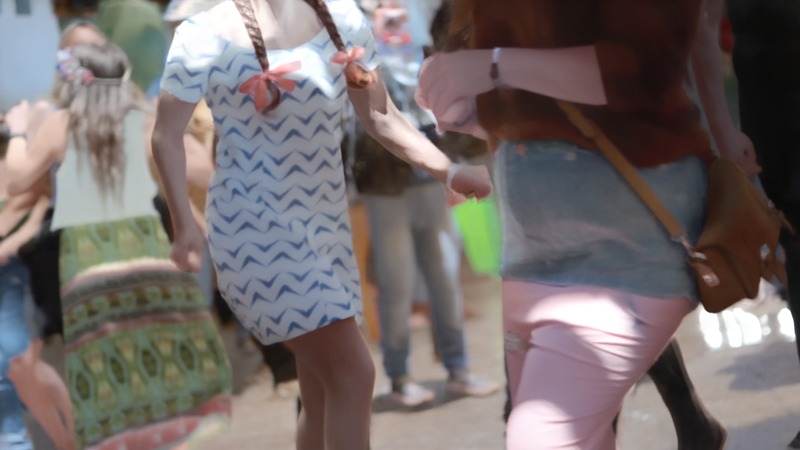}
        \end{overpic}\hfill
        \begin{overpic}[width=0.195\linewidth]{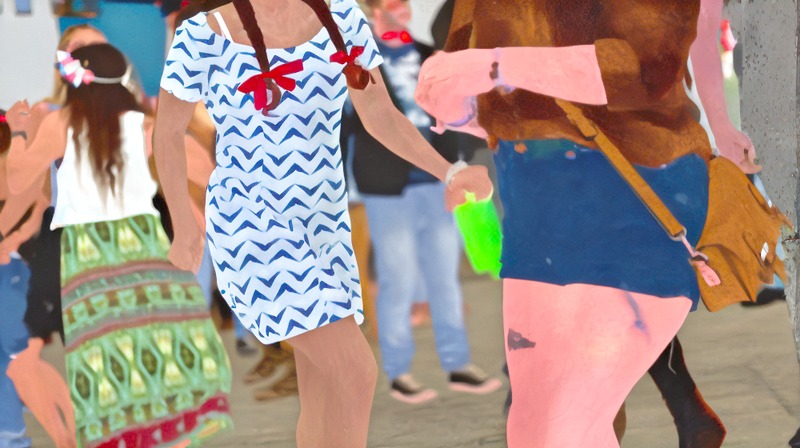}
        \end{overpic}\hfill
        \begin{overpic}[width=0.195\linewidth]{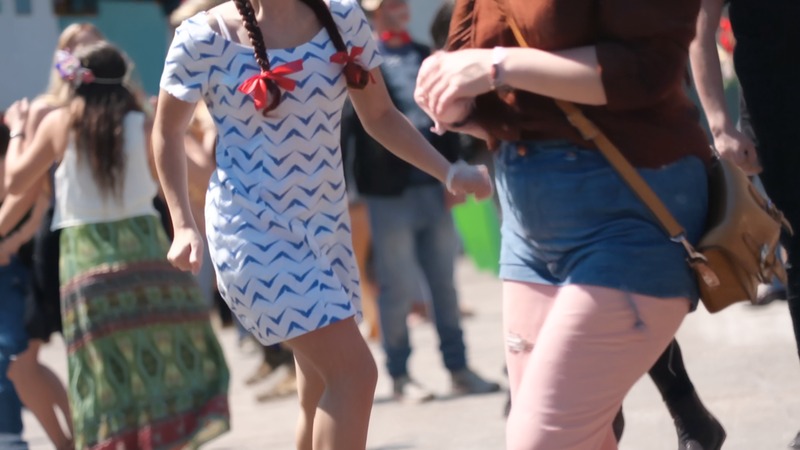}
        \end{overpic}
    \end{minipage}\par
    \begin{minipage}{1.0\linewidth}
        \begin{overpic}[width=0.195\linewidth]{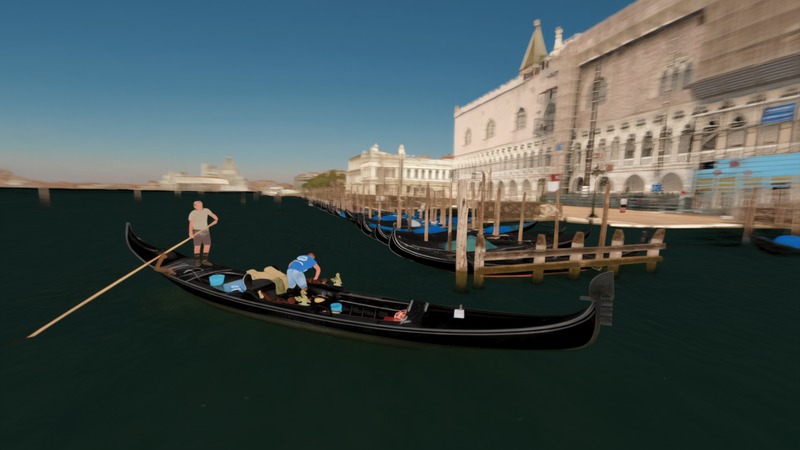}
        \end{overpic}\hfill
        \begin{overpic}[width=0.195\linewidth]{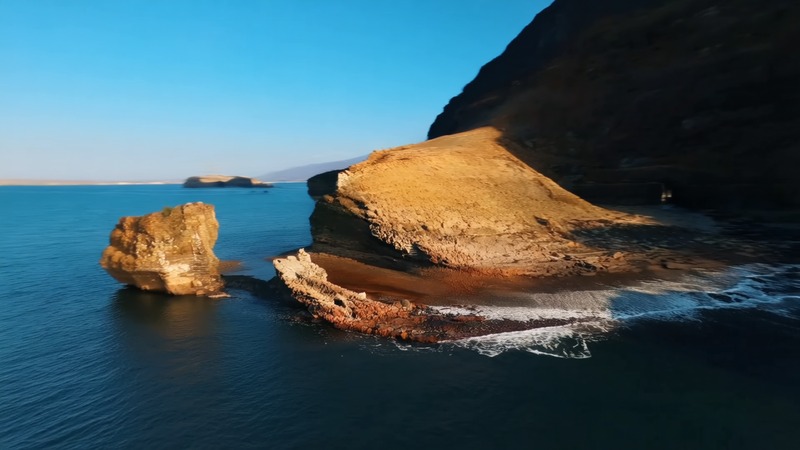}
        \end{overpic}\hfill
        \begin{overpic}[width=0.195\linewidth]{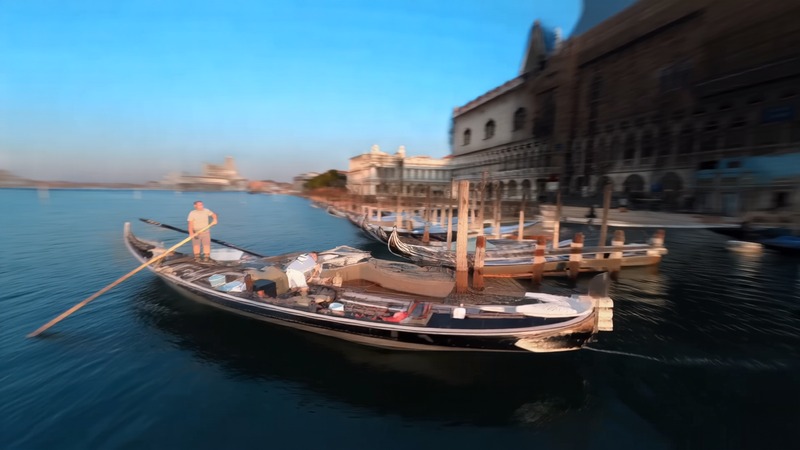}
        \end{overpic}\hfill
        \begin{overpic}[width=0.195\linewidth]{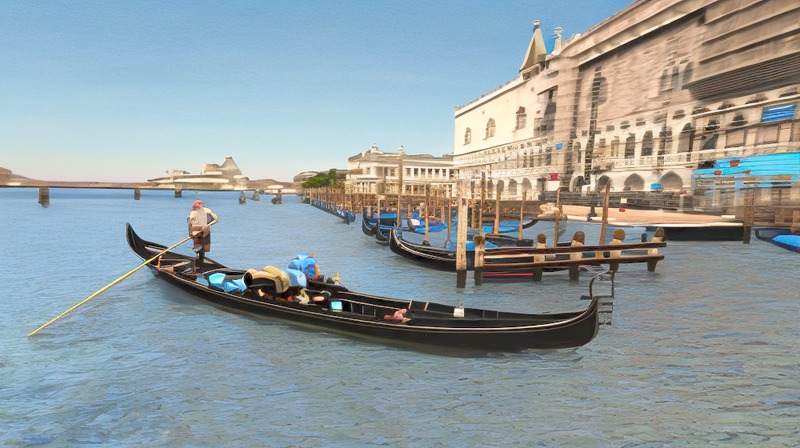}
        \end{overpic}\hfill
        \begin{overpic}[width=0.195\linewidth]{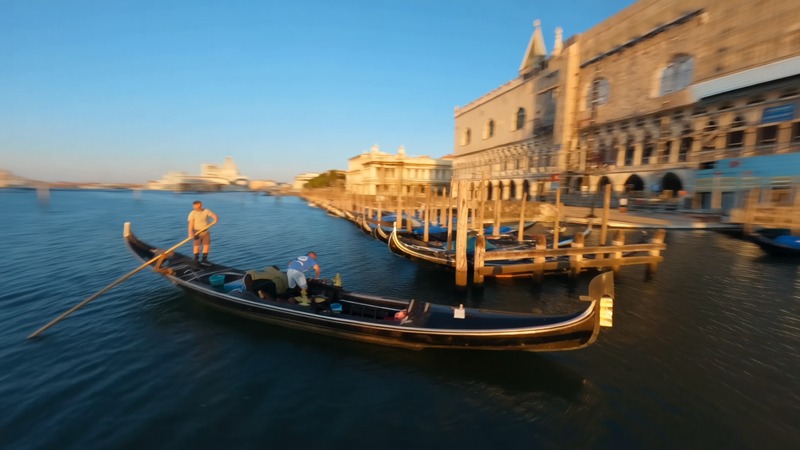}
        \end{overpic}
    \end{minipage}
    \caption{\edit{Qualitative \xtorgb comparison with albedo conditioning against \rgbx~\cite{zeng2024rgbx}, VACE~\cite{jiang2025vace}, and a channel-wise concatenation baseline, which follows the conditioning mechanism used by DiffusionRenderer~\cite{liang2025diffusionrenderer}. All methods except \rgbx are trained on our \textsf{(RGB, X)} dataset from the same base model for the same training time. VACE and channel-wise models produce broadly realistic results but do not follow the guidance closely, while \rgbx shows less realism due to being trained on exclusively indoor scene content.}}
    \label{fig:x2rgb_baseline2}
\end{figure*}

\paragraph{Comparisons} \edit{In \cref{fig:x2rgb_baseline}, we compare our generalized \xtorgb model, trained to condition on albedo, normal, irradiance, and material inputs with modality dropout, with \citet{zeng2024rgbx}. Results in the first row use synthetic G-buffers rendered from 3D scenes, while the other rows use estimated G-buffers from real videos. Outputs from \rgbx \cite{zeng2024rgbx} often appear unrealistic and flat compared to ours. We attribute this improvement to the stronger DiT video backbone and to training on realistic videos annotated with estimated G-buffers, rather than mixing synthetic and real RGB supervision. Further comparisons are shown in the supplementary video; since \rgbx\cite{zeng2024rgbx} is an image model, its outputs also exhibit temporal inconsistencies. In \cref{fig:x2rgb_baseline2}, we further compare \xtorgb with albedo conditioning against \rgbx~\cite{zeng2024rgbx}, VACE~\cite{jiang2025vace}, and a channel-wise concatenation baseline. A direct \xtorgb comparison with DiffusionRenderer~\cite{liang2025diffusionrenderer} is not straightforward because DiffusionRenderer requires all G-buffers to be fed to the model, including a static image-based lighting (IBL) input, and it is unclear whether the IBL is defined around the camera or over the entire scene. Since DiffusionRenderer uses channel-wise concatenation for conditioning, we instead include a channel-wise concatenation baseline trained on our data, in addition to VACE and \rgbx \cite{zeng2024rgbx}. All methods except \rgbx \cite{zeng2024rgbx} are trained on our (RGB, X) dataset from the same base model for the same duration. Our model produces the most realistic results while following the input albedo. VACE sometimes fails to follow the input albedo and produces less realistic results. \rgbx \cite{zeng2024rgbx} outputs are unrealistic and flat. Channel-wise concatenation can produce realistic results from the base model's prior, but it often fails to follow the albedo.}

\begin{table}
\centering
\begin{tabular}{l|c}
\bf Method & \bf FID $\downarrow$ \\
  \hline
Channel-wise concat. & 88.6039 \\
VACE~\cite{jiang2025vace} & 62.2883 \\
\rgbx~\cite{zeng2024rgbx} & 62.3884 \\
Ours & \bf 45.3871 \\
\end{tabular}
\caption{\edit{FID comparison for \xtorgb on the held-out \textsf{(RGB, X)} test set. We compute FID between generated frames and real video frames from the test set. Lower is better.}}
\label{tab:x2rgb_fid}
\end{table}

\paragraph{Quantitative comparisons} \edit{In \cref{tab:x2rgb_fid}, we quantitatively compare \xtorgb with albedo conditioning on the held-out \textsf{(RGB, X)} test set. We run each model on the same test split, which contains 6307 real video frames, and compute Fr\'echet Inception Distance (FID) \cite{heusel2017gans} against the real-frame distribution of the test set. Our model achieves the lowest FID, improving substantially over VACE, \rgbx, and the channel-wise concatenation baseline, which is consistent with the qualitative realism differences shown in \cref{fig:x2rgb_baseline2}.}

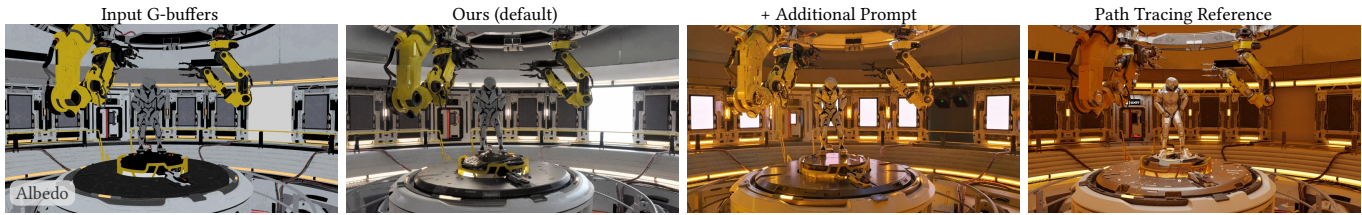
\begin{figure*}[t]
    \vspace{1mm}
    \footnotesize
    \centering
    \begin{minipage}{1.0\linewidth}
        \begin{overpic}[width=0.245\linewidth]{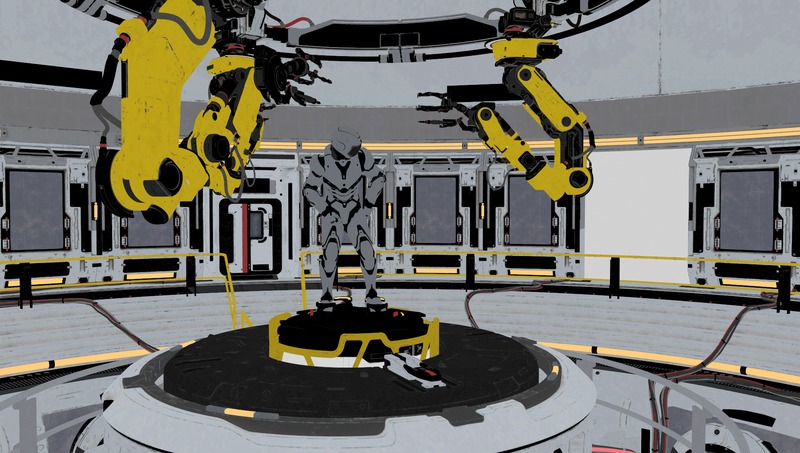}
        \put(29,58){Input G-buffers}
        \put(0,2){
        \begin{tikzpicture} \node[fill=black!10,fill opacity=0.75,rounded corners=1ex, text width=0.7cm,align=left] {Albedo}; \end{tikzpicture}
        }
        \end{overpic}\hfill
        \begin{overpic}[width=0.245\linewidth]{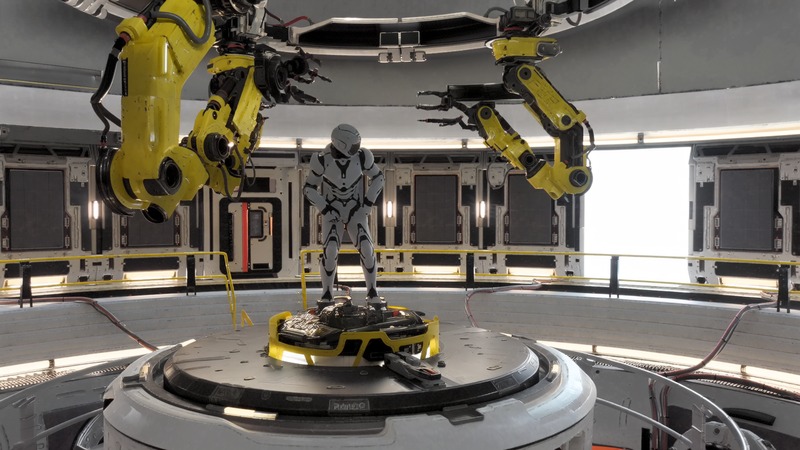}
        \put(32,58){Ours (default)}
        \end{overpic}\hfill
        \begin{overpic}[width=0.245\linewidth]{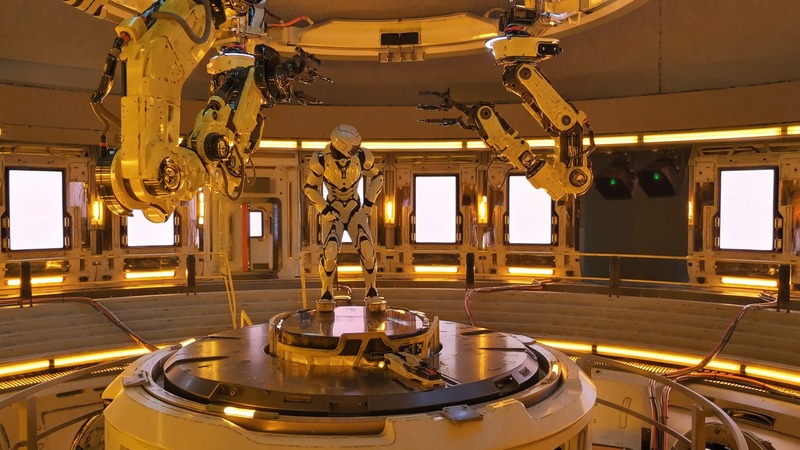}
        \put(22,58){+ Additional Prompt}
        \end{overpic}\hfill
        \begin{overpic}[width=0.245\linewidth]{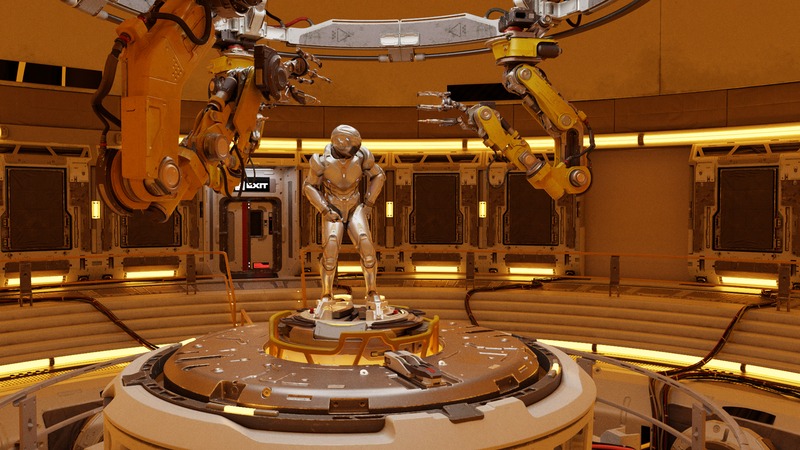}
        \put(20,58){Path Tracing Reference}
        \end{overpic}
    \end{minipage}
    \caption{\edit{Additional text prompts can control lighting and material appearance without additional G-buffer inputs. With the default prompt (``Photograph, high-resolution, 4K, high-quality''), the result follows the albedo tone but misses the warm ambient light from the yellow strip lights. Appending ``yellow strip lights create warm ambient light, metal dancing robot at center'' better matches the path-traced reference and shifts the robot from diffuse to glossy.}}
    \label{fig:x2rgb_addprompt}
\end{figure*}

\paragraph{Additional prompt control} \edit{In \cref{fig:x2rgb_addprompt}, we show that text prompts can further control lighting and material appearance even when the G-buffer input is fixed. With the default prompt, our albedo-conditioned model follows the albedo tone but misses the warm ambient lighting from the yellow strip lights, because the albedo buffer does not encode emissive surfaces. Adding a scene-specific prompt, ``yellow strip lights create warm ambient light, metal dancing robot at center'', produces warmer illumination and a glossier robot, closely matching the path-traced reference.}

\begin{figure}[th]
    \setlength{\tabcolsep}{1pt}
    \centering
        \begin{tabular}{ccc}
            \footnotesize Input albedo & \footnotesize W/o segment dropout & \footnotesize With segment dropout \\[-1pt]
            \includegraphics[width=0.326\linewidth]{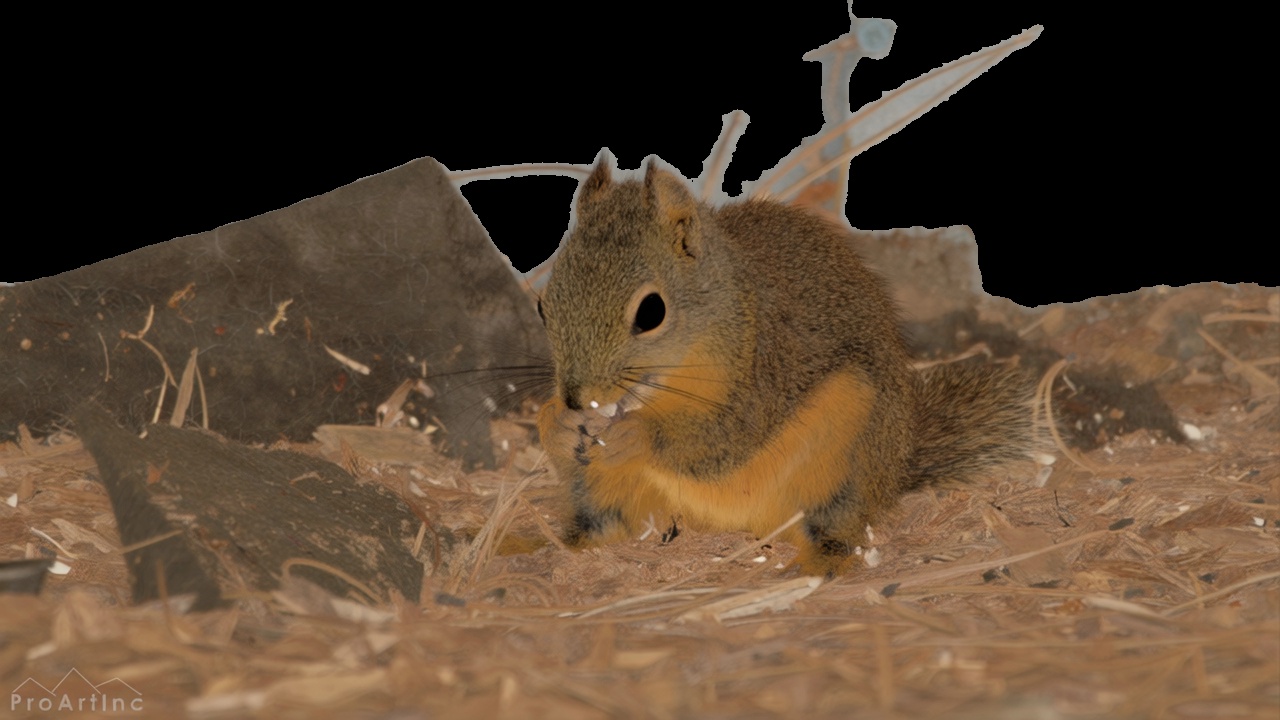} &
            \includegraphics[width=0.326\linewidth]{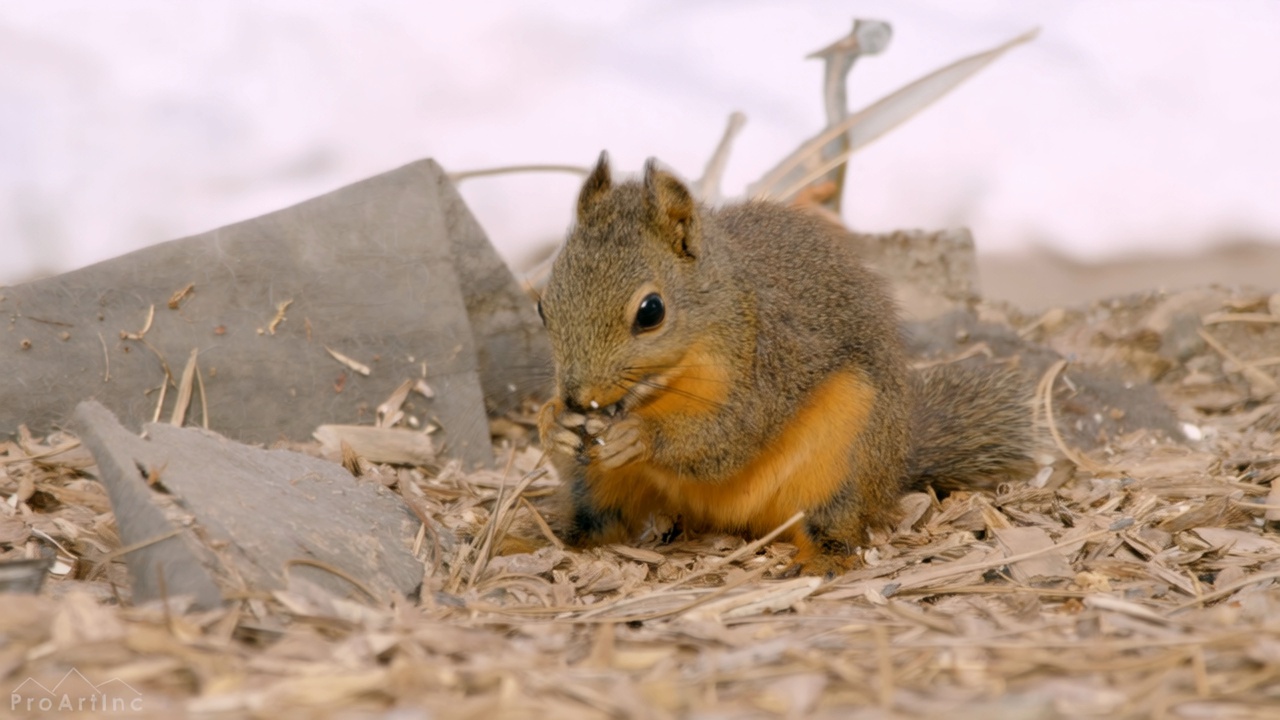} &
            \includegraphics[width=0.326\linewidth]{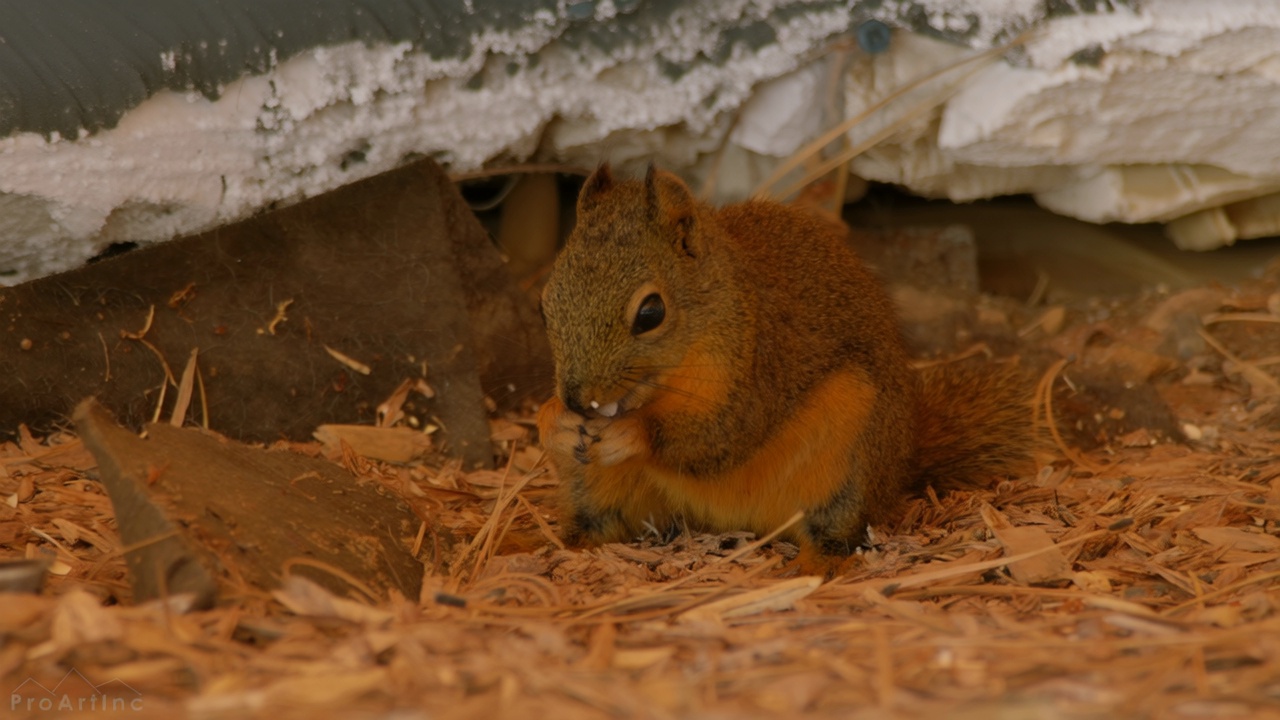} \\[-1pt]
            \includegraphics[width=0.326\linewidth]{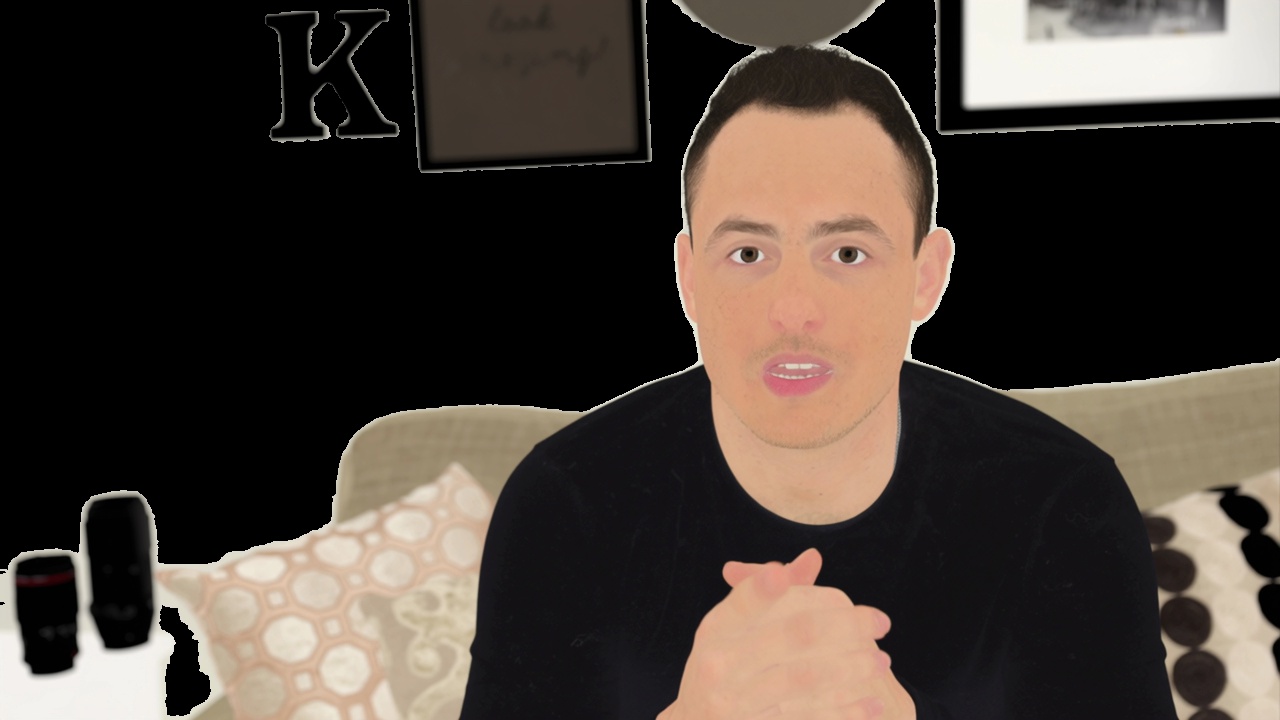} &
            \includegraphics[width=0.326\linewidth]{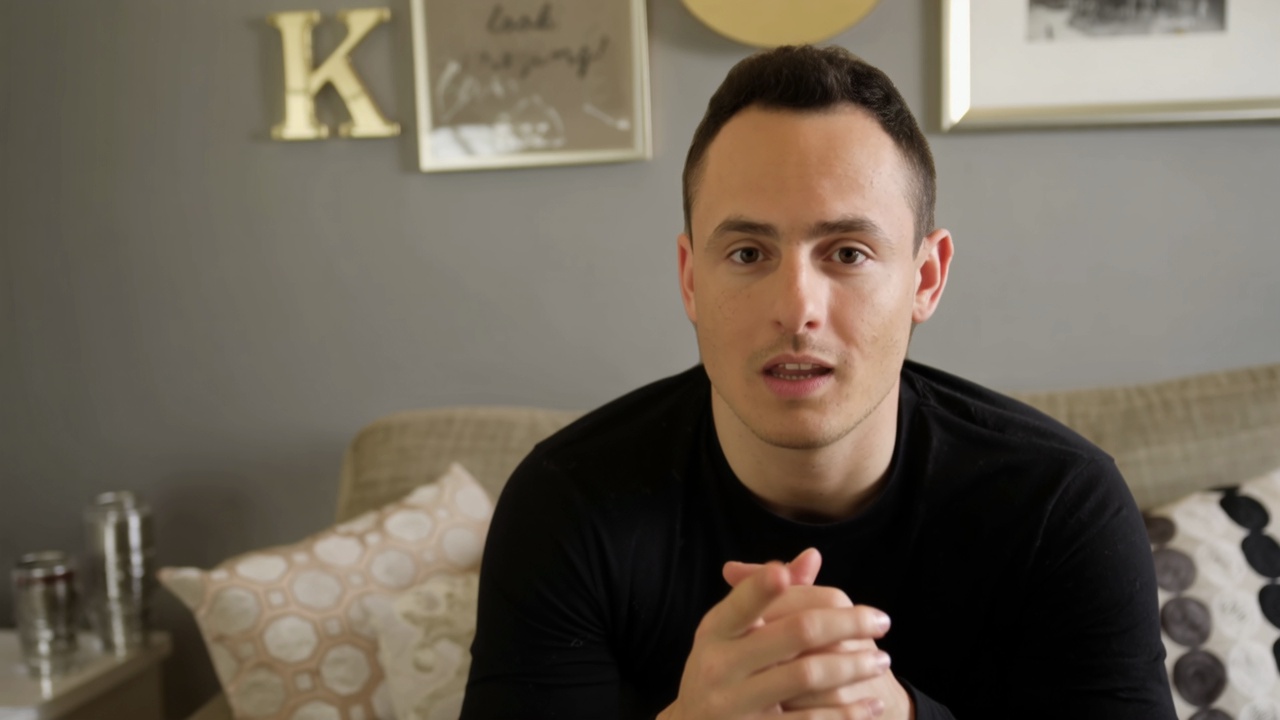} &
            \includegraphics[width=0.326\linewidth]{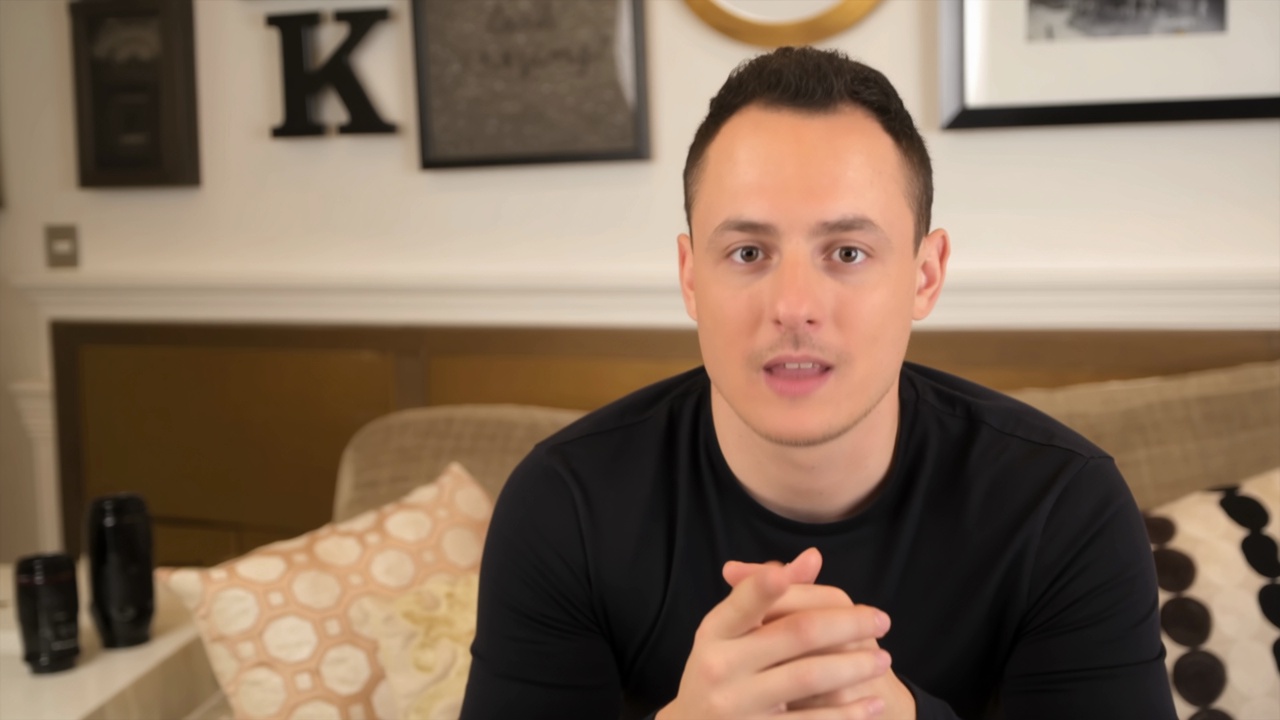} \\[-1pt]
            \includegraphics[width=0.326\linewidth]{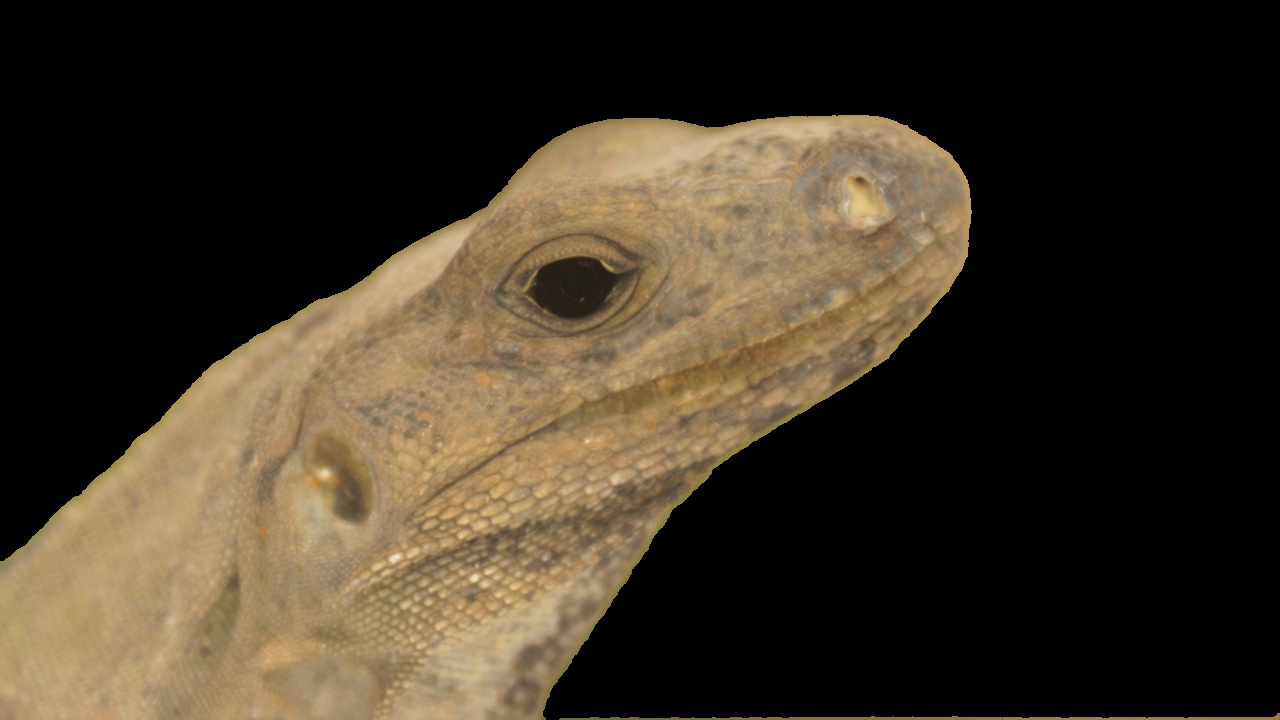} &
            \includegraphics[width=0.326\linewidth]{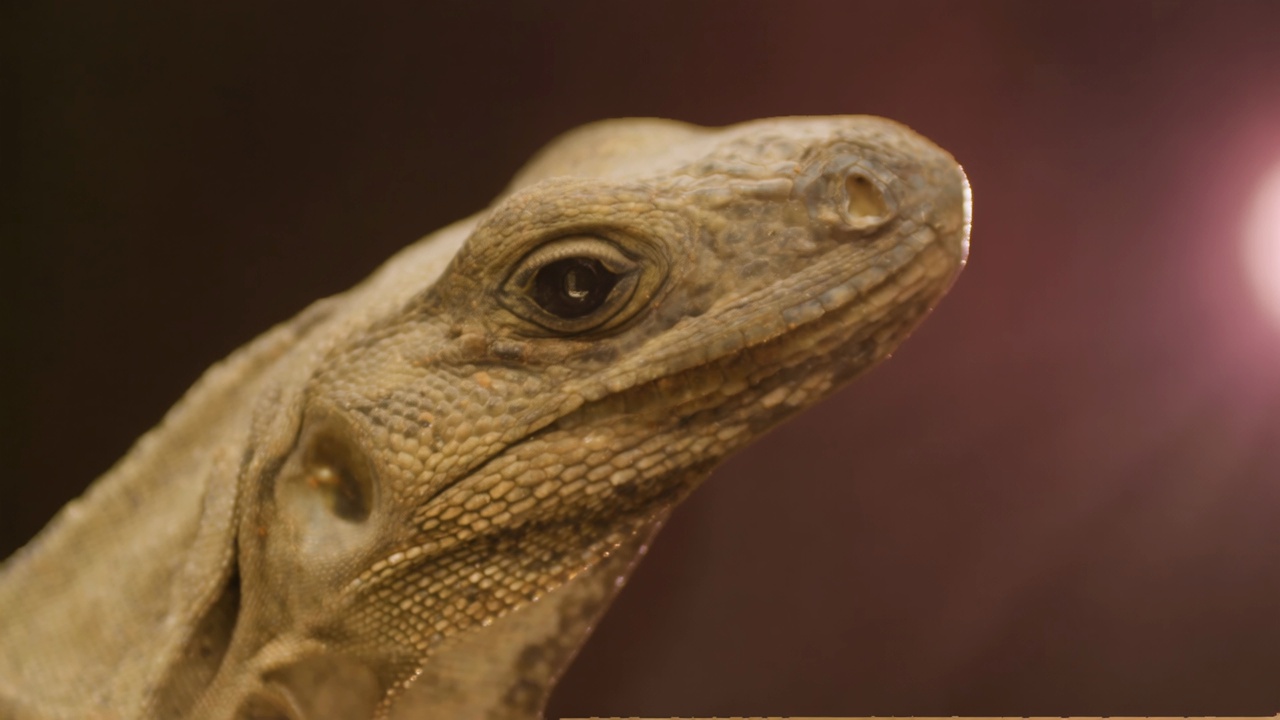} &
            \includegraphics[width=0.326\linewidth]{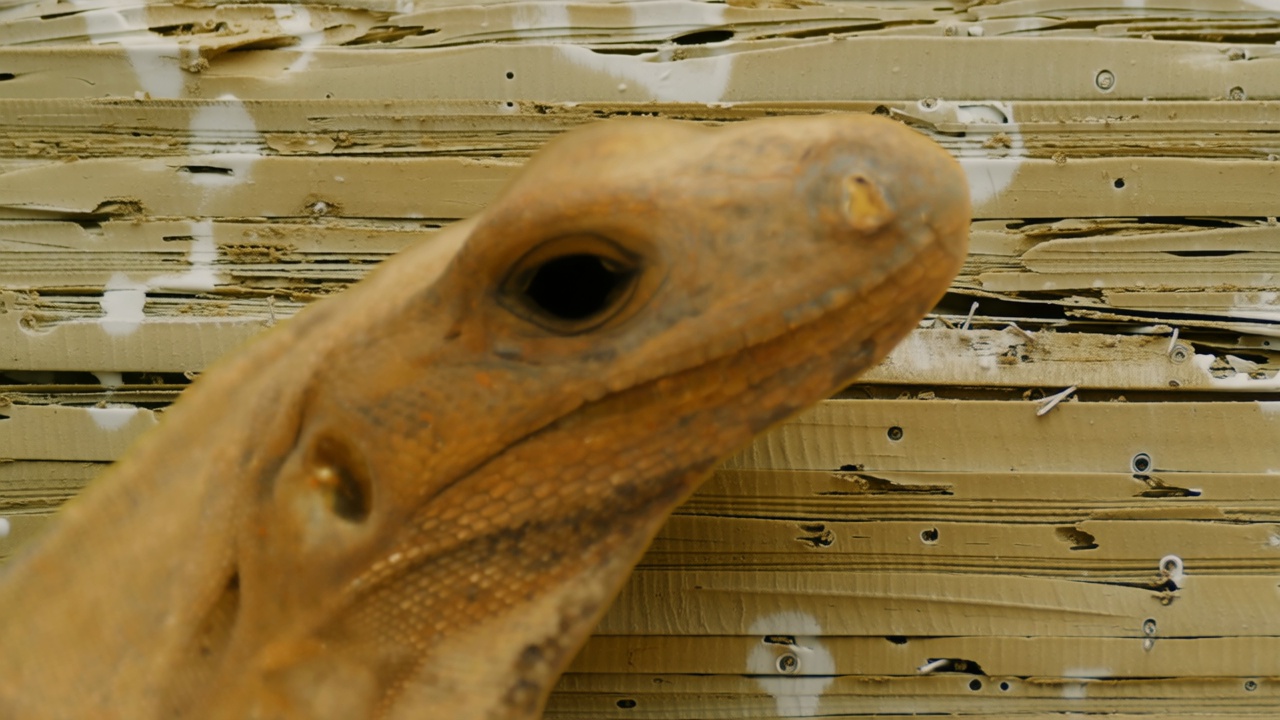}
        \end{tabular}
    \caption{\edit{Spatial conditioning dropout for sparse albedo guidance. We show sparse input albedo, where zero-valued regions mean ``no guidance'' rather than black albedo (left), the baseline \xtorgb model without segment-dropout training, which produces flat visuals in these regions (middle), and our segment-dropout model, which preserves visual richness in unguided regions (right).}}
    \label{fig:x2rgb_spatial_dropout}
\end{figure}

\paragraph{Spatial dropout} \edit{In \cref{fig:x2rgb_spatial_dropout}, we illustrate spatial conditioning dropout, which allows the user to specify albedo only in selected regions and let the model generate content freely elsewhere. The left column shows sparse input G-buffers with zero-valued regions that mean ``no guidance'' rather than ``black albedo''. The baseline \xtorgb model, which is not trained with segment dropout, tends to produce flat-looking visuals in these unguided regions (middle column). Training with randomly dropped segments allows the user to provide spatially sparse signals without losing visual richness in the unconditioned regions (right column).}
\edit{To train this model, we run the SAM 2 video segmentation model \cite{ravi2024sam} on the \textsf{(RGB, X)} training sequences, and randomly zero out albedo in some segments during training.}

\subsection{Streaming results}

\begin{figure*}[t]
    \vspace{1mm}
    \footnotesize
    \begin{minipage}{0.97\linewidth}
        \begin{overpic}[width=0.1915\linewidth]{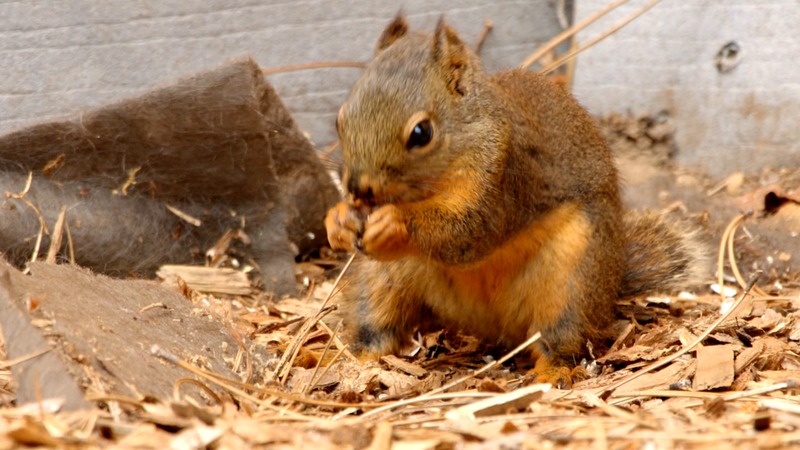}
        \put(29,59){Input RGB}
        \put(-7,24){\rotatebox[origin=c]{90}{Frame 0}} 
        \end{overpic}\hspace{1em}
        \begin{overpic}[width=0.1915\linewidth]{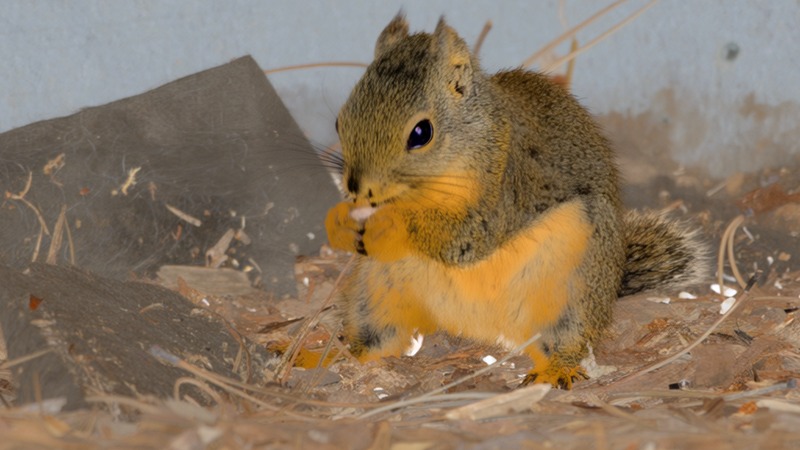}
        \put(17,59){Teacher Forcing (TF)}
        \put(-7,24){\rotatebox[origin=c]{90}{Albedo}} 
        \end{overpic}\hfill
        \begin{overpic}[width=0.1915\linewidth]{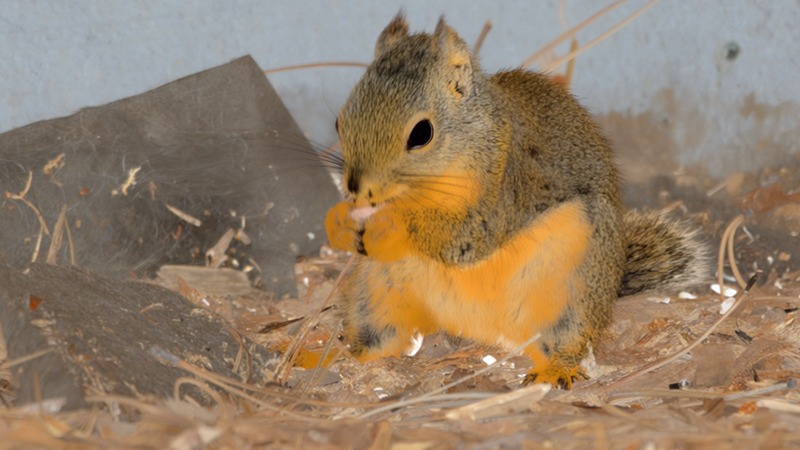}
        \put(18,59){$+$ Augmentations}
        \end{overpic}\hfill
        \begin{overpic}[width=0.1915\linewidth]{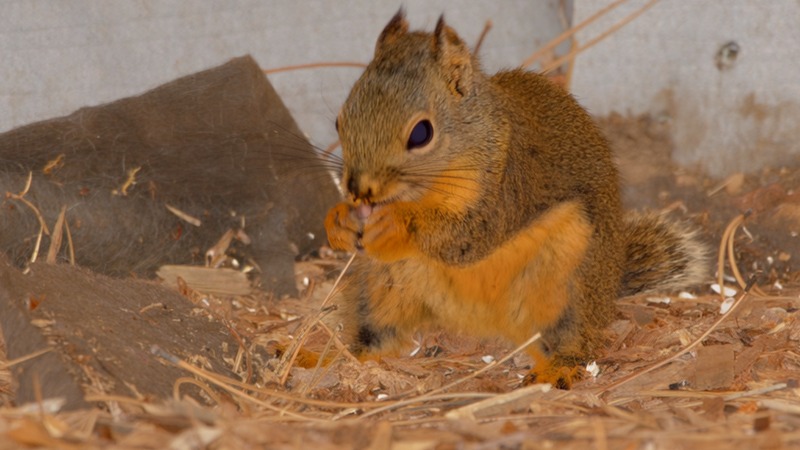}
        \put(24,59){$+$ Long-context}
        \end{overpic}\hfill
        \begin{overpic}[width=0.1915\linewidth]{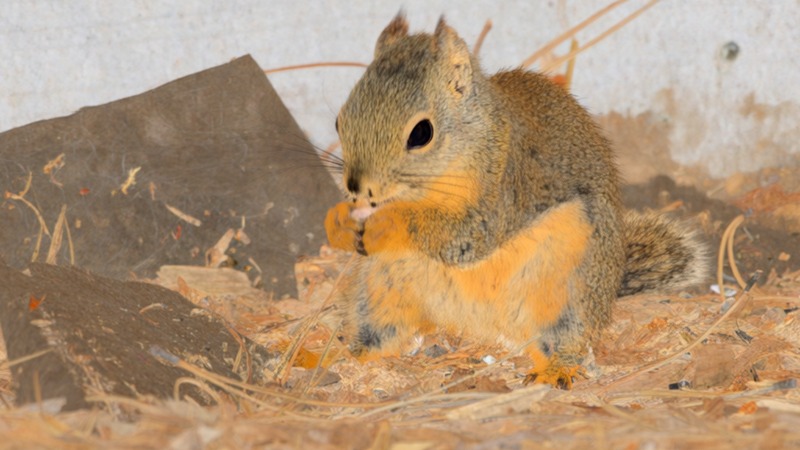}
        \put(18,59){$+$ Self Forcing (SF)}
        \end{overpic}
    \end{minipage}\par\bigskip
    \begin{minipage}{0.97\linewidth}
        \begin{overpic}[width=0.1915\linewidth]{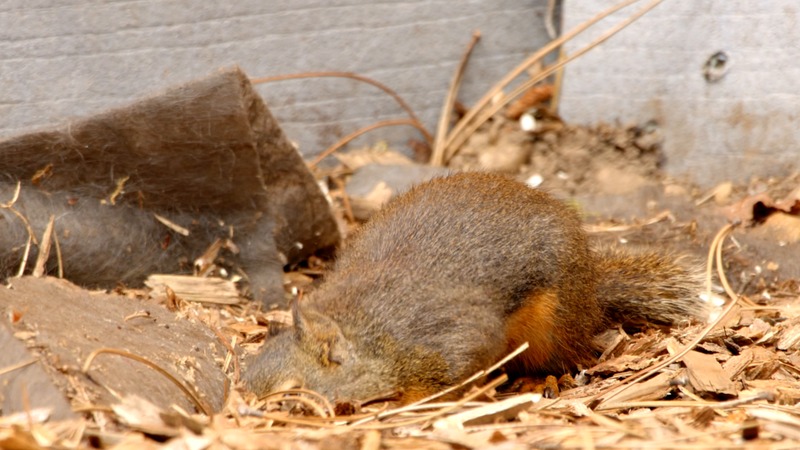}
        \put(45,60){\ldots}
        \put(-7,24){\rotatebox[origin=c]{90}{Frame 305}}
        \end{overpic}\hspace{1em}
        \begin{overpic}[width=0.1915\linewidth]{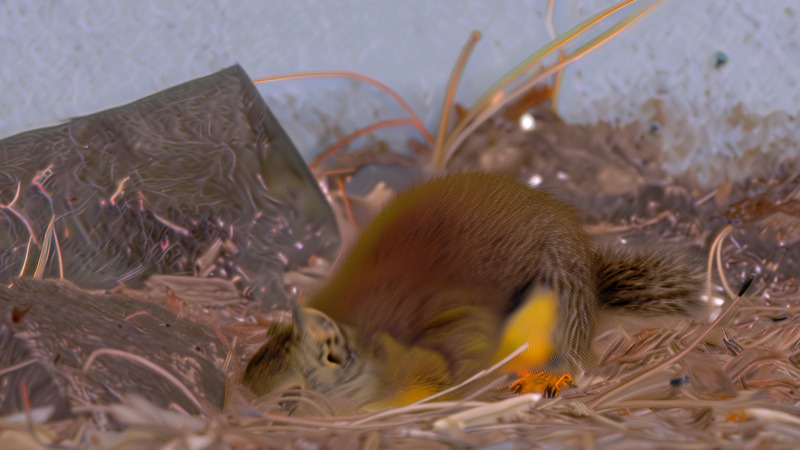}
        \put(-7,24){\rotatebox[origin=c]{90}{Albedo}} 
        \put(45,60){\ldots}
        \end{overpic}\hfill
        \begin{overpic}[width=0.1915\linewidth]{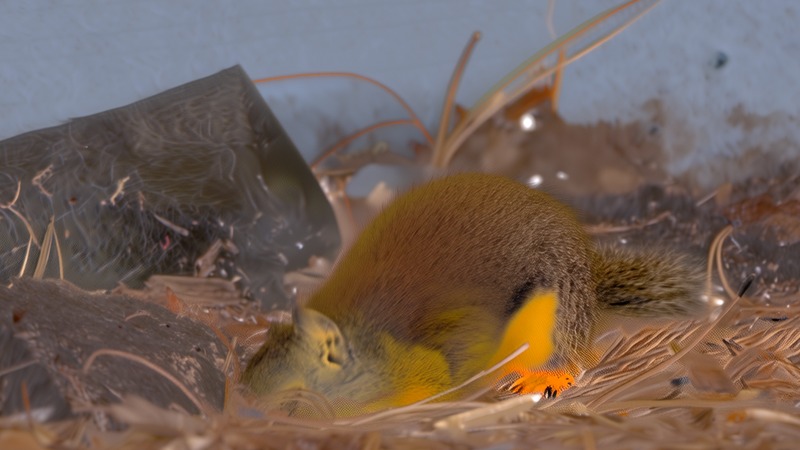}
        \put(45,60){\ldots}
        \end{overpic}\hfill
        \begin{overpic}[width=0.1915\linewidth]{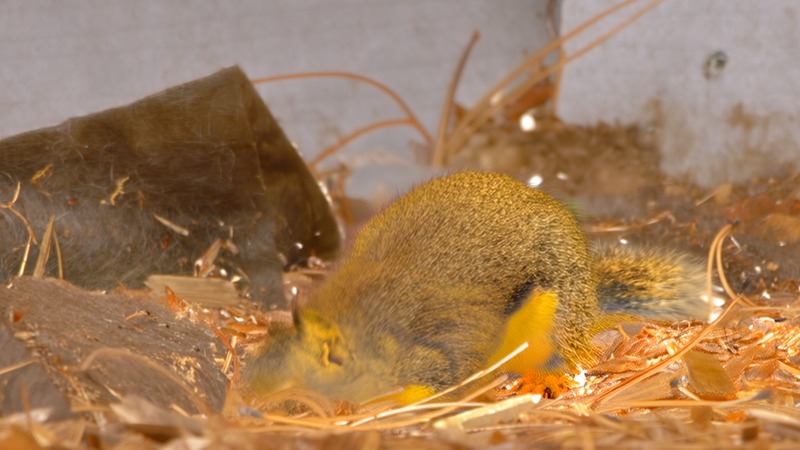}
        \put(45,60){\ldots}
        \end{overpic}\hfill
        \begin{overpic}[width=0.1915\linewidth]{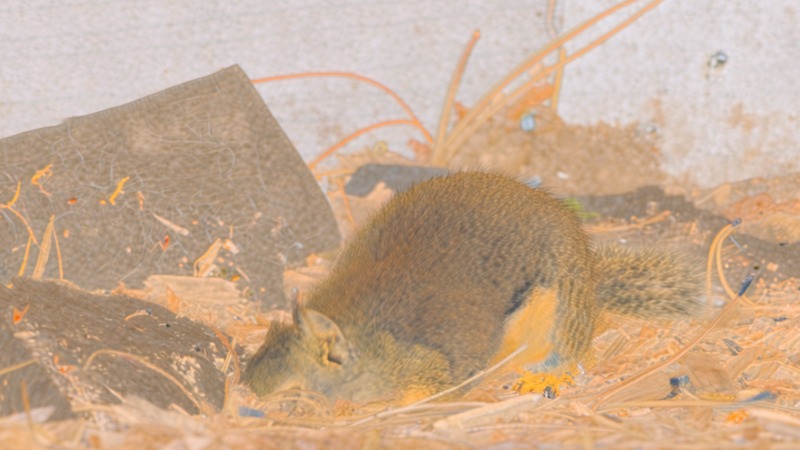}
        \put(45,60){\ldots}
        \end{overpic}
    \end{minipage}\par\bigskip
    \begin{minipage}{0.97\linewidth}
        \begin{overpic}[width=0.1915\linewidth]{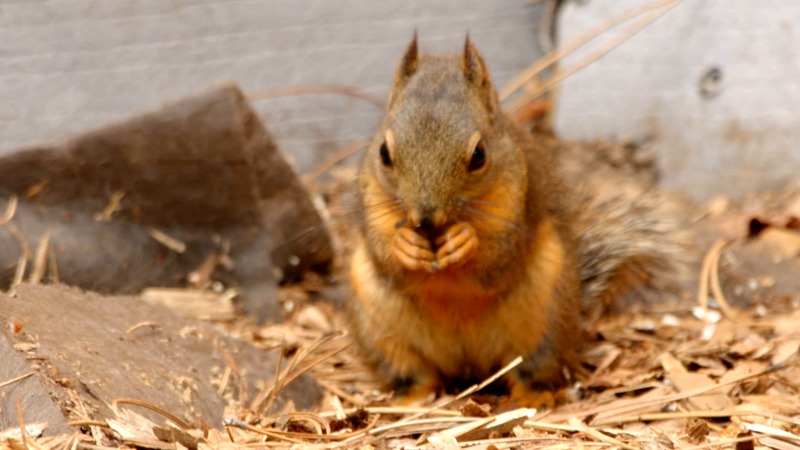}
        \put(45,60){\ldots}
        \put(-7,24){\rotatebox[origin=c]{90}{Frame 656}}
        \end{overpic}\hspace{1em}
        \begin{overpic}[width=0.1915\linewidth]{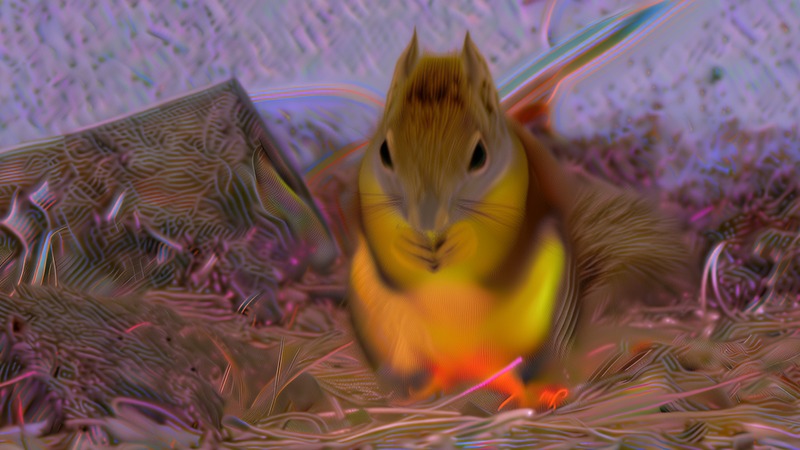}
        \put(-7,24){\rotatebox[origin=c]{90}{Albedo}}
        \put(45,60){\ldots}
        \put(45,-10){(a)}
        \end{overpic}\hfill
        \begin{overpic}[width=0.1915\linewidth]{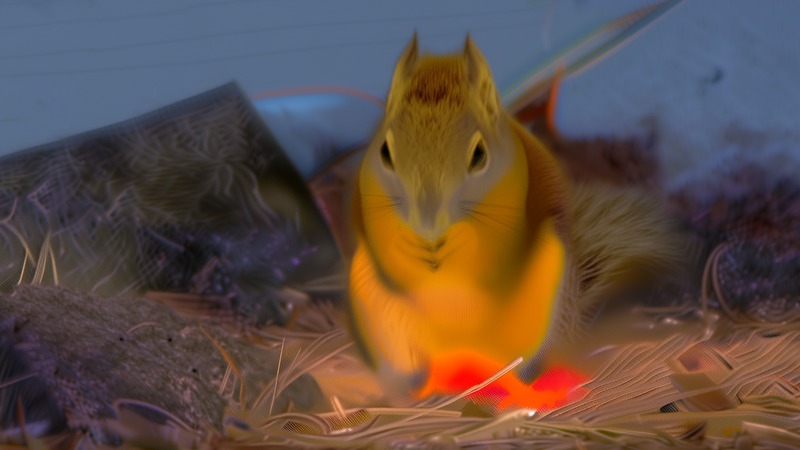}
        \put(45,60){\ldots}
        \put(45,-10){(b)}
        \end{overpic}\hfill
        \begin{overpic}[width=0.1915\linewidth]{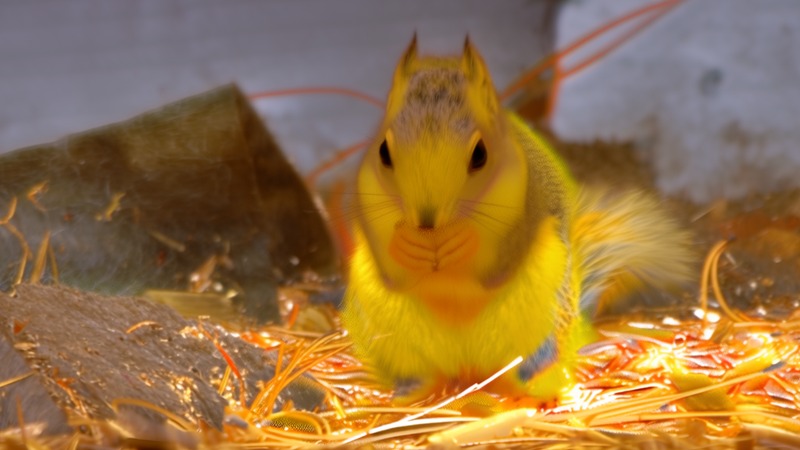}
        \put(45,60){\ldots}
        \put(45,-10){(c)}
        \end{overpic}\hfill
        \begin{overpic}[width=0.1915\linewidth]{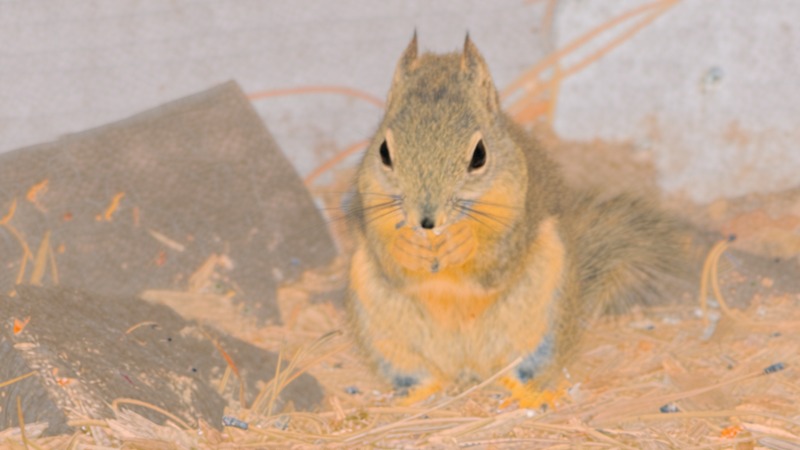}
        \put(45,60){\ldots}
        \put(45,-10){(d)}
        \end{overpic}
    \end{minipage}
    \caption{\edit{Ablation for $1/16$ streaming \rgbtox. Teacher forcing trains on ground-truth stitching frames but uses previous predictions at inference, leading to blur and error accumulation after roughly 100--200 frames (a). Stitching-frame augmentations help only slightly (b), and long-context tokens improve color preservation but do not eliminate drift (c). Combining these components with hybrid self forcing training yields stable estimates over hundreds of frames (d).}}
    \label{fig:rgb2x_streaming}
\end{figure*}

\begin{figure*}[t]
    \vspace{1mm}
    \footnotesize
    \begin{minipage}{1\linewidth}
        \begin{overpic}[width=0.196\linewidth]{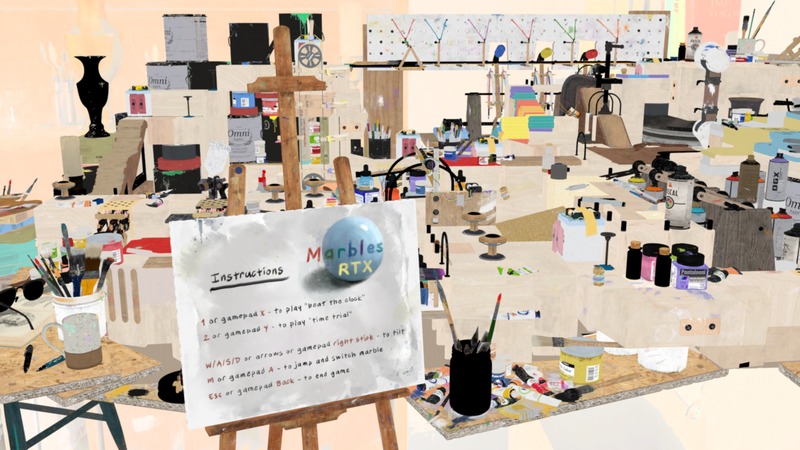}
        \put(27,58){Input G-buffers}
        \put(0,2){
        \begin{tikzpicture} \node[fill=black!10,fill opacity=0.75,rounded corners=1ex, text width=0.7cm,align=left] {Albedo}; \end{tikzpicture}
        }
        \end{overpic}\hfill
        \begin{overpic}[width=0.196\linewidth]{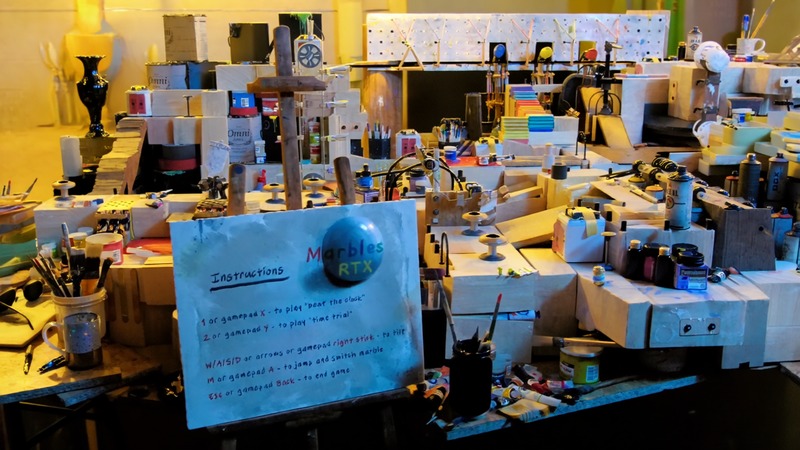}
        \put(35,58){Frame 0}
        \end{overpic}\hfill
        \begin{overpic}[width=0.196\linewidth]{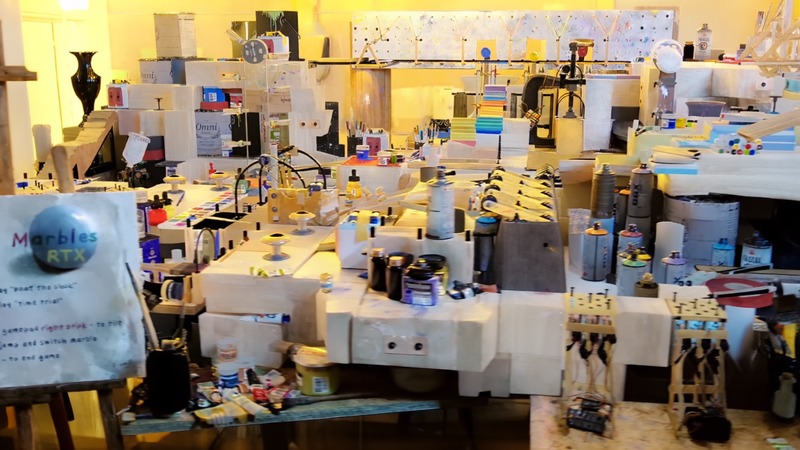}
        \put(35,58){Frame 100}
        \end{overpic}\hfill
        \begin{overpic}[width=0.196\linewidth]{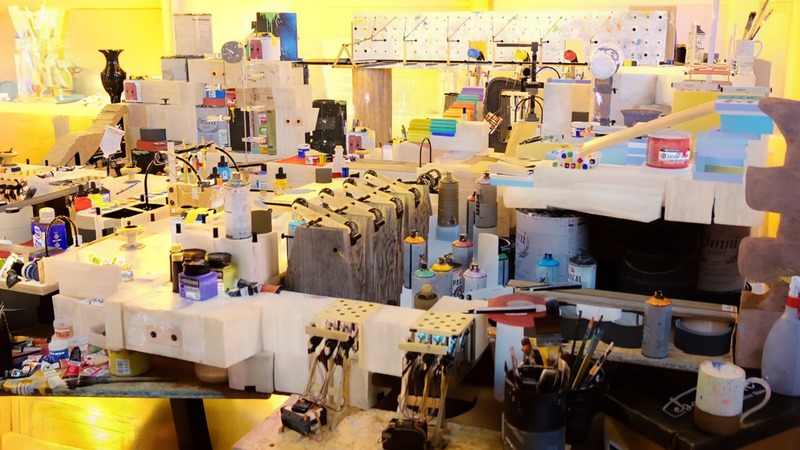}
        \put(35,58){Frame 200}
        \end{overpic}\hfill
        \begin{overpic}[width=0.196\linewidth]{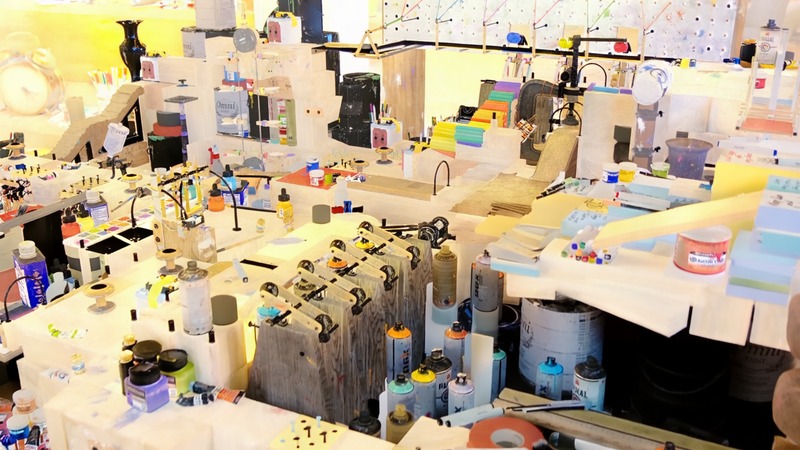}
        \put(35,58){Frame 300}
        \end{overpic}
    \end{minipage}\par\smallskip
    \begin{minipage}{1\linewidth}
    \begin{overpic}[width=0.196\linewidth]{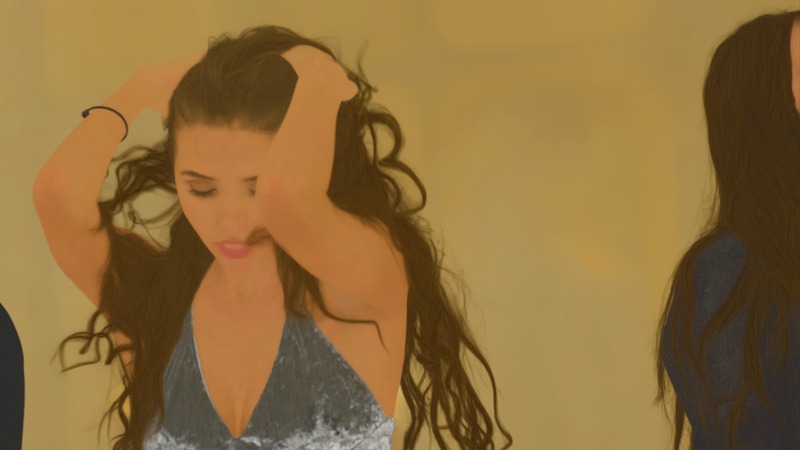}
        \put(0,2){
        \begin{tikzpicture} \node[fill=black!10,fill opacity=0.75,rounded corners=1ex, text width=0.7cm,align=left] {Albedo}; \end{tikzpicture}
        }
        \end{overpic}\hfill
        \begin{overpic}[width=0.196\linewidth]{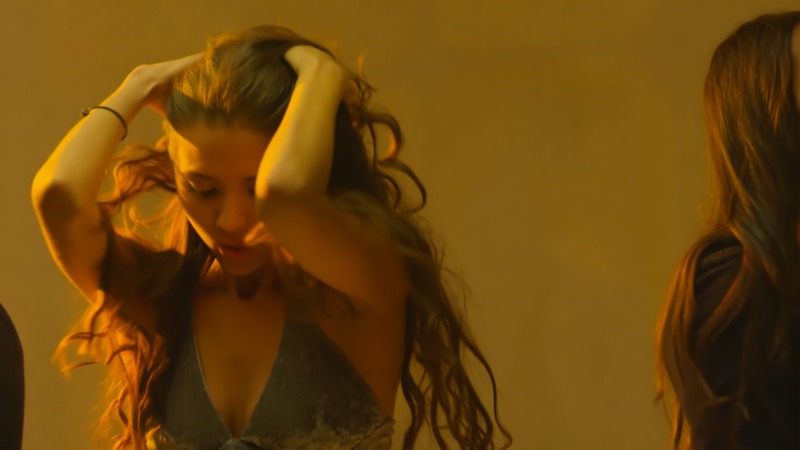}
        \end{overpic}\hfill
        \begin{overpic}[width=0.196\linewidth]{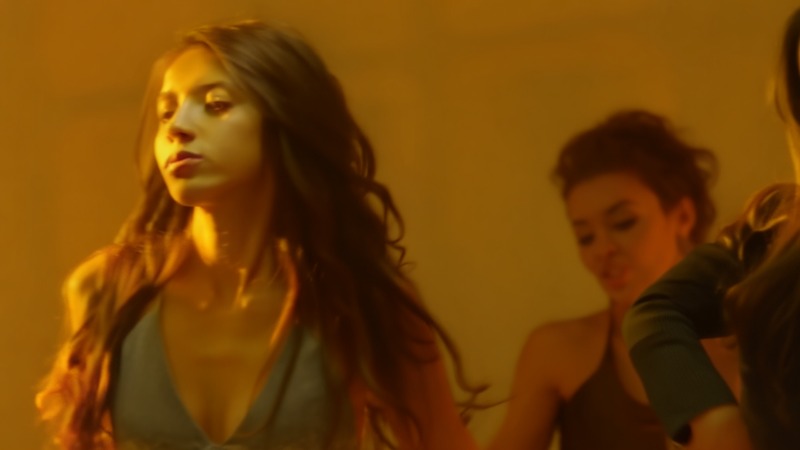}
        \end{overpic}\hfill
        \begin{overpic}[width=0.196\linewidth]{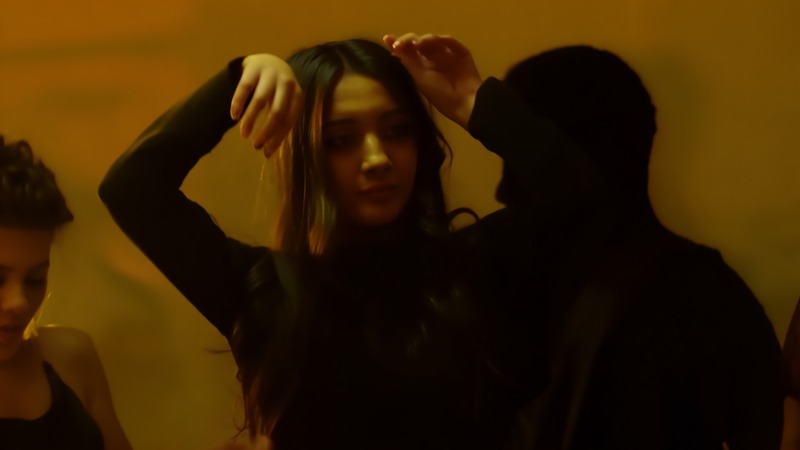}
        \end{overpic}\hfill
        \begin{overpic}[width=0.196\linewidth]{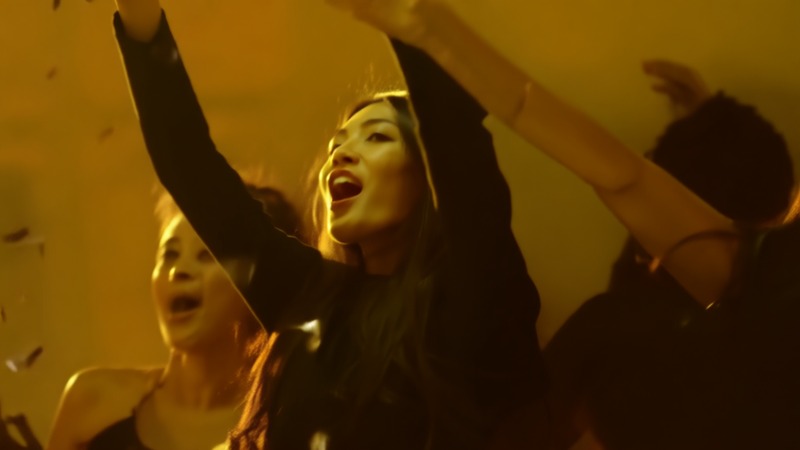}
        \end{overpic}
    \end{minipage}
    \caption{\edit{Streaming \xtorgb results with self forcing training, applied to synthetic albedo sequences (top) and estimated ones from real videos (bottom). The generated videos maintain temporal coherence and lighting consistency across frames; full sequences are provided in the supplementary material.}}
    \label{fig:x2rgb_streaming}
\end{figure*}

In \cref{fig:rgb2x_streaming}, we illustrate our \rgbtox streaming results, as well as ablations comparing to simpler variants. In (a) we illustrate the drifting issues of teacher forcing (training using ground truth data and inference using predicted data from the previous chunk). In (b), we apply augmentations of color and brightness in an attempt to break the perfection of baseline teacher forcing; this leads to only a slight improvement. In (c), we illustrate the more significant improvement from conditioning on a long-context frame (typically but not necessarily the first frame of the sequence). However, serious drifting still eventually becomes apparent on 600+ frame sequences. Finally, in (d) we show our final streaming \rgbtox model, combining the previous improvements with a mixed self forcing / teacher forcing training strategy. This approach leads to stable estimation without catastrophic drifting for about 1000 frames.

In \cref{fig:x2rgb_streaming}, we show the results of our streaming \xtorgb model trained to condition on albedo sequences, applied to synthetic (top) and estimated (bottom) albedo inputs. Despite not being conditioned on lighting, this model shows good temporal coherence and lighting consistency. Please see the supplementary zip file with an HTML page containing more streaming \rgbtox and \xtorgb videos.

\subsection{Discussion and limitations}

The optimality of G-buffers as a form of generative renderer guidance remains to be confirmed by further research, but we believe our experiments provide promising initial evidence, such as our extensive \xtorgb results with albedo and lighting conditioning shown in the video. This form of guidance is capable of controlling geometry, color and lighting, while leaving sufficient space for generative model creativity.

One limitation is that our models are currently no faster than the original, large Wan 2.1 14B model they are finetuned from. Our models could be seen as \emph{teacher models}, and could be distilled to fast interactive models by adapting DMD or similar teacher-student techniques. Rendering G-buffers is certainly fast, but full global
illumination may not be; future GPUs are likely to have more neural inference power and a distilled model based on our framework may soon become faster than classical Monte Carlo rendering, in addition to the benefits of learned realism priors.

 Our single-frame long-context solution could be extended to a whole video sequence or a true 3D context to provide improved memory. Our per-segment spatial dropout could be extended further, where some segments could be partially controlled, e.g., through a neural descriptor of the desired object / material, rather than having to specify either per-pixel detail or nothing. Finally, our models can often achieve realistic \xtorgb results even from fairly coarse and flat G-buffer content, but the right approach to produce 3D content for best results with generative rendering is a completely unstudied problem.

\section{Conclusion}

We introduced \name, a unified framework for generative forward and inverse rendering, capable of estimating G-buffers from images, videos and streams, and rendering realistic RGB outputs from G-buffer inputs. \edit{Our key contribution is a general recipe for finetuning diffusion transformer (DiT) models into generative forward and inverse renderers.

We use frame-wise concatenation to separate clean conditioning tokens from noisy output tokens, and add QK type embeddings and clean input tokens to make each token's role, modality, and noise level explicit. For \xtorgb models, we use \xx-patchify to pack multiple G-buffer inputs into one compact conditioning stream. We further use prompt guidance, condition blurring and dropout, spatial dropout, and lighting buffers to provide flexible control over realism, appearance, and illumination.

Our experiments show strong results in both directions: \rgbtox models recover temporally coherent G-buffers from RGB videos, while \xtorgb models synthesize realistic RGB videos from synthetic or estimated G-buffer inputs. We further extend both directions to streaming by combining teacher forcing, self forcing and long-context techniques.} We believe our trained models, as well as the overall framework introduced in this paper, can be useful for future progress in generative forward and inverse rendering.

% \begin{acks}
% \end{acks}

\bibliographystyle{ACM-Reference-Format}
\bibliography{bibtex}
% \input{10-appendix}
%\newpage

\end{document}